\documentclass[10pt,journal,compsoc]{IEEEtran}

\ifCLASSOPTIONcompsoc
  \usepackage[nocompress]{cite}
\else
  \usepackage{cite}
\fi

\ifCLASSOPTIONcompsoc
\usepackage[caption=false,font=normalsize,labelfon
t=sf,textfont=sf]{subfig}
\else
\usepackage[caption=false,font=footnotesize]{subfi
g}
\fi

\usepackage{graphicx}
\usepackage{amsmath}
\usepackage{xcolor}
\usepackage{booktabs}
\usepackage{lipsum}
\usepackage{multirow}
\usepackage{tabularx}

\usepackage{array}
\newcolumntype{H}{>{\setbox0=\hbox\bgroup}c<{\egroup}@{}}

\newif\ifincludecomment
\includecommenttrue 

\ifincludecomment
  
\else
  \NewEnviron{guidance}{}{}
\fi
\ifincludecomment
\newcommand{\maybecomment}[1]{\todo[color=olive!40]{#1}} 
\newcommand{\maybetohere}[1]{\todo[color=red!40]{#1}} 
\newcommand{\maybedelete}[1]{\todo[color=blue!40]{#1}} 

\else
  \newcommand{\maybecomment}[1]{}
  
\newcommand{\maybedelete}[1]{} 
\fi
\newcommand{\amostohere}[1]{{\color{black}\maybetohere{AMOS HERE}}}

\makeatletter
\newcommand{\ssymbol}[1]{$\ ^{\@fnsymbol{#1}}$}
\makeatother
\newcommand{\specialcell}[2][c]{%
  \begin{tabular}[#1]{@{}c@{}}#2\end{tabular}}
\newcommand\longtail{long$-$tail}

\begin{document}

\title{Few-Shot Learning with Class Imbalance}

\author{Mateusz~Ochal,
        Massimiliano Patacchiola,
        Jose Vazquez,
        Amos Storkey,
        Sen Wang
        \IEEEcompsocitemizethanks{%
        \IEEEcompsocthanksitem Mateusz Ochal and Sen Wang are with School of Engineering and Physical Sciences, Heriot-Watt University, Edinburgh, UK. \protect E-mail: m.ochal@hw.ac.uk%
        \IEEEcompsocthanksitem  Massimiliano Patacchiola and Amos Storkey were with School of Informatics, University of Edinburgh, Edinburgh, UK. Massimiliano is now with the Department of Engineering, University of Cambridge, UK.%
        \IEEEcompsocthanksitem Jose Vazquez is with SeeByte Ltd., Edinburgh, UK.%
}
\thanks{Manuscript received April 30, 2021}
}

\markboth{Under Review}%
{Ochal \MakeLowercase{\textit{et al.}}: Few-Shot Learning with Class Imbalance}

\IEEEtitleabstractindextext{%
\begin{abstract}
Few-Shot Learning (FSL) algorithms are commonly trained through Meta-Learning (ML), which exposes models to batches of tasks sampled from a meta-dataset to mimic tasks seen during evaluation. However, the standard training procedures overlook the real-world dynamics where classes commonly occur at different frequencies. While it is generally understood that class imbalance harms the performance of supervised methods, limited research examines the impact of imbalance on the FSL evaluation task. Our analysis compares 10 state-of-the-art meta-learning and FSL methods on different imbalance distributions and rebalancing techniques. Our results reveal that 1) some FSL methods display a natural disposition against imbalance while most other approaches produce a performance drop by up to 17\% compared to the balanced task without the appropriate mitigation; 2) contrary to popular belief, many meta-learning algorithms will not automatically learn to balance from exposure to imbalanced training tasks; 3) classical rebalancing strategies, such as random oversampling, can still be very effective, leading to state-of-the-art performances and should not be overlooked; 4) FSL methods are more robust against meta-dataset imbalance than imbalance at the task-level with a similar imbalance ratio ($\rho<20$), with the effect holding even in long-tail datasets under a larger imbalance ($\rho=65$).
\end{abstract}

}

\maketitle

\IEEEraisesectionheading{\section{Introduction}\label{sec:introduction}}

\IEEEPARstart{D}{eep} learning methods are well-known for their state-of-the-art performance on a variety of tasks, but they often require training on large labeled datasets to generalize well to new examples. However, large datasets can be costly to obtain and annotate \cite{LeCun2015dl}. This is a particularly limiting issue in many real-world situations due to the need to perform real-time operations, the presence of rare categories, or the desire for a good user experience \cite{Ochal2020oceans,Guan2020aerial,Zhang2020ciloss,Massiceti2021orbit}. Few-Shot Learning (FSL) alleviates this burden by defining a distribution over \emph{tasks}, with each task containing a few labeled data points (\emph{support set}) and a set of target data (\emph{query set}) belonging to the same set of classes. A common way to train FSL methods is through Meta-Learning (ML). One such ML technique is \emph{episodic (meta-) training} \cite{Vinyals2017matching} where the model is repeatedly exposed to batches of tasks sampled from a task distribution and then tested on a different but similar distribution in the meta-testing phase. We use the prefix \emph{“meta”} to distinguish between the high-level training and evaluation routines of meta-learning (outer loop / pre-training) from the training and evaluation routines at the single-task level (inner loop / fine-tuning). 

\begin{figure*}[t]
    \includegraphics[width=\linewidth]{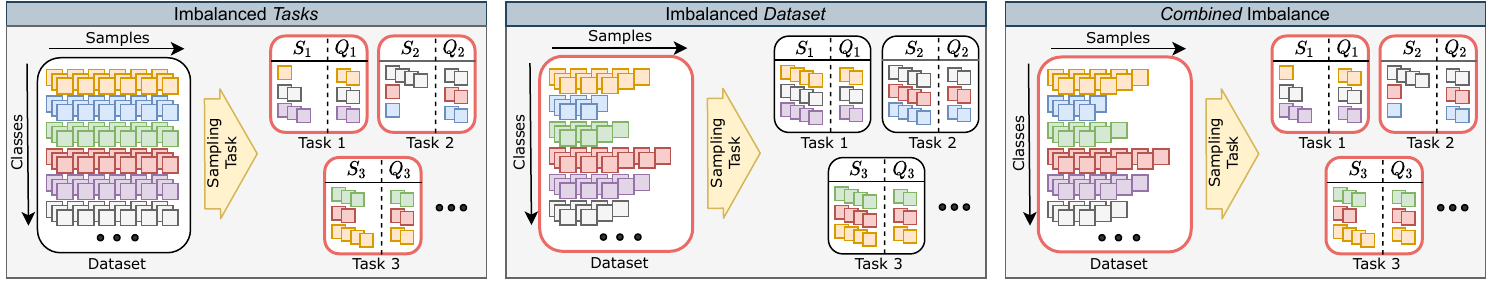}
    \caption{The \emph{Class Imbalanced Few-Shot Learning (CIFSL)} problem investigated in this work. We evaluate imbalance at the \emph{task-level} (left) in Section~\ref{sec imb task}, \emph{dataset-level} (middle) in Section~\ref{sec imb dataset} and \emph{combined} (right) in Section~\ref{sec imb combined}. We study imbalance under several imbalance distributions: $linear$ ({task 1}), $step$ ({task 2}), and $random$ ({task 3}). We also study the effects of $\longtail$ imbalance at the meta-dataset level.}
    \label{fig visualisation_diagram}
\end{figure*}

The FSL task serves as a practical and convenient benchmark for ML methods. However, the standard setting assumes that classes contain the same number of data points. In contrast, in many applications, the number of samples can vary and follow a wide range of imbalance distributions \cite{Guan2020aerial,Zhang2020ciloss,Buda2018imbalance,Leevy2018bigdata,Japkowicz2002imbalance}. While it is generally understood that imbalance harms the performance of supervised methods \cite{Liu2019tailed,Salakhutdinov2011longtail} when trained and evaluated on the same classes, limited research exams its impact on the ML generalization to novel classes and the FSL evaluation task. This paper focuses on these issues and provides a deep analysis into the \emph{Class Imbalance Few-Shot Learning (CIFSL)} problem. We identify and examine three levels of class imbalance: task-level, dataset-level, and combined (task-level and dataset-level) imbalance. In contrast to previous work on CIFSL \cite{Triantafillou2019meta, Lee2020baysiantaml, Chen2020mamlstop, Wertheimer2019metainat}, we explicitly attribute and quantify the impact on the performance caused by class imbalance for each model. Moreover, we study multiple class imbalance distributions, giving a realistic assessment of performance and revealing previously unknown strengths and weaknesses of 10 state-of-the-art algorithms. Additionally, we offer practical advice, identifying ways to boost performance and providing an improved procedure as a baseline for CIFSL. Figure~\ref{fig visualisation_diagram} shows a graphical representation of the CIFSL problem. 

\textbf{Task Level.} Given the nature of FSL task, a small difference in the number of class samples can already introduce significant levels of imbalance, affecting model performance. This source of imbalance can originate from the meta-evaluation tasks (reflecting deployment of meta-trained models in the real-world \cite{Ochal2020oceans,Guan2020aerial}) or from the meta-training tasks (mimicking online meta-learning scenarios \cite{Finn2019online} where the task distribution might not be visible or easily controlled, e.g., due to data privacy and Federated Meta-Learning \cite{Chen2018federated}).

\textbf{Dataset Level.} Standard FSL benchmarks (e.g., Omniglot, Mini-ImageNet) assume an equal number of data-points for each class in the meta-dataset. However, in many real-world applications, it is desirable to combine or exchange those datasets with domain-specific datasets tailored towards a specific application and whose underlying class distribution is likely to be imbalanced \cite{Guan2020aerial,Ochal2020oceans,Massiceti2021orbit}. In some real-world settings, the dataset could be immeasurably large or even unknown, like in online class imbalance \cite{wang2014resampling}. Although recent meta-dataset FSL benchmarks (such as Meta-Dataset \cite{Triantafillou2019meta} or meta-iNat \cite{Wertheimer2019metainat}) contain some degrees of dataset imbalance, its impact on the ML generalization to the FSL evaluation tasks has remained largely unexplored.%

\textbf{Combined Level.} While studying both levels individually is important, realistic settings are likely to induce both levels of imbalance simultaneously. In our paper, we study the impact of combined levels and show that it is compounding -- yielding significantly worse performance drop than the sum of its components taken individually. 

\textbf{Contributions.} Our main contributions and findings can be summarized as follows:

\begin{enumerate}
    \item We present a large empirical investigation on a formalized CIFSL problem, showing its impact on 10 meta- and few-shot learning algorithms under various imbalance distributions (linear, step, random). As far as we know, this is the most exhaustive empirical study of CIFSL available in the literature.
    \item We show that contrary to the spirit of meta-learning, meta-training on imbalanced tasks through Random-Shot Meta-Training \cite{Triantafillou2019meta, Lee2020baysiantaml, Chen2020mamlstop, Guan2020aerial} does not necessarily work as effectively as Standard Meta-Training \cite{Vinyals2017matching} when systematically tested under various task imbalance distributions and conditions.
    \item We adapt methods used in the imbalanced supervised learning literature (e.g., Random Over-Sampling and various re-balancing losses) to the CIFSL setting. These adaptations are easy to implement and show state-of-the-art performance under many conditions (e.g., different backbones, shots, and datasets).
    \item We provide quantitative insights into the dataset imbalance problem and show that meta-learners are robust to meta-dataset imbalance under various imbalance distributions, dataset sizes, and moderate cross-domain shift. The impact of imbalance at the task level is much more significant in comparison. 
\end{enumerate}

\section{Related Work}\label{sec related}
\subsection{Class Imbalance}
Imbalance occurs when at least one class (the majority class) contains a higher number of samples than others (the minority classes). If uncorrected, conventional supervised losses, such as (multi-class) cross-entropy, skew the learning process, introducing bias and poor generalization towards the minority class  \cite{Buda2018imbalance,Leevy2018bigdata}.

Class imbalance has been studied using real-world datasets and distributions that approximate real-world dynamics \cite{Buda2018imbalance,Johnson2019imbalance,Liu2019tailed}. A significant portion of the community uses artificial imbalance distributions, such as \emph{linear} and \emph{step} \cite{Buda2018imbalance}. At large-scale, datasets with many classes and samples tend to follow a \emph{long-tail} distribution \cite{Salakhutdinov2011longtail,Reed2001powerlaw}, with most classes occurring infrequently (tail classes) and only a few occurring with high frequency (head classes). 
Classical long-tail literature commonly concentrates on the generalization performance to all dataset classes and the impact caused by the large imbalance ratios \emph{between} the head and tail classes \cite{Liu2019tailed}. In contrast, our analysis focuses on the generalization performance to novel few-shot tasks, including the impact caused by imbalance present \emph{within} the novel few-shot classes \cite{Vinyals2017matching,Chen2019closer,Triantafillou2019meta}. Moreover, to reflect the unpredictability of real-world class imbalance, our analysis assumes no prior knowledge of imbalance, and we model imbalance according to multiple distributions (for more details, see Section~\ref{sec methodology}).

Classical imbalance approaches can be categorized into three groups: data-level, algorithm-level, and hybrid \cite{Buda2018imbalance}.
Popular data-level methods include Random Over-Sampling (ROS) 
\cite{Japkowicz2002imbalance}, which randomly resamples data points from the minority classes. 
\emph{Algorithm-level} strategies use regularization or minimization of loss/cost functions. Weighted loss is a common approach where the loss of each sample is weighted by the inverse frequency of its class. Focal loss \cite{Lin2017focal} down-weighs easy predictions while focusing the training on samples that are harder to classify. CB loss \cite{Cui2019cbloss} augments the weighted loss based on the effective number of samples. \emph{Hybrid} methods combine one or more strategies (e.g., Two-Phase Training, \cite{Havaei2017twophase}). Although CI strategies have been studied in various object recognition tasks, limited research evaluates their generalization to recent FSL methods.

\subsection{Few-Shot Learning.}
FSL aims to train models using a limited number of samples by meta-learning or pretraining on a base set of classes. FSL methods can be broadly categorized into metric-learning \cite{Snell2017proto,Sung2017relationnet,Edwards2017,Vinyals2017matching}, optimization-based \cite{Finn2017maml,Ravi2017}, hallucination \cite{Zhang2018meta_gan,Antoniou2017gan_augment}, transfer-learning \cite{Chen2019closer,Dhillon2020baseline}, and probabilistic approaches \cite{Patacchiola2019gpshot,zhang2021shallow}. Additionally, these methods can be paired with \emph{transductive inference} mechanisms \cite{Antoniou2019sca, liu2019learning, Dhillon2020baseline, boudiaf2020transductive} which take advantage of the query set as an additional unlabeled set for task adaptation. In this report, we select at least one representative from all model categories, with the exception of hallucination and transductive inference categories that typically work on top of other FSL methods and utilize additional samples beyond the original support set. Related to our work is \emph{Incremental Few-Shot Learning} \cite{Ren2019ifsl,Gidaris2018ifsl,Hariharan2017sgm} also known as \emph{Generalized Few-Shot Learning} \cite{Dwivedi_2019_ICCV, Li_2019_ICCV} that extends the standard FSL task and also measures performance on the base set of classes (e.g. the meta-training dataset). In contrast, our work is more aligned with standard FSL and focuses on the problem of imbalance with novel classes, which suffices in many applications \cite{Guan2020aerial, Ochal2020oceans}.

\subsection{Imbalance in Few-Shot Learning}\label{sec related cifsl}

\subsubsection{Task Imbalance}

Class imbalance, in the context of standard FSL, has received some attention, although the current work is not comprehensive \cite{Guan2020aerial, Triantafillou2019meta, Lee2020baysiantaml, Chen2020mamlstop}. At the task level, class imbalance occurs in the support set or the query-set, directly affecting the learning and evaluation procedures. \cite{Triantafillou2019meta} uses imbalanced support sets to create a more realistic and challenging benchmark for meta-learning. The authors use random-shot tasks with randomly selected classes (way) and samples (shot), which replace the balanced task during meta-training and meta-evaluation. A similar idea is explored by \cite{Lee2020baysiantaml}, \cite{Chen2020mamlstop}, and \cite{Guan2020aerial} with the last two using a fixed number of classes (way). However, none of these works quantify the impact of class imbalance on the FSL task nor the advantages of Random-Shot meta-training. Moreover, previous work mainly focused on a single imbalance distribution \cite{Zhang2020ciloss,Guan2020aerial} or a small selection of methods \cite{Chen2020mamlstop}, or mixed several factors of variation into a single evaluation task \cite{Triantafillou2019meta,Lee2020baysiantaml}, making it challenging to attribute any performance change purely to class imbalance. In comparison, our work provides a more comprehensive view of CIFSL by examining a wider range of methods, imbalance distributions, and rebalancing techniques. 

Related to, but distinct from, our CIFSL settings is \emph{task-distribution} imbalance \cite{Lee2020baysiantaml}, where the underlying task-distribution seen during meta-training differs from meta-evaluation task-distribution. Task-distribution imbalance has already received some attention \cite{Lee2020baysiantaml,Cao2020shotanalysis} and it will only be discussed tangentially in this work.

\subsubsection{Dataset Imbalance}
FSL methods \cite{Snell2017proto,Sung2017relationnet,Edwards2017,Vinyals2017matching,Finn2017maml,Ravi2017,Chen2019closer,Dhillon2020baseline,Patacchiola2019gpshot,zhang2021shallow} are typically compared against balanced benchmarks (eg. Omniglot, MiniImageNet). More recent benchmarks \cite{Triantafillou2019meta, Wertheimer2019metainat} contain some levels of imbalance in the meta-dataset. Specifically, Meta-Dataset \cite{Triantafillou2019meta} combines datasets of different size (e.g. Omniglot, CUB, Fungi, Aircraft, ImageNet etc.) into a single corpus. A different dataset, meta-iNat \cite{Wertheimer2019metainat}, models imbalance according to a long-tail distribution. 
However, no significant work quantifies the impact of imbalance at the \emph{dataset-level}. 
It can be cumbersome to evaluate the imbalance of Meta-Dataset and meta-iNat specifically, as it requires access to a balanced version of these datasets with an equal total number of samples. Therefore, in our analysis, we artificially induce imbalance into datasets and model imbalance according to various distributions approximating many real-world scenarios. 
 
\subsubsection{Combined Imbalance}
Although \cite{Triantafillou2019meta} considers the combined imbalanced setting implicitly through imbalanced tasks and combining datasets of various sizes, little insight is offered for this level. Studying the combined level offers the most realistic settings, and we show that it yields a significantly worse performance drop than the sum of the levels taken individually. 

\section{Problem Definition}\label{sec methodology}
\subsection{Standard Task/Meta-Training.} 
Benchmarking FSL methods typically involves three phases: meta-training/pre-training, meta-validation, and meta-testing. Each phase samples batches of data points or tasks from a separate dataset: $\mathcal{D}_{train}$, $\mathcal{D}_{val}$, and $\mathcal{D}_{test}$, respectively, such that the classes and samples between each of the datasets are non-overlapping. We assume that a dataset $\mathcal{D}$ is $balanced$ when it contains $N^\mathcal{D}$ classes and $K^\mathcal{D}$ samples for each class. Similarly, a standard $K^\mathcal{S}$-shot $N^\mathcal{S}$-way FSL classification task is defined by a small \emph{support set}, $\mathcal{S} = \left\{ (x_1, y_1), \dots , (x_{s}, y_{s}) \right\} \sim \mathcal{D}$, containing $N^\mathcal{S} \times K^\mathcal{S}$ image-label pairs drawn from $N^\mathcal{S}$ unique classes with $K^\mathcal{S}$ samples per class ($|\mathcal{S}|=K^\mathcal{S} \times N^\mathcal{S}$ and $K^\mathcal{S} \ll K^{\mathcal{D}}$ and $N^\mathcal{S} \ll N^{\mathcal{D}}$). The goal is to minimize some loss over a \emph{query set}, $\mathcal{Q} = \left\{ (x_1, y_1), \dots , (x_{q}, y_{q}) \right\}  \sim \mathcal{D}$, containing a different set of $K^\mathcal{Q}$ samples drawn from the same $N^\mathcal{S}$ classes (i.e. $N^\mathcal{Q}=N^\mathcal{S}$, $\mathcal{Q}^{(x)} \cap \mathcal{S}^{(x)} = \emptyset$ and $\mathcal{Q}^{(y)}\equiv \mathcal{S}^{(y)}$).

\subsection{Imbalanced Tasks/Meta-Dataset.} 
For brevity, but without loss of generality, we define a distribution for a set of data points $\mathcal{*}\in\{\mathcal{D},\mathcal{S},\mathcal{Q}\}$ as a tuple ($K_{min}^{\mathcal{*}}$, $K_{max}^{\mathcal{*}}$, $N^{\mathcal{*}}$, $M^{\mathcal{*}}$) for a distribution $\mathcal{I} \in \{linear,step,random,\longtail\}$ \cite{Buda2018imbalance,Wertheimer2019metainat}, where $K_{min}^{\mathcal{*}}$ is the minimum number of samples per class, $K_{max}^{\mathcal{*}}$ is the maximum number of samples per class, $N^{\mathcal{*}}$ is the number of classes, and $M^{\mathcal{*}}$ is an additional parameter used for $step$ imbalance. Specifically, we define $\mathcal{I}$ distributions as:
 \begin{itemize}
     \item \emph{Linear imbalance}. The $K^\mathcal{*}_{i}$ number of samples for each class $i\in\{1..N^\mathcal{*}\}$ is defined by:
     \begin{equation}
        \begin{split}
            K^\mathcal{*}_{i} = \mathtt{round} (
            K^\mathcal{*}_{min}-c + (i - 1) \times \\ (K^\mathcal{*}_{max}+ 2 \times c - K^\mathcal{*}_{min}) / (N^\mathcal{*}-1) ) ,
        \end{split}
     \end{equation}
     where $c=0.499$ for rounding purposes. For example, this means that for $linear$ (1,9,5,-) set,  $K^\mathcal{*}_{i} \in \{1, 3, 5, 7, 9\}$, and for $linear$ (4,6,5,-) set, $K^\mathcal{*}_{i} \in \{4, 4, 5, 6, 6\}$. 
     \item \emph{Step imbalance}. The number of class samples, $K^\mathcal{*}_{i}$, is determined by an additional variable $M^\mathcal{*}$ specifying the number of minority classes. Specifically, for classes $i\in\{1..N^\mathcal{*}\}$:
     \begin{equation}
         K^\mathcal{*}_{i}=
         \begin{cases}
             K^\mathcal{*}_{min},           &    \text{if}~ i \leq M^\mathcal{*},\\
             K^\mathcal{*}_{max},           &    \text{otherwise}.
         \end{cases}
    \end{equation}
     For example, in a $step$ (1,9,5,1) set, there is 1 minority class, and $K^\mathcal{*}_{i} \in \{1, 9, 9, 9, 9\}$. 
     \item \emph{Random imbalance}. The number of class samples, $K^\mathcal{*}_{i}$, is sampled from a uniform distribution, i.e. $ K^\mathcal{*}_{i} \sim \mathtt{Unif}(K^\mathcal{*}_{min}, K^\mathcal{*}_{max})$, with $K^\mathcal{*}_{min}$ and $K^\mathcal{*}_{max}$ inclusive. 
     \item \emph{Long-Tail imbalance}. Imbalance could be modeled by a Zipf's/Power Law \cite{Reed2001powerlaw} for a more realistic imbalance distribution.
 \end{itemize}
  We also report the imbalance ratio $\rho$, which is a scalar identifying the level of class imbalance; this is often reported in the CI literature for the supervised case \cite{Buda2018imbalance}. We define $\rho$ to be the ratio between the number of samples in the majority and minority classes in a set of data points:
 \begin{equation}
  \rho=\frac{K^\mathcal{*}_{max}}{K^\mathcal{*}_{min}}.
 \end{equation}
 
\subsection{Methods and Task-Sampling Procedures}\label{sec solutions}
In this section, we provide a panoramic overview over rebalancing methods and task-sampling procedures, starting from one that has been popularly used in CIFSL (Random-Shot Meta-Training) and then describing the ones used in supervised learning (Random Over-Sampling and Rebalancing Losses). We will show how the latter can be adapted to the CISFL setting, and in Section~\ref{sec exp} we will provide experimental results proving their effectiveness. We have selected these particular techniques because they provide a small computational overhead and can be easily adjusted to work with few-shot learning methods, making their adaption straightforward in real-world applications.

\textbf{Random-Shot Meta-Training.} Offers a realistic task-sampling procedure during meta-training where the task-distribution cannot be controlled easily (e.g.\cite{Finn2019online,Chen2018federated}). Random-Shot exposes a model to imbalanced tasks during the meta-training phase, so it could also be used purposefully to prepare models for imbalanced evaluation tasks. In Section~\ref{sec exp}, we show this setting is overall more challenging when tested across a large variety of methods and conditions. We apply Random-Shot Meta-Training similarly to the standard episodic meta-training \cite{Vinyals2017matching} but with the balanced tasks exchanged with $K_{min}$-$K_{max}$-shot \emph{random}-distribution tasks, as defined above in Section~\ref{sec methodology}. To isolate imbalance, we keep the number of classes between tasks the same (i.e., $N{^S} = 5$ for all meta-phases). We use random-distribution following previous work \cite{Triantafillou2019meta,Lee2020baysiantaml}, since, in real-world applications, the actual imbalance distribution at the few-shot scale is likely to be unknown.

\textbf{Random Over-Sampling.} A strategy used in standard supervised learning to rebalance classes is
Random Over-Sampling (ROS) \cite{Japkowicz2002imbalance}. ROS randomly samples data points from the minority classes, automatically rebalancing them. Adapting ROS to the CIFSL setting consists of matching the number of support samples in the non-majority classes to the number of support samples in the majority class, $K_{i} = \max_i(K_i)$. This means that for $\mathcal{I} \in \{linear, step\}$, the number of samples in each class is equal to $K_{max}$. In particular, we match $K_i$ to $\max_i(K_i)$ by resampling uniformly at random the remaining $\max_i(K_i)-K_{i}$ support samples belonging to class $i$, and then appending them to the support set. When applying ROS with augmentation (\emph{ROS+}), we perform further data transformation on the resampled supports. A visual representation of a class imbalanced task after applying ROS and ROS+ is presented in Appendix~\ref{appendix implementation} (Figure~\ref{fig ros_support}).

\textbf{Rebalancing Losses.} In supervised learning, various loss functions have been introduced to tackle class imbalance. Here, we will focus on three of them: Weighted Loss, Focal Loss, and Class Balancing Loss. The \emph{Weighted Loss} \cite{Buda2018imbalance,Leevy2018bigdata} defines a set of weights for each class, giving more or less weight based on the number of samples associated to the class. We adapted the Weighted Loss to the CIFSL setting by weighting the inner-loop cross-entropy loss of optimization-based algorithms and fine-tune baselines by inverse class frequency of support set samples. Another popular rebalancing loss is the \emph{Focal Loss} \cite{Lin2017focal}, which exploits the confidence of the model to identify hard-to-classify data points that deserve more attention and create greater influence on the backpropagation of the gradients. Focal Loss effectively combats class imbalance in one-stage object detectors but has not been used in the CIFSL setting. In our adaptation, we exchanged the inner-loop cross-entropy loss of optimization-based algorithms and fine-tune baselines with the Focal Loss with $\gamma=2$ and $\alpha=1$. Finally, the \emph{Class Balancing Loss (CB Loss)} is a rebalancing loss multiplier commonly used in the long-tail literature \cite{Cui2019cbloss}. The loss associated with each class is re-weighted based on the effective number of samples for each class. In our adaptation of the CB Loss for CIFSL, we weighted the inner-loop cross-entropy loss of optimization-based algorithms and fine-tune baselines using the class multiplier. The multiplier is calculated as $(1-\beta^n)/(1-\beta)$ where $n$ is the number of samples per class, and $\beta=0.8$ based on $N-1/N$ formula, where $N$ is the number of classes. We also tried different values for $\beta$, which are discussed in Appendix~\ref{appendix implementation}.

\begin{figure*}[tbh]
    \centering
    \includegraphics[width=\linewidth]{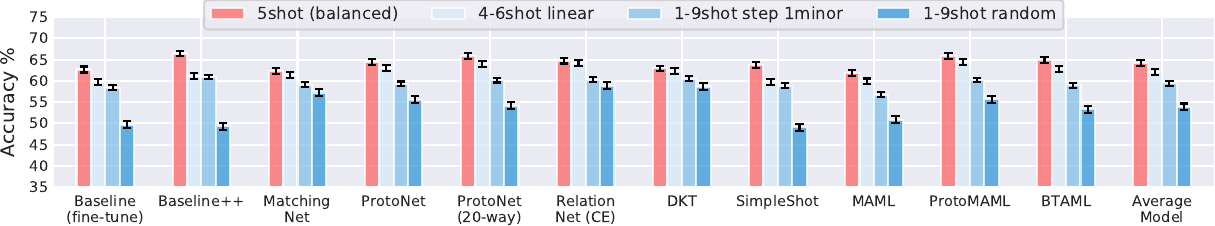}
    \caption{Overview of the Class Imbalance Few-Shot Learning (CIFSL) problem in Standard Meta-Training. Performance of models meta-trained with standard (balanced) tasks, meta-evaluated on balanced and unbalanced distributions. All models show a consistent performance drop when evaluated on imbalance distributions (blue bars) with respect to evaluation on standard balanced tasks (red bar). The task with 1-9shot step imbalance with 1 minority class contains 37 support samples in total, while all other tasks contain only 25 support samples. Most methods perform worse on the imbalanced tasks, despite having the same or higher number of support samples.}
    \label{fig balanced_vs_imbalanced}
\end{figure*}

\section{Experiments}\label{sec exp}

In this section, we provide experimental evidence for our claims. In particular, we show the following:
(i) evidence that the CIFSL is indeed a significant problem -- all methods underperform when meta-tested on imbalanced tasks; (ii) contrary to popular belief, Random-Shot Meta-Training has a serious disadvantage on imbalanced evaluation tasks, when tested on a large number of conditions and methods; (iii) techniques used in supervised learning to combat imbalance (e.g. ROS and Rebalancing Losses) can be adapted to CIFSL with state-of-the-art performances; (iv) dataset imbalance has can have a small negative effect on performance which is not as significant as task-level imbalance; and (v) combining dataset and task imbalance results has compounding effect on the degradation in performance. In the following subsections we outline the experimental setup in Section~\ref{sec exp setup}, then we present results showing the model robustness to each of the imbalance levels: \emph{task-level} in Section~\ref{sec imb task}, \emph{dataset-level} in Section~\ref{sec imb dataset} and \emph{combined} in Section~\ref{sec imb combined}.

\subsection{Setup}\label{sec exp setup}
We adapted 10 unique baselines and FSL methods into our framework: Fine-tune baseline \cite{Pan2010transfer}, Baseline++ \cite{Chen2019closer}, SimpleShot \cite{Wang2019simpleshot}, Prototypical Networks \cite{Snell2017proto}, Matching Networks \cite{Vinyals2017matching}, Relation Networks \cite{Sung2017relationnet}, MAML \cite{Finn2017maml}, ProtoMAML \cite{Triantafillou2019meta}, DKT \cite{Patacchiola2019gpshot}, and Bayesian TAML (BTAML) \cite{Lee2020baysiantaml}. We used a 4 layer convolutional network as backbone for each model, following common practice \cite{Chen2019closer}. All experiments were repeated three times with different initialization seeds. Each data point represents the average performance of 600 meta-testing tasks per run. Error bars in figures and numbers in brackets in tables show 95\% confidence interval, unless stated otherwise. 

For \emph{task-level} imbalance, we trained FSL models using \emph{Standard} (episodic) meta-training \cite{Vinyals2017matching} using 5-shot 5-way tasks, as well as \emph{Random-Shot} meta-training \cite{Triantafillou2019meta,Lee2020baysiantaml,Chen2020mamlstop} using 1-9shot 5-way random-distribution tasks (as described in Section~\ref{sec methodology}).  We trained and evaluated all methods on MiniImageNet \cite{Ravi2017,Vinyals2017matching}, containing 64 classes with 600 images each. For most experiments, we focused on the low-data range with a support set of 25 samples (5 avr. shot), but we also explored mid-shot settings (see in Section~\ref{sec exp ablation} for more details). We applied the mitigation techniques during meta-training or meta-evaluation. We pre-trained baselines (i.e., Fine-Tune, 1-NN, Baseline++) using mini-batch gradient descent and then fine-tuned on the support or perform a 1-NN classification. We evaluated all baselines and models using a wide range of imbalanced meta-testing tasks. To isolate the effects of support-set imbalance, we kept the query set balanced. For clarity throughout this report, we distinguish between the meta-training and meta-evaluation phases using the right arrow ($\rightarrow$). For example, \emph{``Standard $\rightarrow$ ROS''} indicates Standard meta-training with ROS applied at meta-test time. Similarly, \emph{``Random-Shot \& ROS $\rightarrow$ ROS+"} indicates the use of ROS during Randoms-Shot Meta-Training and the use of ROS+ at meta-test time. 

For \emph{dataset-level} imbalance, we halved the total number of samples from the original $\mathcal{D}_{train}$ of Mini-ImageNet to get enough samples for the majority classes. We denote this dataset as $\mathcal{D}'_{train}$, where $|\mathcal{D}'_{train}| \approx 64 \times 300 = 19200$. We induced imbalance into the dataset according to one of the $\mathcal{I}$-distributions outlined in Section~\ref{sec methodology}. For simplicity, we did not modify the meta-validation and meta-testing datasets ($\mathcal{D}_{val}$ and $\mathcal{D}_{test}$), keeping them like in the original Min-ImageNet. To emulate stronger domain-shift, we also evaluated models on 50 randomly selected classes from CUB-2011 \cite{Wah2011cub}.

For the \emph{combined-level} imbalance setting, we trained models using linear 1-9shot tasks with linearly imbalanced dataset $\mathcal{D}'_{train}$. More details and experiments can be found in Appendix~\ref{appendix implementation}.
 
\subsection{Robustness to Task Imbalance}\label{sec imb task}

\subsubsection{Standard Meta-Training}\label{sec exp standard}
In the first set of experiments, we highlight the magnitude of the CIFSL problem with Standard meta-training. We meta-trained all the models using the standard (balanced) episodic meta-training schedule (5-shot 5-way), and then we meta-evaluated them on both balanced and imbalanced tasks under various imbalance distributions. The results are reported in Figure~\ref{fig balanced_vs_imbalanced} and demonstrate the crux of CIFSL problem. Specifically, the figure shows that even a small level of imbalance (linear 4-6-shot 5-way, $\rho=1.5$) produces a significant\footnote{Non-overlapping 95\% confidence intervals indicate `significant' performance difference.} performance drop for 7 out of 11 models. The average accuracy drop is $-1.5\%$ for metric-based models and $-8.2\%$ for optimization-based models. On tasks with a larger imbalance (1-9shot random, $\rho=9.0$) the effect is even more evident, with a drop of $-8.4\%$ for metric-based models and $-17.1\%$ for optimization-based models. Optimization-based methods and fine-tune baselines suffer more as they use conventional supervised loss functions in the inner-loop which are known to be particularly susceptible to imbalance \cite{Buda2018imbalance,Johnson2019imbalance,Japkowicz2002imbalance}. Interestingly, despite the additional 12 samples in the support set in 1-9shot step tasks with 1 minority class ($\rho=9.0$), the average model performance drops by $-5.0\%$ to the balanced task with 25 samples. Note that, although all methods are susceptible to class imbalance, some are more robust. Matching Net, Relation Net, and DKT perform better than the others with an average drop of only $-5\%$ on the task with largest imbalance. Training on balanced tasks, renders evaluation on imbalanced tasks out-of-distribution \cite{Lee2020baysiantaml}. In section~\ref{sec exp random-shot}, we explore another meta-training procedure, Random-Shot, that exposes models to imbalanced tasks before testing. 

\begin{table}[tbh]
    \centering
    \caption{Table showing the average model accuracy after Standard meta-/pre-training, and meta-testing on balanced and imbalanced tasks while applying ROS/ROS+ at the meta-testing time only.}
    \scalebox{0.89}{\begin{tabular}{cr|cccc}
\toprule
           \specialcell{Model}          &                \specialcell{Train$\rightarrow$Test}                   &          \specialcell{5shot\\balanced} &                     \specialcell{4-6shot\\linear} &                       \specialcell{1-9shot\\step} &                     \specialcell{1-9shot\\random} \\
\midrule
\multirow{3}{*}{\specialcell{Baseline\\(fine-tune)}} & \specialcell{Standard} &  62.67\textcolor{gray}{\tiny{$\pm$0.70}} &           59.78\textcolor{gray}{\tiny{$\pm$0.70}} &           58.39\textcolor{gray}{\tiny{$\pm$0.57}} &           49.72\textcolor{gray}{\tiny{$\pm$0.85}} \\
                    & \specialcell{$\rightarrow$ ROS} &                               \textbf{-} &           59.42\textcolor{gray}{\tiny{$\pm$0.69}} &           57.80\textcolor{gray}{\tiny{$\pm$0.58}} &           51.11\textcolor{gray}{\tiny{$\pm$0.81}} \\
                    & \specialcell{$\rightarrow$ ROS+} &                                        - &  \textbf{61.07\textcolor{gray}{\tiny{$\pm$0.70}}} &  \textbf{61.11\textcolor{gray}{\tiny{$\pm$0.64}}} &  \textbf{57.99\textcolor{gray}{\tiny{$\pm$0.74}}} \\
\midrule
\multirow{3}{*}{\specialcell{Baseline++}} & \specialcell{Standard} &  66.43\textcolor{gray}{\tiny{$\pm$0.66}} &           61.17\textcolor{gray}{\tiny{$\pm$0.69}} &           60.93\textcolor{gray}{\tiny{$\pm$0.54}} &           49.28\textcolor{gray}{\tiny{$\pm$0.89}} \\
                    & \specialcell{$\rightarrow$ ROS} &                               \textbf{-} &           64.43\textcolor{gray}{\tiny{$\pm$0.67}} &           61.52\textcolor{gray}{\tiny{$\pm$0.55}} &           58.76\textcolor{gray}{\tiny{$\pm$0.76}} \\
                    & \specialcell{$\rightarrow$ ROS+} &                                        - &  \textbf{64.97\textcolor{gray}{\tiny{$\pm$0.66}}} &  \textbf{65.06\textcolor{gray}{\tiny{$\pm$0.63}}} &  \textbf{62.87\textcolor{gray}{\tiny{$\pm$0.69}}} \\
\midrule
\multirow{3}{*}{\specialcell{Matching\\Net}} & \specialcell{Standard} &  62.27\textcolor{gray}{\tiny{$\pm$0.69}} &  \textbf{61.41\textcolor{gray}{\tiny{$\pm$0.68}}} &           59.09\textcolor{gray}{\tiny{$\pm$0.60}} &           57.26\textcolor{gray}{\tiny{$\pm$0.73}} \\
                    & \specialcell{$\rightarrow$ ROS} &                               \textbf{-} &           60.10\textcolor{gray}{\tiny{$\pm$0.66}} &           60.41\textcolor{gray}{\tiny{$\pm$0.66}} &           58.21\textcolor{gray}{\tiny{$\pm$0.68}} \\
                    & \specialcell{$\rightarrow$ ROS+} &                                        - &           60.17\textcolor{gray}{\tiny{$\pm$0.66}} &  \textbf{60.70\textcolor{gray}{\tiny{$\pm$0.65}}} &  \textbf{58.57\textcolor{gray}{\tiny{$\pm$0.67}}} \\
\midrule
\multirow{3}{*}{\specialcell{ProtoNet}} & \specialcell{Standard} &  64.37\textcolor{gray}{\tiny{$\pm$0.71}} &  \textbf{63.10\textcolor{gray}{\tiny{$\pm$0.70}}} &           59.28\textcolor{gray}{\tiny{$\pm$0.57}} &           55.54\textcolor{gray}{\tiny{$\pm$0.82}} \\
                    & \specialcell{$\rightarrow$ ROS} &                               \textbf{-} &           60.06\textcolor{gray}{\tiny{$\pm$0.68}} &           58.30\textcolor{gray}{\tiny{$\pm$0.56}} &           52.37\textcolor{gray}{\tiny{$\pm$0.79}} \\
                    & \specialcell{$\rightarrow$ ROS+} &                                        - &           62.30\textcolor{gray}{\tiny{$\pm$0.69}} &  \textbf{61.38\textcolor{gray}{\tiny{$\pm$0.62}}} &  \textbf{59.08\textcolor{gray}{\tiny{$\pm$0.72}}} \\
\midrule
\multirow{3}{*}{\specialcell{ProtoNet\\(20-way)}} & \specialcell{Standard} &  65.76\textcolor{gray}{\tiny{$\pm$0.70}} &  \textbf{63.92\textcolor{gray}{\tiny{$\pm$0.71}}} &           60.18\textcolor{gray}{\tiny{$\pm$0.56}} &           54.14\textcolor{gray}{\tiny{$\pm$0.87}} \\
                    & \specialcell{$\rightarrow$ ROS} &                               \textbf{-} &           59.75\textcolor{gray}{\tiny{$\pm$0.69}} &           59.04\textcolor{gray}{\tiny{$\pm$0.55}} &           50.90\textcolor{gray}{\tiny{$\pm$0.83}} \\
                    & \specialcell{$\rightarrow$ ROS+} &                                        - &           63.52\textcolor{gray}{\tiny{$\pm$0.70}} &  \textbf{62.12\textcolor{gray}{\tiny{$\pm$0.61}}} &  \textbf{59.79\textcolor{gray}{\tiny{$\pm$0.75}}} \\
\midrule
\multirow{3}{*}{\specialcell{Relation\\Net (CE)}} & \specialcell{Standard} &  64.76\textcolor{gray}{\tiny{$\pm$0.68}} &  \textbf{64.14\textcolor{gray}{\tiny{$\pm$0.69}}} &           60.28\textcolor{gray}{\tiny{$\pm$0.58}} &           58.86\textcolor{gray}{\tiny{$\pm$0.78}} \\
                    & \specialcell{$\rightarrow$ ROS} &                               \textbf{-} &           61.85\textcolor{gray}{\tiny{$\pm$0.68}} &           59.07\textcolor{gray}{\tiny{$\pm$0.57}} &           56.08\textcolor{gray}{\tiny{$\pm$0.75}} \\
                    & \specialcell{$\rightarrow$ ROS+} &                                        - &           63.14\textcolor{gray}{\tiny{$\pm$0.69}} &  \textbf{62.51\textcolor{gray}{\tiny{$\pm$0.65}}} &  \textbf{60.50\textcolor{gray}{\tiny{$\pm$0.71}}} \\
\midrule
\multirow{3}{*}{\specialcell{DKT}} & \specialcell{Standard} &  62.92\textcolor{gray}{\tiny{$\pm$0.67}} &  \textbf{62.32\textcolor{gray}{\tiny{$\pm$0.67}}} &           60.54\textcolor{gray}{\tiny{$\pm$0.63}} &           58.66\textcolor{gray}{\tiny{$\pm$0.73}} \\
                    & \specialcell{$\rightarrow$ ROS} &                               \textbf{-} &           61.56\textcolor{gray}{\tiny{$\pm$0.66}} &           60.11\textcolor{gray}{\tiny{$\pm$0.61}} &           58.47\textcolor{gray}{\tiny{$\pm$0.69}} \\
                    & \specialcell{$\rightarrow$ ROS+} &                                        - &           61.54\textcolor{gray}{\tiny{$\pm$0.67}} &  \textbf{61.62\textcolor{gray}{\tiny{$\pm$0.65}}} &  \textbf{59.43\textcolor{gray}{\tiny{$\pm$0.69}}} \\
\midrule
\multirow{3}{*}{\specialcell{Simple-\\Shot}} & \specialcell{Standard} &  63.74\textcolor{gray}{\tiny{$\pm$0.69}} &           59.77\textcolor{gray}{\tiny{$\pm$0.70}} &           58.89\textcolor{gray}{\tiny{$\pm$0.56}} &           49.05\textcolor{gray}{\tiny{$\pm$0.89}} \\
                    & \specialcell{$\rightarrow$ ROS} &                               \textbf{-} &           55.15\textcolor{gray}{\tiny{$\pm$0.71}} &           58.12\textcolor{gray}{\tiny{$\pm$0.58}} &           45.36\textcolor{gray}{\tiny{$\pm$0.85}} \\
                    & \specialcell{$\rightarrow$ ROS+} &                                        - &  \textbf{61.14\textcolor{gray}{\tiny{$\pm$0.72}}} &  \textbf{59.63\textcolor{gray}{\tiny{$\pm$0.61}}} &  \textbf{55.57\textcolor{gray}{\tiny{$\pm$0.77}}} \\
\midrule
\multirow{3}{*}{\specialcell{MAML}} & \specialcell{Standard} &  61.83\textcolor{gray}{\tiny{$\pm$0.71}} &           59.87\textcolor{gray}{\tiny{$\pm$0.69}} &           56.77\textcolor{gray}{\tiny{$\pm$0.59}} &           50.83\textcolor{gray}{\tiny{$\pm$0.80}} \\
                    & \specialcell{$\rightarrow$ ROS} &                               \textbf{-} &           59.94\textcolor{gray}{\tiny{$\pm$0.70}} &           57.93\textcolor{gray}{\tiny{$\pm$0.63}} &           54.96\textcolor{gray}{\tiny{$\pm$0.76}} \\
                    & \specialcell{$\rightarrow$ ROS+} &                                        - &  \textbf{60.17\textcolor{gray}{\tiny{$\pm$0.72}}} &  \textbf{60.30\textcolor{gray}{\tiny{$\pm$0.68}}} &  \textbf{58.15\textcolor{gray}{\tiny{$\pm$0.75}}} \\
\midrule
\multirow{3}{*}{\specialcell{Proto-\\MAML}} & \specialcell{Standard} &  65.87\textcolor{gray}{\tiny{$\pm$0.71}} &           64.36\textcolor{gray}{\tiny{$\pm$0.68}} &           60.19\textcolor{gray}{\tiny{$\pm$0.57}} &           55.66\textcolor{gray}{\tiny{$\pm$0.80}} \\
                    & \specialcell{$\rightarrow$ ROS} &                               \textbf{-} &           63.25\textcolor{gray}{\tiny{$\pm$0.70}} &           60.72\textcolor{gray}{\tiny{$\pm$0.57}} &           56.32\textcolor{gray}{\tiny{$\pm$0.78}} \\
                    & \specialcell{$\rightarrow$ ROS+} &                                        - &  \textbf{65.14\textcolor{gray}{\tiny{$\pm$0.70}}} &  \textbf{64.50\textcolor{gray}{\tiny{$\pm$0.65}}} &  \textbf{62.20\textcolor{gray}{\tiny{$\pm$0.74}}} \\
\midrule
\multirow{3}{*}{\specialcell{BTAML}} & \specialcell{Standard} &  64.96\textcolor{gray}{\tiny{$\pm$0.70}} &           62.88\textcolor{gray}{\tiny{$\pm$0.69}} &           58.96\textcolor{gray}{\tiny{$\pm$0.57}} &           53.32\textcolor{gray}{\tiny{$\pm$0.81}} \\
                    & \specialcell{$\rightarrow$ ROS} &                               \textbf{-} &           64.04\textcolor{gray}{\tiny{$\pm$0.71}} &           62.89\textcolor{gray}{\tiny{$\pm$0.66}} &           59.59\textcolor{gray}{\tiny{$\pm$0.76}} \\
                    & \specialcell{$\rightarrow$ ROS+} &                                        - &  \textbf{64.28\textcolor{gray}{\tiny{$\pm$0.72}}} &  \textbf{63.93\textcolor{gray}{\tiny{$\pm$0.68}}} &  \textbf{61.49\textcolor{gray}{\tiny{$\pm$0.75}}} \\
\bottomrule
\end{tabular}}
    \label{tbl standard_vs_imbalanced_tasks}
\end{table}

\textbf{Mitigating Class Imbalance.} The poor performance of algorithms on imbalanced evaluation tasks perhaps is not too surprising. After all, the models were never exposed to imbalance during meta-training, and these models do not necessarily have mechanisms to deal with imbalance per se. Table~\ref{tbl standard_vs_imbalanced_tasks} shows that performance on imbalanced evaluation tasks can be effectively improved using simple approaches, such as random-oversampling with or without augmentation \cite{Japkowicz2002imbalance}, applied during the meta-testing time of meta-/pre-trained models. Overall, we see that for higher levels of imbalance (step and random), ROS+ effectively improves performances for all models. In the lower levels of imbalance (linear), the ROS+ has at least as good performance as standard inference for 7 out of 11 models. For the other 4 models, the performance is around 1\% lower but still just within overlapping confidence intervals. The models that do not see an improvement are the three classical metric-learning based methods: ProtoNet, Matching Net, and Relation Net. These metric-based models might be unable to adjust to feature shifts caused by augmentation in ROS+, which could indicate sub-optimal transformations performed by ROS+, skewing the feature representation away from what the models have learned during meta-training. Future work could investigate different augmentations that would affect performance - some work already studies augmentations in the general FSL setting \cite{Rajendran2020meta,Ni2020data,Antoniou2019unsupervised,Antoniou2017gan_augment}. Despite this drawback, augmenting resampled images has a clear advantage over the plain ROS in many settings. ROS (without augmentation) performs even worse than Standard for ProtoNet, Relation Net, and SimpleShot, showing a clear advantage of augmenting resampled features. 

\textbf{Precision and Recall.} We dive deeper into the effect caused by task-level imbalance imposed on individual classes. The top two sub-tables in Table~\ref{tbl pr_standard} show the average precision and recall for each class in a linear 1-9shot 5-way task after Standard meta-training with no mitigation strategy. The lower recall and precision scores for the minority classes ($K_i \leq 3$) reflect the accuracy performance drops observed earlier. Interestingly, although the precision remains low for many algorithms, the precision remains relatively high for mid-shot classes ($K_i\in\{3,5\}$) and before dropping on the high-shot classes ($K_i\in\{7,9\}$). We posit that this precision drop is caused by the mislabeling of low-shot classes as a high-shot class, resulting in poorer precision in those majority classes. Some methods (i.e., DKT, Matching Net, and Relation Net) perform better compared to other methods. We posit that Matching Net's ability to look at data-points individually and independently means the model is agnostic to the support set size, resulting in higher precision and recall. In the case of DKT, which uses Gaussian Processes, the algorithm is able to naturally account for imbalance by adjusting its confidence in prediction based on the number of samples per class, thus, able to effectively adjust its prediction scores based on the number and quality of shots, giving DKT the best precision and recall for the minority class. Relation Net shares part of its mechanism for calculating class mean embeddings with ProtoNet, and yet, the additional relation module yields significantly higher robustness against imbalance. The relation module allows for non-linear separability between classes in embedding space \cite{Sung2017relationnet}, and therefore, it is less reliant on getting a good prototype representation. Intuitively, in ProtoNet, a single sample is unlikely to be a good representation of a class prototype, and using Euclidean distance to compare between query and prototype vectors forces linear class boundaries, which may be inadequate for low number of shots. In contrast, the relation module can pay more attention to the presence and lack of specific features within query-prototype embeddings, effectively defining non-linear class boundaries. Our results show that Relation Net has a higher recall rate for the minority classes than ProtoNet, which is consistent with our hypothesis. Additional precision-recall tables for Standard $\rightarrow$ ROS+ are in the Appendix~\ref{sec pr_tables}.

\begin{table}[tbh]
    \centering
    \caption{Table showing the average precision (top) and recall (bottom) for each class of a linear 1-9shot evaluation task after Standard training. The number of support samples for each class in denoted by $K_i$.}
    \scalebox{0.82}{

\begin{tabular}{cl|ccccc}
\toprule
               &       &                                          $K_0=1$ &                                          $K_1=3$ &                                          $K_2=5$ &                                          $K_3=7$ &                                          $K_4=9$ \\
\midrule
\parbox[t]{1mm}{\multirow{11}{*}{\rotatebox[origin=c]{90}{\specialcell{Precision (95\%CI)}}}} & Baseline (fine-tune) &           0.05\textcolor{gray}{\tiny{$\pm$0.02}} &           0.63\textcolor{gray}{\tiny{$\pm$0.03}} &           0.63\textcolor{gray}{\tiny{$\pm$0.02}} &           0.50\textcolor{gray}{\tiny{$\pm$0.01}} &           0.41\textcolor{gray}{\tiny{$\pm$0.01}} \\
               & Baseline++ &           0.02\textcolor{gray}{\tiny{$\pm$0.01}} &           0.45\textcolor{gray}{\tiny{$\pm$0.04}} &  \textbf{0.66\textcolor{gray}{\tiny{$\pm$0.02}}} &           0.52\textcolor{gray}{\tiny{$\pm$0.01}} &           0.39\textcolor{gray}{\tiny{$\pm$0.01}} \\
               & Matching Net &           0.54\textcolor{gray}{\tiny{$\pm$0.04}} &           0.67\textcolor{gray}{\tiny{$\pm$0.02}} &           0.62\textcolor{gray}{\tiny{$\pm$0.01}} &           0.55\textcolor{gray}{\tiny{$\pm$0.01}} &           0.48\textcolor{gray}{\tiny{$\pm$0.01}} \\
               & ProtoNet &           0.18\textcolor{gray}{\tiny{$\pm$0.03}} &           0.70\textcolor{gray}{\tiny{$\pm$0.02}} &           0.62\textcolor{gray}{\tiny{$\pm$0.02}} &           0.54\textcolor{gray}{\tiny{$\pm$0.01}} &           0.51\textcolor{gray}{\tiny{$\pm$0.01}} \\
               & ProtoNet (20-way) &           0.10\textcolor{gray}{\tiny{$\pm$0.02}} &  \textbf{0.71\textcolor{gray}{\tiny{$\pm$0.03}}} &           0.64\textcolor{gray}{\tiny{$\pm$0.02}} &           0.55\textcolor{gray}{\tiny{$\pm$0.01}} &           0.49\textcolor{gray}{\tiny{$\pm$0.01}} \\
               & Relation Net (CE) &           0.42\textcolor{gray}{\tiny{$\pm$0.04}} &           0.67\textcolor{gray}{\tiny{$\pm$0.02}} &           0.62\textcolor{gray}{\tiny{$\pm$0.01}} &           0.56\textcolor{gray}{\tiny{$\pm$0.01}} &           0.54\textcolor{gray}{\tiny{$\pm$0.01}} \\
               & DKT &  \textbf{0.57\textcolor{gray}{\tiny{$\pm$0.03}}} &           0.63\textcolor{gray}{\tiny{$\pm$0.02}} &           0.61\textcolor{gray}{\tiny{$\pm$0.01}} &  \textbf{0.57\textcolor{gray}{\tiny{$\pm$0.01}}} &  \textbf{0.55\textcolor{gray}{\tiny{$\pm$0.01}}} \\
               & SimpleShot &           0.01\textcolor{gray}{\tiny{$\pm$0.01}} &           0.59\textcolor{gray}{\tiny{$\pm$0.04}} &           0.66\textcolor{gray}{\tiny{$\pm$0.02}} &           0.51\textcolor{gray}{\tiny{$\pm$0.01}} &           0.41\textcolor{gray}{\tiny{$\pm$0.01}} \\
               & MAML &           0.00\textcolor{gray}{\tiny{$\pm$0.00}} &           0.65\textcolor{gray}{\tiny{$\pm$0.02}} &           0.58\textcolor{gray}{\tiny{$\pm$0.02}} &           0.48\textcolor{gray}{\tiny{$\pm$0.01}} &           0.42\textcolor{gray}{\tiny{$\pm$0.01}} \\
               & ProtoMAML &           0.28\textcolor{gray}{\tiny{$\pm$0.04}} &           0.71\textcolor{gray}{\tiny{$\pm$0.02}} &           0.63\textcolor{gray}{\tiny{$\pm$0.01}} &           0.53\textcolor{gray}{\tiny{$\pm$0.01}} &           0.45\textcolor{gray}{\tiny{$\pm$0.01}} \\
               & BTAML &           0.11\textcolor{gray}{\tiny{$\pm$0.03}} &           0.69\textcolor{gray}{\tiny{$\pm$0.03}} &           0.61\textcolor{gray}{\tiny{$\pm$0.02}} &           0.51\textcolor{gray}{\tiny{$\pm$0.01}} &           0.43\textcolor{gray}{\tiny{$\pm$0.01}} \\
\midrule
\parbox[t]{1mm}{\multirow{11}{*}{\rotatebox[origin=c]{90}{\specialcell{Recall (95\%CI)}}}}  & Baseline (fine-tune) &           0.00\textcolor{gray}{\tiny{$\pm$0.00}} &           0.19\textcolor{gray}{\tiny{$\pm$0.02}} &           0.53\textcolor{gray}{\tiny{$\pm$0.02}} &           0.75\textcolor{gray}{\tiny{$\pm$0.01}} &           0.90\textcolor{gray}{\tiny{$\pm$0.01}} \\
               & Baseline++ &           0.00\textcolor{gray}{\tiny{$\pm$0.00}} &           0.13\textcolor{gray}{\tiny{$\pm$0.02}} &           0.48\textcolor{gray}{\tiny{$\pm$0.02}} &           0.76\textcolor{gray}{\tiny{$\pm$0.02}} &  \textbf{0.94\textcolor{gray}{\tiny{$\pm$0.01}}} \\
               & Matching Net &           0.15\textcolor{gray}{\tiny{$\pm$0.02}} &           0.41\textcolor{gray}{\tiny{$\pm$0.02}} &           0.61\textcolor{gray}{\tiny{$\pm$0.02}} &           0.73\textcolor{gray}{\tiny{$\pm$0.02}} &           0.84\textcolor{gray}{\tiny{$\pm$0.01}} \\
               & ProtoNet &           0.03\textcolor{gray}{\tiny{$\pm$0.01}} &           0.37\textcolor{gray}{\tiny{$\pm$0.02}} &           0.65\textcolor{gray}{\tiny{$\pm$0.02}} &           0.78\textcolor{gray}{\tiny{$\pm$0.01}} &           0.84\textcolor{gray}{\tiny{$\pm$0.01}} \\
               & ProtoNet (20-way) &           0.01\textcolor{gray}{\tiny{$\pm$0.00}} &           0.31\textcolor{gray}{\tiny{$\pm$0.02}} &           0.63\textcolor{gray}{\tiny{$\pm$0.02}} &  \textbf{0.79\textcolor{gray}{\tiny{$\pm$0.01}}} &           0.88\textcolor{gray}{\tiny{$\pm$0.01}} \\
               & Relation Net (CE) &           0.09\textcolor{gray}{\tiny{$\pm$0.01}} &  \textbf{0.48\textcolor{gray}{\tiny{$\pm$0.02}}} &  \textbf{0.68\textcolor{gray}{\tiny{$\pm$0.02}}} &           0.76\textcolor{gray}{\tiny{$\pm$0.01}} &           0.81\textcolor{gray}{\tiny{$\pm$0.01}} \\
               & DKT &  \textbf{0.22\textcolor{gray}{\tiny{$\pm$0.02}}} &           0.48\textcolor{gray}{\tiny{$\pm$0.02}} &           0.65\textcolor{gray}{\tiny{$\pm$0.02}} &           0.73\textcolor{gray}{\tiny{$\pm$0.02}} &           0.79\textcolor{gray}{\tiny{$\pm$0.01}} \\
               & SimpleShot &           0.00\textcolor{gray}{\tiny{$\pm$0.00}} &           0.15\textcolor{gray}{\tiny{$\pm$0.02}} &           0.52\textcolor{gray}{\tiny{$\pm$0.02}} &           0.77\textcolor{gray}{\tiny{$\pm$0.01}} &           0.90\textcolor{gray}{\tiny{$\pm$0.01}} \\
               & MAML &           0.00\textcolor{gray}{\tiny{$\pm$0.00}} &           0.26\textcolor{gray}{\tiny{$\pm$0.02}} &           0.55\textcolor{gray}{\tiny{$\pm$0.02}} &           0.75\textcolor{gray}{\tiny{$\pm$0.01}} &           0.85\textcolor{gray}{\tiny{$\pm$0.01}} \\
               & ProtoMAML &           0.04\textcolor{gray}{\tiny{$\pm$0.01}} &           0.34\textcolor{gray}{\tiny{$\pm$0.02}} &           0.62\textcolor{gray}{\tiny{$\pm$0.02}} &           0.78\textcolor{gray}{\tiny{$\pm$0.01}} &           0.87\textcolor{gray}{\tiny{$\pm$0.01}} \\
               & BTAML &           0.01\textcolor{gray}{\tiny{$\pm$0.00}} &           0.29\textcolor{gray}{\tiny{$\pm$0.02}} &           0.58\textcolor{gray}{\tiny{$\pm$0.02}} &           0.77\textcolor{gray}{\tiny{$\pm$0.01}} &           0.88\textcolor{gray}{\tiny{$\pm$0.01}} \\
\bottomrule
\end{tabular}}
    \label{tbl pr_standard}
\end{table}

\begin{figure*}[bth]
    \centering
    \hspace{0.015\linewidth}
    \includegraphics[width=0.96\linewidth]{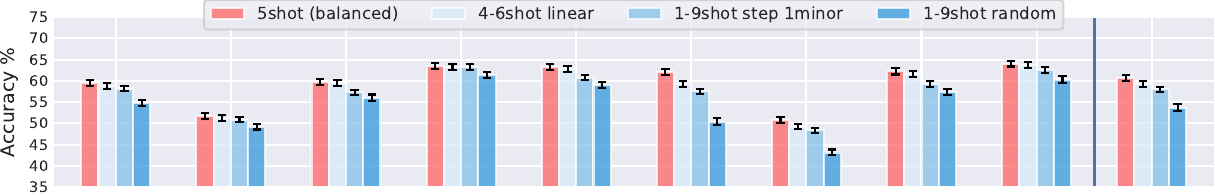}
    \includegraphics[width=0.97\linewidth]{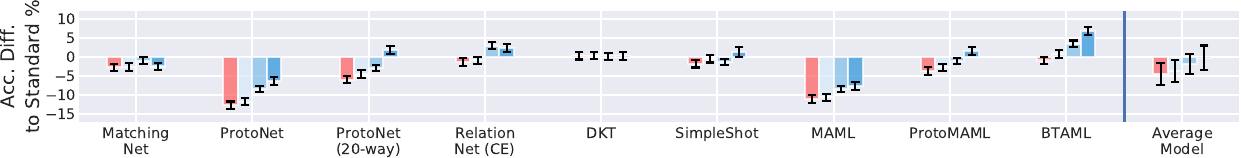}
    \caption{Overview of the Class Imbalance Few-Shot Learning (CIFSL) problem in Random-Shot Meta-Training. \emph{Top:} Performance of models meta-trained with random-shot tasks, meta-evaluated on balanced and unbalanced distributions. All models show a consistent performance drop when evaluated on imbalance distributions (blue bars) respect to evaluation on standard balanced tasks (red bar). \emph{Bottom:} Performance difference between Random-Shot Meta-Training \cite{Triantafillou2019meta} against Standard (balanced) episodic training \cite{Vinyals2017matching} (i.e. results in Figure~\ref{fig balanced_vs_imbalanced}) - a positive score indicates an improvement in Random-Shot. The results show that Random-Shot meta-training offers a harder training setting to most algorithms while only some algorithms have better performance on random-shot evaluation tasks.}
    \label{fig imbalanced_vs_imbalanced}
\end{figure*}

\subsubsection{Random-Shot Meta-Training.}\label{sec exp random-shot}
Meta-training on balanced tasks while evaluating on imbalanced tasks creates a problem of out-of-distribution evaluation tasks \cite{Lee2020baysiantaml}. In the spirit of meta-learning, where the training tasks aim to mimic the evaluation task \cite{Vinyals2017matching}, Random-Shot Meta-Training can be used as a possible way to prepare models for imbalanced at evaluation time \cite{Triantafillou2019meta}. Random-Shot meta-training also offers a more realistic meta-training procedure \cite{Lee2020baysiantaml}, that is more reflective of real-world settings where the training task distribution cannot be easily controlled \cite{Chen2018federated,Massiceti2021orbit}. In this set of experiments, we show that: (i) Random-Shot Meta-Training is a more challenging problem than the Standard (balanced) counterpart; (ii) counter-intuitively, using Random-Shot meta-training does not necessarily help prepare models to imbalanced evaluation tasks and only a limited number of methods benefit from this procedure; (iii) the Random-Shot procedure can be paired with re-balancing procedures during meta-training and meta-testing to alleviate the problem of class imbalance.

\textbf{Does Random-Shot help?} The top of Figure~\ref{fig imbalanced_vs_imbalanced} shows the accuracy on various imbalanced tasks after Random-Shot meta-training. At first glance, the performance drop between 5shot (balanced) tasks and the other (imbalanced) tasks appears less significant than for Standard in Figure~\ref{fig balanced_vs_imbalanced}. However, the bottom of Figure~\ref{fig imbalanced_vs_imbalanced} shows a task-for-task difference in scores between Random-Shot vs Standard, and reveals that Random-Shot performs worse overall compared to Standard. Only 5 out of 9 ML algorithms perform (significantly) better on the 1-9shot random distribution task; however, this comes at the cost of worse performance on other tasks, especially on the balanced 5-shot task. From a practical point of view, Standard meta-training, as a way to prepare models for imbalanced evaluation tasks, offers higher accuracy performance than Random-Shot. A closer look at the task-distribution reveals that a model sees the same type of imbalanced tasks at a rate of 1 in 59k tasks, $(K_{max}-K_{min}+1)^{N} = 9^{5} \approx 59k$. Although this still a low rate, it is still more common than for Standard that never exposes a model to imbalanced tasks during training. In Appendix~\ref{appendix verification}, we report the meta-validation performance through epochs, showing that the models have mostly converged and additional training brings minimal benefit. As a sanity check, we also run a few methods for an increased number of epochs, increasing the number of training tasks five times. However, this did not yield a significant improvement with respect to Standard meta-training (see Appendix~\ref{appendix longer}) and suggests something more fundamental happens during Random-Shot meta-training resulting in poorer generalization performance on other tasks, and further suggests that methods are not suited to out-of-the-box Random-Shot meta-training.

\textbf{Details into Random-Shot trained models.} BTAML is 1 of 2 algorithms that sees improvement by Random-Shot in at least two tasks. Exposure to imbalanced training tasks allows the model to meta-learn the appropriate parameters for its Bayesian Inference Network conditioned on the whole support set and generates task-specific per-class learning rates -- effectively allowing the model to learn a weighing scheme to balance the inner-loop loss. Interestingly, Relation Net trained through Random-Shot also appears to have a slight advantage compared to Standard showing an advantage in the presence of the relation module discussed in Section~\ref{sec exp standard} -- exposure to imbalanced training tasks can help the model focus on specific features and learn a better non-linear class boundary. In contrast to BTAML and Relation Net, the other FSL methods see less stable improvements from Random-Shot. For instance, MAML is known to be a challenging method to train \cite{Antoniou2018mamlpp}. The vanilla batch gradient descent mechanism in the inner-loop may find itself heavily skewed towards majority classes in the support set, which results in larger task variance and renders finding a common network parameter initialization a lot more challenging. Interestingly, DKT performs similarly after Random-Shot and Standard meta-training. This may be due to the partitioned Bayesian one-vs-rest scheme used for classification by DKT, with a separate Gaussian Process for each class; this mechanism could be agnostic to the distribution of class samples within a task. 

\begin{table}[tbh]
    \centering
    \caption{Table showing the average model accuracy after Random-Shot meta-training, and meta-testing on balanced and imbalanced tasks while applying ROS/ROS+ at the meta-testing time only.}
    \scalebox{0.85}{\begin{tabular}{cr|cccc}
\toprule
    \specialcell{Model}          &             \specialcell{Train$\rightarrow$Test}        &                   \specialcell{5shot\\(balanced)} &                     \specialcell{4-6shot\\linear} &                       \specialcell{1-9shot\\step} &                     \specialcell{1-9shot\\random} \\
\midrule
\multirow{4}{*}{\specialcell{Matching\\Net}} & \specialcell{Random-Shot} &           59.58\textcolor{gray}{\tiny{$\pm$0.69}} &           58.79\textcolor{gray}{\tiny{$\pm$0.69}} &           58.19\textcolor{gray}{\tiny{$\pm$0.65}} &           54.79\textcolor{gray}{\tiny{$\pm$0.74}} \\
                    & \specialcell{$\rightarrow$ ROS+} &           58.12\textcolor{gray}{\tiny{$\pm$0.67}} &           58.09\textcolor{gray}{\tiny{$\pm$0.66}} &           58.72\textcolor{gray}{\tiny{$\pm$0.67}} &           56.77\textcolor{gray}{\tiny{$\pm$0.69}} \\
                    & \specialcell{ROS $\rightarrow$ ROS} &  \textbf{61.75\textcolor{gray}{\tiny{$\pm$0.70}}} &  \textbf{61.50\textcolor{gray}{\tiny{$\pm$0.69}}} &  \textbf{61.92\textcolor{gray}{\tiny{$\pm$0.69}}} &  \textbf{59.53\textcolor{gray}{\tiny{$\pm$0.73}}} \\
                    & \specialcell{ROS+ $\rightarrow$ ROS+} &           60.05\textcolor{gray}{\tiny{$\pm$0.68}} &           59.59\textcolor{gray}{\tiny{$\pm$0.67}} &           60.36\textcolor{gray}{\tiny{$\pm$0.67}} &           57.28\textcolor{gray}{\tiny{$\pm$0.70}} \\
\midrule
\multirow{4}{*}{\specialcell{ProtoNet}} & \specialcell{Random-Shot} &           51.65\textcolor{gray}{\tiny{$\pm$0.68}} &           51.31\textcolor{gray}{\tiny{$\pm$0.68}} &           50.89\textcolor{gray}{\tiny{$\pm$0.63}} &           49.24\textcolor{gray}{\tiny{$\pm$0.69}} \\
                    & \specialcell{$\rightarrow$ ROS+} &           50.67\textcolor{gray}{\tiny{$\pm$0.68}} &           50.02\textcolor{gray}{\tiny{$\pm$0.68}} &           50.45\textcolor{gray}{\tiny{$\pm$0.66}} &           48.32\textcolor{gray}{\tiny{$\pm$0.69}} \\
                    & \specialcell{ROS $\rightarrow$ ROS} &           55.35\textcolor{gray}{\tiny{$\pm$0.68}} &           54.40\textcolor{gray}{\tiny{$\pm$0.68}} &           53.76\textcolor{gray}{\tiny{$\pm$0.62}} &           51.32\textcolor{gray}{\tiny{$\pm$0.70}} \\
                    & \specialcell{ROS+ $\rightarrow$ ROS+} &  \textbf{60.05\textcolor{gray}{\tiny{$\pm$0.71}}} &  \textbf{59.98\textcolor{gray}{\tiny{$\pm$0.69}}} &  \textbf{60.57\textcolor{gray}{\tiny{$\pm$0.66}}} &  \textbf{57.35\textcolor{gray}{\tiny{$\pm$0.74}}} \\
\midrule
\multirow{4}{*}{\specialcell{ProtoNet\\(20-way)}} & \specialcell{Random-Shot} &           59.79\textcolor{gray}{\tiny{$\pm$0.70}} &           59.52\textcolor{gray}{\tiny{$\pm$0.71}} &           57.32\textcolor{gray}{\tiny{$\pm$0.62}} &           56.03\textcolor{gray}{\tiny{$\pm$0.76}} \\
                    & \specialcell{$\rightarrow$ ROS+} &           58.31\textcolor{gray}{\tiny{$\pm$0.72}} &           57.76\textcolor{gray}{\tiny{$\pm$0.70}} &           57.59\textcolor{gray}{\tiny{$\pm$0.67}} &           55.55\textcolor{gray}{\tiny{$\pm$0.71}} \\
                    & \specialcell{ROS $\rightarrow$ ROS} &           59.52\textcolor{gray}{\tiny{$\pm$0.71}} &           58.30\textcolor{gray}{\tiny{$\pm$0.72}} &           56.89\textcolor{gray}{\tiny{$\pm$0.63}} &           54.84\textcolor{gray}{\tiny{$\pm$0.74}} \\
                    & \specialcell{ROS+ $\rightarrow$ ROS+} &  \textbf{61.21\textcolor{gray}{\tiny{$\pm$0.72}}} &  \textbf{61.41\textcolor{gray}{\tiny{$\pm$0.71}}} &  \textbf{62.84\textcolor{gray}{\tiny{$\pm$0.66}}} &  \textbf{59.89\textcolor{gray}{\tiny{$\pm$0.75}}} \\
\midrule
\multirow{4}{*}{\specialcell{Relation\\Net (CE)}} & \specialcell{Random-Shot} &           63.50\textcolor{gray}{\tiny{$\pm$0.70}} &           63.30\textcolor{gray}{\tiny{$\pm$0.70}} &           63.33\textcolor{gray}{\tiny{$\pm$0.69}} &  \textbf{61.33\textcolor{gray}{\tiny{$\pm$0.74}}} \\
                    & \specialcell{$\rightarrow$ ROS+} &           62.53\textcolor{gray}{\tiny{$\pm$0.71}} &           61.97\textcolor{gray}{\tiny{$\pm$0.71}} &           62.30\textcolor{gray}{\tiny{$\pm$0.68}} &           59.96\textcolor{gray}{\tiny{$\pm$0.73}} \\
                    & \specialcell{ROS $\rightarrow$ ROS} &  \textbf{64.12\textcolor{gray}{\tiny{$\pm$0.71}}} &  \textbf{63.31\textcolor{gray}{\tiny{$\pm$0.70}}} &  \textbf{64.20\textcolor{gray}{\tiny{$\pm$0.68}}} &           61.32\textcolor{gray}{\tiny{$\pm$0.77}} \\
                    & \specialcell{ROS+ $\rightarrow$ ROS+} &           60.48\textcolor{gray}{\tiny{$\pm$0.71}} &           60.50\textcolor{gray}{\tiny{$\pm$0.70}} &           61.44\textcolor{gray}{\tiny{$\pm$0.69}} &           58.20\textcolor{gray}{\tiny{$\pm$0.73}} \\
\midrule
\multirow{4}{*}{\specialcell{DKT}} & \specialcell{Random-Shot} &  \textbf{63.23\textcolor{gray}{\tiny{$\pm$0.66}}} &           62.72\textcolor{gray}{\tiny{$\pm$0.67}} &           60.75\textcolor{gray}{\tiny{$\pm$0.62}} &           58.92\textcolor{gray}{\tiny{$\pm$0.71}} \\
                    & \specialcell{$\rightarrow$ ROS+} &           62.33\textcolor{gray}{\tiny{$\pm$0.66}} &           62.07\textcolor{gray}{\tiny{$\pm$0.67}} &  \textbf{62.01\textcolor{gray}{\tiny{$\pm$0.64}}} &  \textbf{60.05\textcolor{gray}{\tiny{$\pm$0.69}}} \\
                    & \specialcell{ROS $\rightarrow$ ROS} &           63.21\textcolor{gray}{\tiny{$\pm$0.67}} &  \textbf{62.85\textcolor{gray}{\tiny{$\pm$0.68}}} &           61.52\textcolor{gray}{\tiny{$\pm$0.65}} &           59.80\textcolor{gray}{\tiny{$\pm$0.72}} \\
                    & \specialcell{ROS+ $\rightarrow$ ROS+} &           61.16\textcolor{gray}{\tiny{$\pm$0.67}} &           60.17\textcolor{gray}{\tiny{$\pm$0.67}} &           61.87\textcolor{gray}{\tiny{$\pm$0.66}} &           58.18\textcolor{gray}{\tiny{$\pm$0.72}} \\
\midrule
\multirow{4}{*}{\specialcell{Simple-\\Shot}} & \specialcell{Random-Shot} &  \textbf{61.99\textcolor{gray}{\tiny{$\pm$0.71}}} &           59.35\textcolor{gray}{\tiny{$\pm$0.71}} &           57.57\textcolor{gray}{\tiny{$\pm$0.59}} &           50.41\textcolor{gray}{\tiny{$\pm$0.88}} \\
                    & \specialcell{$\rightarrow$ ROS+} &           61.12\textcolor{gray}{\tiny{$\pm$0.70}} &  \textbf{59.50\textcolor{gray}{\tiny{$\pm$0.71}}} &           58.39\textcolor{gray}{\tiny{$\pm$0.61}} &  \textbf{54.88\textcolor{gray}{\tiny{$\pm$0.76}}} \\
                    & \specialcell{ROS $\rightarrow$ ROS} &           59.52\textcolor{gray}{\tiny{$\pm$0.73}} &           55.39\textcolor{gray}{\tiny{$\pm$0.71}} &           55.67\textcolor{gray}{\tiny{$\pm$0.60}} &           47.63\textcolor{gray}{\tiny{$\pm$0.80}} \\
                    & \specialcell{ROS+ $\rightarrow$ ROS+} &           59.44\textcolor{gray}{\tiny{$\pm$0.69}} &           57.89\textcolor{gray}{\tiny{$\pm$0.70}} &  \textbf{59.84\textcolor{gray}{\tiny{$\pm$0.60}}} &           49.95\textcolor{gray}{\tiny{$\pm$0.83}} \\
\midrule
\multirow{4}{*}{\specialcell{MAML}} & \specialcell{Random-Shot} &           50.79\textcolor{gray}{\tiny{$\pm$0.67}} &           49.25\textcolor{gray}{\tiny{$\pm$0.67}} &           48.32\textcolor{gray}{\tiny{$\pm$0.61}} &           43.14\textcolor{gray}{\tiny{$\pm$0.71}} \\
                    & \specialcell{$\rightarrow$ ROS+} &           50.39\textcolor{gray}{\tiny{$\pm$0.69}} &           49.70\textcolor{gray}{\tiny{$\pm$0.69}} &           49.95\textcolor{gray}{\tiny{$\pm$0.67}} &           48.11\textcolor{gray}{\tiny{$\pm$0.70}} \\
                    & \specialcell{ROS $\rightarrow$ ROS} &  \textbf{59.10\textcolor{gray}{\tiny{$\pm$0.70}}} &  \textbf{58.30\textcolor{gray}{\tiny{$\pm$0.69}}} &  \textbf{56.71\textcolor{gray}{\tiny{$\pm$0.63}}} &  \textbf{54.09\textcolor{gray}{\tiny{$\pm$0.73}}} \\
                    & \specialcell{ROS+ $\rightarrow$ ROS+} &           54.60\textcolor{gray}{\tiny{$\pm$0.72}} &           54.41\textcolor{gray}{\tiny{$\pm$0.70}} &           56.35\textcolor{gray}{\tiny{$\pm$0.71}} &           51.04\textcolor{gray}{\tiny{$\pm$0.76}} \\
\midrule
\multirow{4}{*}{\specialcell{Proto-\\MAML}} & \specialcell{Random-Shot} &           62.22\textcolor{gray}{\tiny{$\pm$0.73}} &           61.64\textcolor{gray}{\tiny{$\pm$0.72}} &           59.23\textcolor{gray}{\tiny{$\pm$0.64}} &           57.36\textcolor{gray}{\tiny{$\pm$0.75}} \\
                    & \specialcell{$\rightarrow$ ROS+} &           62.22\textcolor{gray}{\tiny{$\pm$0.73}} &           61.60\textcolor{gray}{\tiny{$\pm$0.74}} &           61.71\textcolor{gray}{\tiny{$\pm$0.70}} &           59.49\textcolor{gray}{\tiny{$\pm$0.75}} \\
                    & \specialcell{ROS $\rightarrow$ ROS} &           62.16\textcolor{gray}{\tiny{$\pm$0.71}} &           60.94\textcolor{gray}{\tiny{$\pm$0.72}} &           59.88\textcolor{gray}{\tiny{$\pm$0.66}} &           57.29\textcolor{gray}{\tiny{$\pm$0.75}} \\
                    & \specialcell{ROS+ $\rightarrow$ ROS+} &  \textbf{63.55\textcolor{gray}{\tiny{$\pm$0.71}}} &  \textbf{63.02\textcolor{gray}{\tiny{$\pm$0.71}}} &  \textbf{63.08\textcolor{gray}{\tiny{$\pm$0.69}}} &  \textbf{60.65\textcolor{gray}{\tiny{$\pm$0.75}}} \\
\midrule
\multirow{4}{*}{\specialcell{BTAML}} & \specialcell{Random-Shot} &           64.07\textcolor{gray}{\tiny{$\pm$0.70}} &           63.71\textcolor{gray}{\tiny{$\pm$0.70}} &           62.52\textcolor{gray}{\tiny{$\pm$0.65}} &           60.31\textcolor{gray}{\tiny{$\pm$0.74}} \\
                    & \specialcell{$\rightarrow$ ROS+} &           64.07\textcolor{gray}{\tiny{$\pm$0.70}} &           63.44\textcolor{gray}{\tiny{$\pm$0.70}} &  \textbf{63.71\textcolor{gray}{\tiny{$\pm$0.68}}} &  \textbf{61.11\textcolor{gray}{\tiny{$\pm$0.74}}} \\
                    & \specialcell{ROS $\rightarrow$ ROS} &  \textbf{64.17\textcolor{gray}{\tiny{$\pm$0.70}}} &  \textbf{63.72\textcolor{gray}{\tiny{$\pm$0.69}}} &           62.81\textcolor{gray}{\tiny{$\pm$0.66}} &           60.47\textcolor{gray}{\tiny{$\pm$0.74}} \\
                    & \specialcell{ROS+ $\rightarrow$ ROS+} &           63.56\textcolor{gray}{\tiny{$\pm$0.72}} &           63.02\textcolor{gray}{\tiny{$\pm$0.72}} &           63.54\textcolor{gray}{\tiny{$\pm$0.70}} &           61.11\textcolor{gray}{\tiny{$\pm$0.74}} \\
\bottomrule
\end{tabular}}
    \label{tbl randomshot_vs_imbalanced_tasks}
\end{table}

\textbf{Improving Random-Shot.} Most meta-learners lack a class balancing mechanism that could allow them to balance the incoming data. In this paragraph, we explore whether we can leverage ROS/ROS+ as rebalancing strategies to improve the model performance. Table~\ref{tbl randomshot_vs_imbalanced_tasks} shows the average performance between various meta-training/meta-evaluation techniques and the baseline Random-Shot Meta-Training. The right arrow ($\rightarrow$) separates the method applied at training time from the method applied at testing time. The table reveals that ROS/ROS+ are effective. Specifically, Matching Net, Relation Net, and MAML see the greatest improvement over plain Random-Shot from applying ROS during meta-training and meta-evaluation, yielding an average improvement of $+3.3\%$, $+0.4\%$, $+9.2\%$, respectively, across all four tasks. ROS applied to MAML is able to balance the inner-loop steps yielding a significant improvement. On the other hand, ProtoNet, ProtoNet (20-way), and ProtoMAML benefit the most from ROS+ applied at both the meta-phases, yielding on average $+8.7\%$, $+3.2\%$, $+2.5\%$, respectively. This advantage could stem from the additional augmentation during meta-training, regularizing the prototypes, and resulting in more linearly separable prototype representations. DKT, SimpleShot, and BTAML seem to benefit the most from ROS+ applied at the meta-testing time only, giving $+1.2\%$, $+2.6\%$, $+1.0\%$ on average on the last two tasks. SimpleShot sees minor improvements, which could be due to its backbone being pre-trained on mini-batches (instead of FSL tasks) and being exposed to imbalance tasks only during meta-validation; therefore, ROS/ROS+ are not applied on the full pipeline and do not affect the training of the backbone. Overall, methods that learn to balance or those naturally robust to imbalance (i.e., Relation Net, BTAML, or DKT) benefit less from oversampling techniques. This is likely due to diminishing returns as the accuracy increases.

\subsubsection{Ablation Studies: Depth of the CIFSL Problem}\label{sec exp ablation}
In this section, we provide additional experiments and ablations to provide further insight and depth of the class imbalance problem using more shots, different backbones, and re-balancing methods. Overall, the experiments show that imbalance persists in larger tasks and deeper backbones, although its impact gets less significant (by about $-3\%$) as the task-size increases compared to the smallest tasks. Additional results are presented in Appendix~\ref{appendix ablation}.

\begin{figure}[htb]
    \centering
    \includegraphics[width=0.99\linewidth]{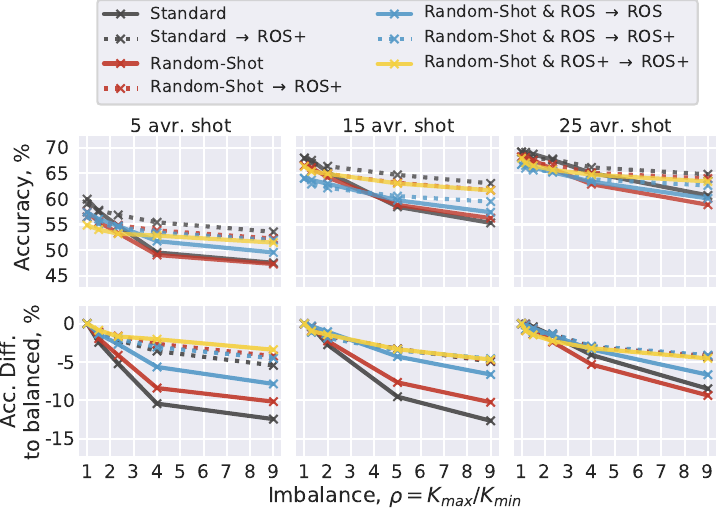}
    \caption{Comparing imbalance levels via support sets of different size. Each line represents the average across all models in each training and imbalance setting.}\label{fig linear_more_diff}
\end{figure}

\textbf{Imbalance with More Shots.} 
We explored additional settings with a higher number of shots. Specifically, we trained models using Random-Shot Meta-Training with 1-29 shot and 1-49 random episodes. We then evaluated those models on imbalanced tasks with an average number of 15 shots and 25 shots, respectively. The bottom row of Figure~\ref{fig linear_more_diff} shows the difference in performance between the balanced ($\rho=1$) and unbalanced ($\rho>1$) tasks. From the plot, we can make at least three observations. First, models achieve $55$-$60\%$ of their performance on the balanced task within the first 5 average shots. Increasing three times the number of shots only boosts the performance by $+7\%$. Secondly, as the task size increases, models get less affected by imbalance. Thirdly, even with 25 average shots, the gap between the balanced and the imbalanced tasks remains significant.

\begin{figure}[htb]
    \centering
    \includegraphics[width=0.99\linewidth]{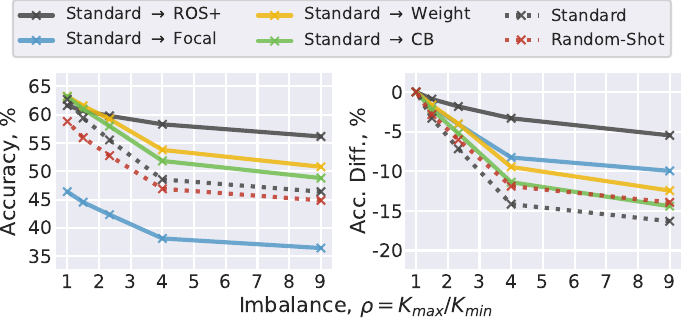}
    \caption{Average model performance against re-balancing strategies applied at test-time only. \emph{Left:} all models and training scenarios. \emph{Right:} performance w.r.t. the balanced task.}\label{fig cost_func_mean}
\end{figure}

\textbf{Rebalancing Losses vs ROS+.} Here, we investigate whether other class imbalance approaches can offer an improvement in performance. For this, we pair optimization-based models \cite{Triantafillou2019meta,Finn2017maml,Chen2019closer} with rebalancing loss functions: Focal Loss \cite{Lin2017focal}, Weighted Loss \cite{Buda2018imbalance,Japkowicz2002imbalance}, and Class Balancing (CB) Loss \cite{Cui2019cbloss}. We apply each loss to the inner-loop of optimization-based methods and fine-tuning baselines at meta-test time. Note that other methods have not been considered because rebalancing losses do not directly apply (no inner-loop). Results in Figure~\ref{fig cost_func_mean} and Appendix~\ref{appendix loss} show overall the loss functions are not as effective as ROS+ applied at inference. Specifically, we observe that Weighted Loss has a drop of $-3\%$ in accuracy compared to Standard. CB Loss did not offer any advantage over the Weighted Loss suggesting that long-tail rebalancing mechanisms do not necessarily generalize to the few-shot setting. Note that a key advantage of ROS/ROS+ is versatility; any FSL algorithm can use ROS, while rebalancing cost functions do not extend to metric-learning methods with no inner-loop. A per-model break-down of Figure~\ref{fig cost_func_mean} is in Appendix~\ref{appendix loss}.

\begin{figure}[htb]
    \centering
    \includegraphics[width=0.95\linewidth]{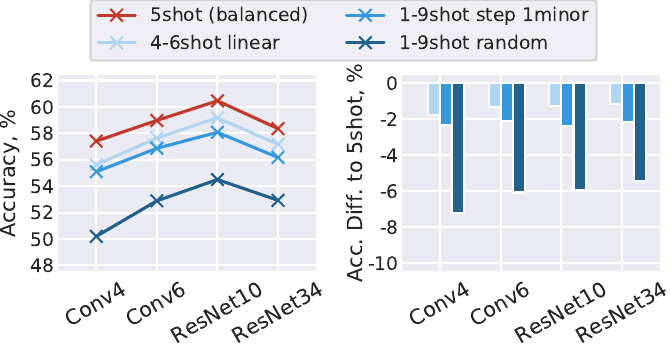}
    \caption{Average model performance against different backbones and imbalanced tasks. \emph{Left:} combined performance of all models and training scenarios. \emph{Right:} relative performance w.r.t. the balanced task.}\label{fig fsl_backbone}
\end{figure}

\textbf{Backbones.} In Figure~\ref{fig fsl_backbone}, we report the combined average accuracy of all models and imbalance strategies against different backbones (Conv4, Conv6, ResNet10, ResNet34). Overall, deeper backbones seem to perform slightly better on the imbalanced tasks, suggesting a higher tolerance for imbalance. For instance, using Conv4 gave $-7.2\%$ difference between the balanced and the 1-9shot random task, while using ResNet10, the gap is smaller ($-6.0\%$). The performance degradation observed with ResNet34 is similar to that reported by \cite{Chen2019closer}, and it is caused by the intrinsic instability of meta-training routines on larger backbones.

\begin{table*}[tbh]
    \caption{Evaluation accuracy after meta-training on $\mathcal{D}'_{train}$ derived from Mini-ImageNet with various imbalance distributions. Small differences in accuracy between $balanced$ and other distributions, suggest a small effect of imbalance at dataset level. \emph{Left:} Evaluation on the meta-testing dataset of Mini-ImageNet. \emph{Right:} Evaluation on the meta-testing dataset of CUB.
    } \label{tbl imbalanced_meta_dataset_5shot_5way}
    \centering
    \scalebox{0.91}{ \begin{tabular}{r|cHccc|cHccc}
\toprule
$\mathcal{D}'_{train} \rightarrow \mathcal{D}_{test}$ & \multicolumn{5}{c|}{Mini-ImageNet $\rightarrow$ Mini-ImageNet} & \multicolumn{5}{c}{Mini-ImageNet $\rightarrow$ CUB} \\
Imbalance in $\mathcal{D}'_{train}$  &  \specialcell{ $balanced$ \\ {\scriptsize(300, 300, 64, -) }} &  \specialcell{ $linear$ \\ {\scriptsize (30, 570, 64, -)} } & \specialcell{  $step$-32 \\ {\scriptsize(30, 570, 64, 32)} } & \specialcell{  $step$-22 \\ {\scriptsize(25, 444, 64, 22)} } & \specialcell{  $step$-animal \\ {\scriptsize(25, 444, 64, 22)} }  &  \specialcell{ $balanced$ \\ {\scriptsize(300, 300, 64, -) }} &  \specialcell{ $linear$ \\ {\scriptsize (30, 570, 64, -)} }  & \specialcell{  $step$-32 \\ {\scriptsize(30, 570, 64, 32)} } & \specialcell{  $step$-22 \\ {\scriptsize(25, 444, 64, 22)} }  & \specialcell{  $step$-animal \\ {\scriptsize(25, 444, 64, 22)} } \\
\midrule
Baseline (fine-tune) &           61.23\textcolor{gray}{\tiny{$\pm$0.72}} &           60.93\textcolor{gray}{\tiny{$\pm$0.70}} &           60.44\textcolor{gray}{\tiny{$\pm$0.71}} &           60.36\textcolor{gray}{\tiny{$\pm$0.71}} &           59.32\textcolor{gray}{\tiny{$\pm$0.73}} &           56.61\textcolor{gray}{\tiny{$\pm$0.72}} &           56.43\textcolor{gray}{\tiny{$\pm$0.73}} &           55.83\textcolor{gray}{\tiny{$\pm$0.71}} &           56.35\textcolor{gray}{\tiny{$\pm$0.72}} &           52.83\textcolor{gray}{\tiny{$\pm$0.73}} \\
Baseline++           &           63.86\textcolor{gray}{\tiny{$\pm$0.67}} &           63.97\textcolor{gray}{\tiny{$\pm$0.67}} &  \textbf{63.23\textcolor{gray}{\tiny{$\pm$0.66}}} &  \textbf{62.91\textcolor{gray}{\tiny{$\pm$0.67}}} &           62.03\textcolor{gray}{\tiny{$\pm$0.70}} &           57.71\textcolor{gray}{\tiny{$\pm$0.72}} &           57.33\textcolor{gray}{\tiny{$\pm$0.74}} &           56.62\textcolor{gray}{\tiny{$\pm$0.75}} &  \textbf{57.07\textcolor{gray}{\tiny{$\pm$0.75}}} &           52.07\textcolor{gray}{\tiny{$\pm$0.71}} \\
Matching Net         &           62.05\textcolor{gray}{\tiny{$\pm$0.69}} &           61.39\textcolor{gray}{\tiny{$\pm$0.70}} &           58.23\textcolor{gray}{\tiny{$\pm$0.69}} &           58.82\textcolor{gray}{\tiny{$\pm$0.70}} &           59.17\textcolor{gray}{\tiny{$\pm$0.71}} &           52.64\textcolor{gray}{\tiny{$\pm$0.74}} &           51.52\textcolor{gray}{\tiny{$\pm$0.76}} &           51.10\textcolor{gray}{\tiny{$\pm$0.74}} &           50.20\textcolor{gray}{\tiny{$\pm$0.75}} &           50.78\textcolor{gray}{\tiny{$\pm$0.75}} \\
ProtoNet             &           64.00\textcolor{gray}{\tiny{$\pm$0.71}} &           62.55\textcolor{gray}{\tiny{$\pm$0.71}} &           60.40\textcolor{gray}{\tiny{$\pm$0.70}} &           60.52\textcolor{gray}{\tiny{$\pm$0.70}} &           61.22\textcolor{gray}{\tiny{$\pm$0.70}} &           54.33\textcolor{gray}{\tiny{$\pm$0.73}} &           52.78\textcolor{gray}{\tiny{$\pm$0.74}} &           52.19\textcolor{gray}{\tiny{$\pm$0.73}} &           52.78\textcolor{gray}{\tiny{$\pm$0.73}} &           52.34\textcolor{gray}{\tiny{$\pm$0.73}} \\
ProtoNet (20-way)    &  \textbf{65.41\textcolor{gray}{\tiny{$\pm$0.70}}} &           63.56\textcolor{gray}{\tiny{$\pm$0.70}} &           59.75\textcolor{gray}{\tiny{$\pm$0.71}} &           60.47\textcolor{gray}{\tiny{$\pm$0.70}} &           61.17\textcolor{gray}{\tiny{$\pm$0.72}} &           55.46\textcolor{gray}{\tiny{$\pm$0.74}} &           54.50\textcolor{gray}{\tiny{$\pm$0.74}} &           52.73\textcolor{gray}{\tiny{$\pm$0.73}} &           52.97\textcolor{gray}{\tiny{$\pm$0.71}} &           50.56\textcolor{gray}{\tiny{$\pm$0.72}} \\
Relation Net (CE)    &           63.78\textcolor{gray}{\tiny{$\pm$0.70}} &           63.29\textcolor{gray}{\tiny{$\pm$0.71}} &           58.15\textcolor{gray}{\tiny{$\pm$0.70}} &           60.50\textcolor{gray}{\tiny{$\pm$0.69}} &           59.66\textcolor{gray}{\tiny{$\pm$0.70}} &           55.83\textcolor{gray}{\tiny{$\pm$0.73}} &           55.67\textcolor{gray}{\tiny{$\pm$0.74}} &           52.90\textcolor{gray}{\tiny{$\pm$0.69}} &           52.87\textcolor{gray}{\tiny{$\pm$0.72}} &           52.35\textcolor{gray}{\tiny{$\pm$0.73}} \\
DKT                  &           62.31\textcolor{gray}{\tiny{$\pm$0.67}} &           61.79\textcolor{gray}{\tiny{$\pm$0.68}} &           59.01\textcolor{gray}{\tiny{$\pm$0.68}} &           59.27\textcolor{gray}{\tiny{$\pm$0.69}} &           59.82\textcolor{gray}{\tiny{$\pm$0.67}} &  \textbf{58.01\textcolor{gray}{\tiny{$\pm$0.74}}} &  \textbf{57.91\textcolor{gray}{\tiny{$\pm$0.72}}} &  \textbf{56.97\textcolor{gray}{\tiny{$\pm$0.73}}} &           56.72\textcolor{gray}{\tiny{$\pm$0.73}} &  \textbf{56.26\textcolor{gray}{\tiny{$\pm$0.72}}} \\
SimpleShot           &           62.53\textcolor{gray}{\tiny{$\pm$0.71}} &           62.47\textcolor{gray}{\tiny{$\pm$0.72}} &           62.34\textcolor{gray}{\tiny{$\pm$0.72}} &           62.03\textcolor{gray}{\tiny{$\pm$0.71}} &           61.26\textcolor{gray}{\tiny{$\pm$0.75}} &           56.64\textcolor{gray}{\tiny{$\pm$0.74}} &           56.16\textcolor{gray}{\tiny{$\pm$0.74}} &           55.54\textcolor{gray}{\tiny{$\pm$0.73}} &           56.01\textcolor{gray}{\tiny{$\pm$0.73}} &           52.43\textcolor{gray}{\tiny{$\pm$0.72}} \\
MAML                 &           61.20\textcolor{gray}{\tiny{$\pm$0.72}} &           60.91\textcolor{gray}{\tiny{$\pm$0.72}} &           58.29\textcolor{gray}{\tiny{$\pm$0.73}} &           58.97\textcolor{gray}{\tiny{$\pm$0.71}} &           59.36\textcolor{gray}{\tiny{$\pm$0.71}} &           55.62\textcolor{gray}{\tiny{$\pm$0.74}} &           55.36\textcolor{gray}{\tiny{$\pm$0.74}} &           55.13\textcolor{gray}{\tiny{$\pm$0.74}} &           54.50\textcolor{gray}{\tiny{$\pm$0.73}} &           54.57\textcolor{gray}{\tiny{$\pm$0.74}} \\
ProtoMAML            &           64.78\textcolor{gray}{\tiny{$\pm$0.70}} &  \textbf{64.30\textcolor{gray}{\tiny{$\pm$0.70}}} &           60.67\textcolor{gray}{\tiny{$\pm$0.71}} &           61.77\textcolor{gray}{\tiny{$\pm$0.71}} &  \textbf{62.41\textcolor{gray}{\tiny{$\pm$0.71}}} &           57.02\textcolor{gray}{\tiny{$\pm$0.74}} &           57.37\textcolor{gray}{\tiny{$\pm$0.74}} &           55.32\textcolor{gray}{\tiny{$\pm$0.73}} &           55.82\textcolor{gray}{\tiny{$\pm$0.71}} &           54.88\textcolor{gray}{\tiny{$\pm$0.73}} \\
\midrule
\specialcell{Avr. Diff. to $balanced$} &          -     & - & -3.1 &	-2.6 &	-2.6	& - & - &	-1.6	& -1.5	& -3.1 \\
\bottomrule
\end{tabular}

  } 
\end{table*}

\subsection{Robustness to Dataset Imbalance}\label{sec imb dataset}
In this section, we are keeping tasks balanced and isolate class imbalance in the meta-training dataset. Overall, we find that meta-dataset imbalance has just a small negative effect on the generalization of ML models to novel tasks.

\subsubsection{Step Distribution}
\textbf{Robustness against domain-shift.} In this paragraph, we examine various $step$ imbalance distributions and evaluate performance with domain-shift. Step imbalance emulates the scenario of joining two balanced datasets. Table~\ref{tbl imbalanced_meta_dataset_5shot_5way} shows the meta-testing performance on $\mathcal{D}_{test}$ of Mini-ImageNet (left column group) or CUB (right column group) after training on $step$ imbalanced $\mathcal{D}'_{train}$. The bottom row shows the average model difference in accuracy between $balanced$ and the imbalanced datasets. The results show a small negative difference ($-3.1\%$) between the $balanced$ dataset and the step scenario with 32 minority classes ($step$-32). Evaluation of models trained under the $step$-32 distribution on CUB shows a small drop ($-1.6$) under the domain-shift condition. Note that $D'_{train}$ contains 22 animal classes, including 3 classes of birds, which might not represent a very large domain-shift.

\textbf{Robustness against larger domain-shift.} Next, we examined a stronger domain-shift scenario. We set all 22 animal classes in $\mathcal{D}'_{train}$ to contain $K_{min}^{\mathcal{D}}=25$ samples each, and we set the other 42 classes to have $K_{max}^{\mathcal{D}}=444$ samples each. We call this setting ``$step$-animal'' as it reduces the diversity of animal samples seen during meta-training. We examine a similar $step$ imbalance with 22 minority classes picked uniformly at random ($step$-22) as a control variable. Results in Table~\ref{tbl imbalanced_meta_dataset_5shot_5way} suggest that $step$-animal performs slightly worse ($-1.6\%$) compared to $step$-22 on average. SimpleShot and the two Baselines are particularly affected by the larger domain-shift on CUB, with a performance drop of $-4.0\%$ on $step$-animal compared to $step$-22. Perhaps, this drop is due to the particular training procedure used for these methods, which are pre-trained on mini-batches instead of tasks. This suggests an implicit strength of ML algorithms against larger domain-shift. 

\subsubsection{Small Datasets}
\begin{figure}[tbh]
    \centering
    \includegraphics[width=0.75\linewidth]{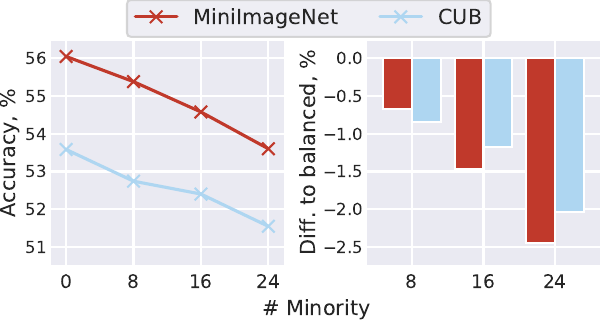}
    \caption{Combined model average performance with increasing minority classes. \emph{Left:} Combined accuracy of all models. \emph{Right:} Performance difference to the balanced dataset. Full results are in Appendix~\ref{appendix reduced}.}\label{fig reduced_meta}
\end{figure}

In this section, we examine whether a smaller dataset size could influence the effect of imbalance. Specifically, we construct a new set of datasets denoted by $\mathcal{D}''_{train}$ containing 1/8\textsuperscript{th} of samples in $\mathcal{D}_{train}$ of Mini-ImageNet, and 32 classes selected uniformly at random, $|\mathcal{D}''_{train}|=4800$. In Figure~\ref{fig reduced_meta}, we see that overall the performance drops as the number of minority classes increases, but the effect still remains quite low ($<3\%$ absolute difference).

\subsubsection{Long-Tail Distribution}
\textbf{Robustness against larger imbalance.} In this section, we examine $\longtail$ distributions with an imbalance ratio $\rho = 65$ on a larger dataset ($6.5\times$ larger than $\mathcal{D}'_{train}$, and derived directly from ImageNet). This setting reflects the observation made by \cite{Reed2001powerlaw} about the real-world imbalance that the top $20\%$ majority classes account for $80\%$ of all data points in $\mathcal{D}_{train}$. Specifically, we partitioned ImageNet to contain 900 classes for $\mathcal{D}_{train}$ distributed according to $balanced$ or $\longtail$, while $\mathcal{D}_{val}$ and $\mathcal{D}_{test}$ contained 50 balanced classes each with 500 randomly selected samples per class. The set of classes for each dataset partition were kept the same for all experiment, but classes within each partition were shuffled between three repeats. We induce $\longtail$ imbalance using the Power-Law distribution \cite{Reed2001powerlaw} with a power of 10, a minimum of 20 samples per class (to allow 5-shot 15-query task), and a maximum of 1300 samples per class for ImageNet. The $balanced$ dataset contained samples distributed uniformly among the 900 classes (i.e. 137 samples per class) such that $|\mathcal{D}_{train}|\approx123300$ for both distributions. We used ResNet-10 as the backbone. The results in Table~\ref{tbl longtail} show the accuracy performance after training on 1800 balanced tasks across 3 seeds. The results suggest that meta-learner models are more susceptible to the larger imbalance, experiencing between $-4.1\%$ and $-6.3\%$ drop in accuracy. This suggests that a more natural imbalance distribution and a higher imbalance ratio could cause a larger performance drop. Interestingly, the Baseline and SimpleShot are least affected with a non-significant drop in performance ($-0.1\%$). This discrepancy could be caused by the natural property of standard task-sampling procedure \cite{Vinyals2017matching} which sample all classes with equal probability and lead to overexposure to minority class samples and over-fitting.

\begin{table}[tbh]
    \centering
    \caption{Accuracy performance on Long-Tail distribution vs balanced with approximately the same number of samples in total.}
    \scalebox{0.99}{\begin{tabular}{l|cc}
\toprule
$I$ in $\mathcal{D}_{train}$ & \specialcell{$balanced$\\{\scriptsize(137, 137, 900,-)}} & \specialcell{$\longtail$\\{\scriptsize(20, 1300, 900,-)}} \\
\midrule
Baseline & 62.57\textcolor{gray}{\tiny{$\pm$0.85}} & 62.47\textcolor{gray}{\tiny{$\pm$0.85}} \\
SimpleShot & 63.36\textcolor{gray}{\tiny{$\pm$0.86}} & 62.20\textcolor{gray}{\tiny{$\pm$0.87}} \\
Matching Net &	64.51\textcolor{gray}{\tiny{$\pm$0.88}} & 59.39\textcolor{gray}{\tiny{$\pm$0.88}} \\
DKT & 64.77\textcolor{gray}{\tiny{$\pm$0.86}} & 59.76\textcolor{gray}{\tiny{$\pm$0.86}} \\
MAML     & 62.83\textcolor{gray}{\tiny{$\pm$0.90}} & 58.69\textcolor{gray}{\tiny{$\pm$0.90}} \\
ProtoNet & 67.97\textcolor{gray}{\tiny{$\pm$0.83}} & 61.67\textcolor{gray}{\tiny{$\pm$0.86}} \\
ProtoMAML & 65.44\textcolor{gray}{\tiny{$\pm$0.88}} & 60.97\textcolor{gray}{\tiny{$\pm$0.89}} \\
\bottomrule
\end{tabular}}
    \label{tbl longtail}
\end{table}

\begin{figure*}
    \centering
    \includegraphics[width=0.99\textwidth]{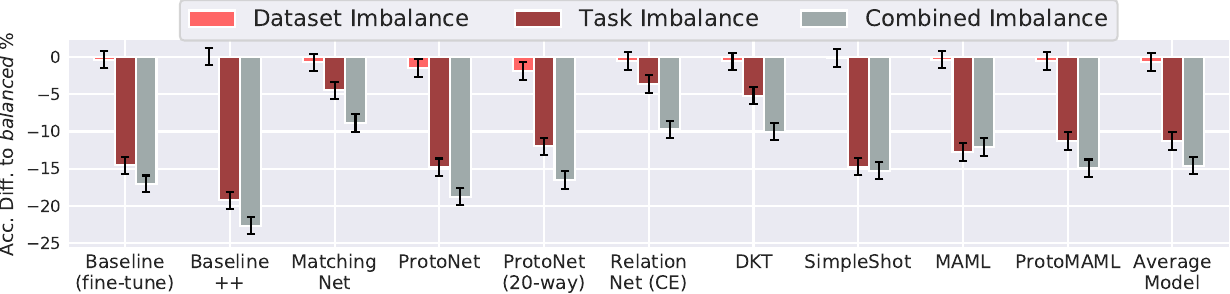}
    \caption{Difference in accuracy between dataset imbalance (light red), task imbalance (dark red), and combined imbalance (grey) against the $balanced$ meta-training baseline. Negative performance indicates a lower accuracy compared to $balanced$. The results suggest that models are quite robust to dataset imbalance compared to imbalance at the task-level.}
    \label{fig linear_task_vs_dataset}
\end{figure*}

\subsection{Robustness to Combined Imbalance}\label{sec imb combined}
Finally, we compare dataset imbalance in Mini-ImageNet to task imbalance using a similar distribution. In the \emph{dataset imbalance} condition, tasks (5-shot, 5-way) are sampled from a linearly-imbalanced dataset $\mathcal{D}'_{train}$ (30,570,64,-). In the \emph{task imbalance} condition, tasks are sampled from a linearly-imbalanced distribution (1,9,5,-) -- also called 1-9shot 5-way -- and from a balanced dataset $\mathcal{D}'_{train}$ (300,300,64,-). Finally, in the \emph{combined imbalance} condition imbalanced tasks (1,9,5,-) are sampled from the imbalanced dataset $\mathcal{D}'_{train}$ (30,570,64,-).  Figure~\ref{fig linear_task_vs_dataset} shows the difference in $\mathcal{D}_{test}$ accuracy between dataset imbalance (light red) and task imbalance (dark red) compared against the score of standard training ($balanced$ at both dataset and task levels). Dataset imbalance does not cause a significant drop in performance, being mostly within error-bars with respect to $balanced$. On the other hand, the imbalanced task condition causes a significant drop in performance, up to $-20\%$ for some methods despite the slightly smaller imbalance magnitude ($\rho=9$ c.f. $\rho=19$). This result suggests that meta-learners are quite robust to dataset imbalance.  Interestingly, the combined task and dataset imbalance seems to have a slight compounding effect on performance, yielding $-15\%$ on average, and performing worse than the combined sum of the two imbalance conditions taken individually ($-12\%$).

\section{Discussion}\label{discussion}

\textbf{Robustness to Task Imbalance.} All examined FSL methods are susceptible to class imbalance, although some show more robustness (e.g., Matching Net, Relation Net, and DKT). Optimization-based methods and fine-tune baselines suffer more as they use conventional supervised loss functions in the inner-loop, which are known to be particularly susceptible to imbalance \cite{Buda2018imbalance,Johnson2019imbalance,Japkowicz2002imbalance}. Moreover, the problem of class imbalance persists as the backbone depth and support set size increase. Those results suggest that current solutions will offer sub-optimal performance in real-world few-shot problems.

\textbf{Effectiveness of Random-Shot meta-training.} Our experiments evaluated a popular meta-training procedure in the meta-learning community -- Random-Shot  \cite{Triantafillou2019meta,Lee2020baysiantaml,Chen2020mamlstop}, which exposes models to random distribution tasks during meta-training. Our findings reveal that this procedure is scarcely effective when evaluated on a variety of imbalance distributions (linear, step, balanced), even though they are still considered \emph{in-distribution} \cite{Lee2020baysiantaml} of the random training tasks. This finding suggests that many popular meta-learning models cannot obtain robustness to imbalance by the simple exposure to imbalanced tasks.

\textbf{Effectiveness of re-balancing procedures.} The results suggest that simple procedures, such as Random Over-Sampling (ROS) and Random Over-Sampling with augmentation (ROS+), are quite effective in tackling class imbalance issues, showing that they are still relevant and should not be forgotten. ROS and ROS+ yield large advantages on optimization-based methods, with improvements in accuracy between $+5\%$ and $+20\%$ in the 1-9shot random task. Using random oversampling during Random-Shot Meta-Training can also boost the performance of this method. ROS and ROS+ can be used as simple yet effective baselines, since thanks to their flexibility, they can be applied to any episodic training algorithm. However, we also note that ROS does not provide any particular advantage to methods in the highest performance ranks (e.g. MatchingNet, DKT, and Relation Net). This finding highlights the strength of those methods and shows the limitations of ROS and the need for further investigation in the CIFSL setting.

\textbf{Robustness to Dataset Imbalance.} In this work, we also provided insights into the meta-dataset imbalance problem in meta-learning, showing that models are quite robust to meta-dataset imbalance, experiencing only small drops in accuracy points when exposed to $\rho \leq 20$. In contrast, the support-set imbalance yields a significantly larger (an order of magnitude) performance drop. This seems to point to an implicit strength of meta-learning algorithms, which are able to learn generalizable features if exposed to balanced tasks during the inner-loop optimization. Higher levels of step and long-tail imbalance will likely produce more dramatic performance differences. Additionally, combining task and dataset imbalance yields compounding degradation in performance. In this study, we only evaluated generalization to novel tasks and classes, but it is likely that imbalance in $\mathcal{D}_{train}$ would impact performance on base tasks and classes sampled from $D_{train}$. Moreover, our experiments were constrained by the size of the meta-training dataset of Mini-ImageNet and ImageNet as we had to control its size for a fair comparison. In the real-world, meta-datasets can be very large with very high imbalance ratios ($\rho\gg100$) and following long-tail distributions \cite{Liu2019tailed,Salakhutdinov2011longtail}. Future work should investigate how these larger imbalance levels could affect meta-learning and cross-domain generalization.

\section{Conclusion}\label{sec conclude}
In this work, we have provided a detailed analysis of CIFSL, showing that class imbalance at the support set level is problematic for many methods, yielding up to $-17\%$ performance drop compared to balanced tasks. Metric-based models show built-in robustness to support-set imbalance and largely outperform optimization-based models even when combined with classical rebalancing approaches. Random-Shot Meta-Training provides minimal benefit but can be improved if coupled with classical rebalancing techniques. Surprisingly, simple techniques such as ROS and Rebalancing Losses, overlooked in CIFSL, can be very effective and lead to state-of-the-art performance. In future work, the insights gained with this investigation could be used to design novel few-shot methods that can guarantee a stable performance under realistic class imbalance settings.

\section*{Acknowledgments}
We want to thank Eleni Triantafillou, Hae Beom Lee, Hayeon Lee, and the members of the Bayesian and Neural Systems group at the University of Edinburgh for valuable comments, suggestions, and discussions offered at various stages of this work. This work was supported by the EPSRC Centre for Doctoral Training in Robotics and Autonomous Systems, funded by the UK Engineering and Physical Sciences Research Council (Grant No. EP/S515061/1) and SeeByte Ltd, Edinburgh, UK.

\bibliographystyle{IEEEtran}
\bibliography{bib}

\clearpage
\appendices

\onecolumn

\section{Implementation Details}\label{appendix implementation}

\subsection{Datasets}\label{appendix dataset details}
For the imbalanced support set experiments, we used MiniImageNet \cite{Vinyals2017matching, Ravi2017} following \cite{Ravi2017,Cheng2019transfer}. All metric and optimization models used 64, 16, 20 classes for the meta-training ($\mathcal{D}_{train}$), meta-validation ($\mathcal{D}_{val}$), and meta-testing ($\mathcal{D}_{test}$) datasets, respectively, with each class containing 600 samples. All images are resized to 84 by 84 pixels. For the feature-transfer baselines (fine-tune, and Baseline\texttt{++}), we used a conventionally partitioned training and validation datasets. Specifically, we combined $\mathcal{D}_{train}$ and $\mathcal{D}_{val}$ classes (i.e., $64 + 16 = 80$ classes), then partitioned the samples of each class into 80\% - 20\% split for pre-training - validation, forming $\mathcal{D}'_{train}$ and $\mathcal{D}'_{val}$ (where $\mathcal{D}'^{(y)}_{train} \equiv \mathcal{D}'^{(y)}_{val}$). Thus, the baseline were trained on the same number of training samples as the meta-learning methods, albeit with more classes and less samples per class. All models (the baselines and other FSL methods) were evaluated on FSL tasks sampled from $\mathcal{D}_{test}$. 

\subsection{Training Procedure.}
All methods follow a similar three-phase learning procedure: meta-training, meta-validation, and meta-testing. At the end of the meta-training phase, the best model was evaluated on tasks sampled from $\mathcal{D}_{test}$. The baselines (i.e., fine-tune, Baseline\texttt{++}) followed a similar three-phase procedure but with the meta-training / meta-validation phases exchanged for conventional pre-training / validation on mini-batches. We used 15 query samples per class in all tasks, except for the 20-way Prototypical Network, where we used 5 query samples per class during meta-training to allow for a higher number of samples in the support set. 

\textbf{Meta-Training / Pre-Training.} During meta-training, the FSL models were exposed to 100k tasks sampled from $\mathcal{D}_{train}$. The imbalance distribution of a training task was either $balanced$ (when Standard Meta-Training) or $random$ (when Random-Shot Meta-Training) with the number of shots detailed in the main paper. The baselines were trained on 100k mini-batches of size 128 sampled from $\mathcal{D}'_{train}$ (defined in Appendix~\ref{appendix dataset details}). After every 500 training tasks/mini-batches, the model was validated on tasks from $\mathcal{D}_{val}$ / $\mathcal{D'}_{val}$ and the best performing model was updated. We tuned the models with the following learning rate ($lr$) combinations: $\{(1 \times 10^2, 1 \times 10^2), (1 \times 10^2, 1 \times 10^3), (1 \times 10^3, 1 \times 10^3), (1 \times 10^3, 1 \times 10^4), (1 \times 10^4, 1 \times 10^4), (1 \times 10^4, 1 \times 10^5)\}$, where the first number indicates the $lr$ for the first 50k tasks / mini-batches, and the second number is the $lr$ is the learning rate for the remaining tasks. We found that ($1 \times 10^{-3}, 1 \times 10^{-4}$) offered the best hyper-parameter combination for most models, and we used it in all our main experiments. 

\textbf{Meta-Testing.} We evaluate tasks using various imbalance severity levels and distributions, as specified in figures and tables. The final test performances were measured on a random sample of 600 tasks. We report the average $95\%$ mean confidence interval in brackets/error bars. 

\textbf{Data Augmentation.} During the meta-/pre-training phases, we apply standard data augmentation techniques, similarly to \cite{Chen2019closer}, with a random rotation of 10 degrees, scaling, random color/contrast/brightness jitter. Meta-validation and meta-testing had no augmentation apart from the \emph{Random-Shot (ROS+)} setting where the same augmentations were applied on the oversampled support images. All images are resized to 84 by 84 pixels.

\subsection{Backbone Architectures}\label{sec backbone}
All methods shared the same backbone architecture. For the core contribution of our work, we used Conv4 architecture consisting of 4 convolutional layers with 64 channels (padding 1), interleaved by batch normalization \cite{Ioffe2015batchnorm}, ReLU activation function, and max-pooling (kernel size 2, and stride 2) \cite{Chen2019closer}. Relation Network used max-pooling only for the last 2 layers of the backbone to account for the Relation Module. The Relation Module consisted of two additional convolutional layers, each followed by batch norm, ReLU, max-pooling).
For experiments with different backbones: Conv6, ResNet10, and ResNet34 \cite{Chen2019closer}. Conv6 extended the Conv4 backbone to 6 convolutional layers, and max-pooling applied after each first 4 layers. ResNet models \cite{he2015residual} followed the same setup as \cite{Chen2019closer}.

\subsection{FSL Methods and Baselines}
In our experiments, we used a wide range of FSL methods (full details can be found in our source code):
\begin{enumerate}
    \item \textbf{Baseline (fine-tune)} \cite{Pan2010transfer} represents a classical way of applying transfer learning, where a neural network is pre-trained on a large dataset, then fine-tuned on a smaller domain-specific dataset. The backbone is followed by a single linear layer with an output for each class in the meta-dataset. The whole network was trained during pre-training. During meta-testing, the pre-trained linear layer was removed and replaced by another randomly initialized classification layer with outputs matching the number of classes in the task ($N$-way). Fine-tuning was performed on the randomly initialized classification layer using the support set $\mathcal{S}$.
    \item \textbf{Baseline\texttt{++}} \cite{Chen2019closer} augments the fine-tune baseline by using Cosine Similarity on the last layer.
    \item \textbf{Matching Network (Matching Net)} \cite{Vinyals2017matching} uses context embeddings with an LSTM to effectively perform k-nearest neighbor in embedding space using cosine similarity to classify the query set.
    \item \textbf{Prototypical Networks (ProtoNet)} \cite{Snell2017proto} maps images into a feature space and estimates class means (called prototypes). The query samples are then classified based on Euclidian similarity to prototypes. We evaluate two models, one meta-trained like the others on 5-way episodes and another variation trained on 20-way episodes. During 20-way meta-training, we set the query size to 5.
    \item \textbf{Relation Networks (Relation Net)} \cite{Sung2017relationnet} augment the classical Prototypical Networks by introducing a relation module (another neural network) for comparing the similarities between support and query points. The original method uses Mean Squared Error to minimize the relation score between samples of the same type. However, we follow \cite{Chen2019closer} using cross-entropy (CE) loss which is more effective. The relation module is described in Section~\ref{sec backbone}. 
    \item \textbf{DKT} (formally called GPShot) proposed by \cite{Patacchiola2019gpshot} is a Bayesian approach that exploits deep kernels and Gaussian Processes to replace the inner-loop optimization of meta-learning with a Bayesian integral. The deep network is used as part of the kernel function and optimized across tasks. We used Batch Norm Cosine distance for the kernel type, same as reported by the authors; this was the most effective kernel. 
    \item \textbf{SimpleShot} \cite{Wang2019simpleshot} augments the 1-NN baseline model by normalizing and centering the feature vector using the mean feature vector of the dataset. The query samples are assigned to the nearest prototype class according to Euclidian distance. In contrast to the baseline models, pre-training is performed on the meta-training dataset like other meta-learning algorithms, and meta-validation is used to select the best model based on performance on tasks sampled from the validation set. 
    \item \textbf{MAML} \cite{Finn2017maml} is a meta-learning technique that learns a common initialization of weights that can be quickly adapted for the task using fine-tuning on the support set. The task adaptation process uses a standard gradient descent algorithm minimizing Cross-Entropy loss on the support set. The original method can be used with both first-order and second-order derivates; we used the first-order MAML, which has been shown to work just as well while being more efficient and stable. We set the inner-learning rate to 0.1 with 10 iteration steps. We optimize the meta-learner model on batches of 4 meta-training tasks. These hyperparameters were selected based on our hyperparameter fine-tuning.
    \item \textbf{ProtoMAML} \cite{Triantafillou2019meta} augments traditional first-order MAML by reinitializing the last classification layer between tasks. Specifically, the weights of the layer are assigned to the prototype for each corresponding output. This extra step combines the fine-tuning ability of MAML and the class regularization ability of Prototypical Networks. We set the inner-loop learning rate to 0.1 with 10 iterations. Unlike MAML, we found that updating the meta-learner after a single meta-training task gave the best performance.
    \item \textbf{Bayesian-TAML (BTAML)} \cite{Lee2020baysiantaml} This method augments MAML with four main changes: 1) task-dependent parameter initialization $z$, 2) task-dependent per-layer learning rate multipliers $\gamma$, 3) task-dependent per-class learning rate multiplier $\omega$, 4) meta-learned per-parameter learning rate $\alpha$ (similar to Meta-SGD, \cite{Li2017metasgd}). We explored several variations to this method with various parameters ($z,\omega,\gamma,\alpha$) turned on and off, as well as different hyperparameters. We found that using a meta-learning rate of $10^{-4}$ performed better than $10^{-3}$ in contrast to the other models. We set the inner-learning rate to 0.1 with 10 iteration steps and optimize the meta-learner model on batches of 4 meta-training tasks.
\end{enumerate}

\begin{figure}[htb]
    \centering
    \fbox{\includegraphics[height=4.25cm]{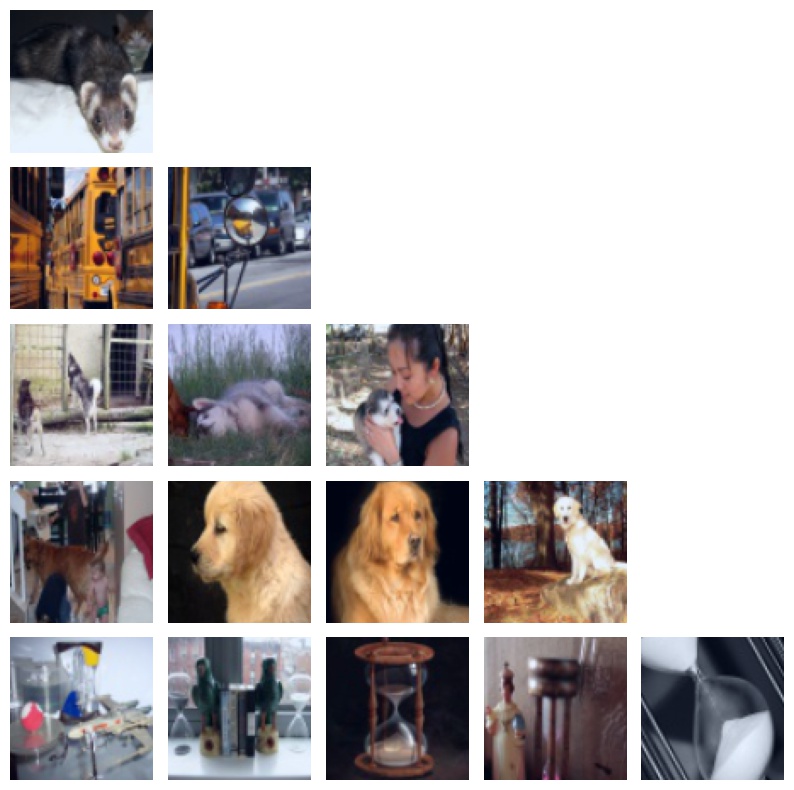}}
    \hspace{0.25cm}
    \fbox{\includegraphics[height=4.25cm]{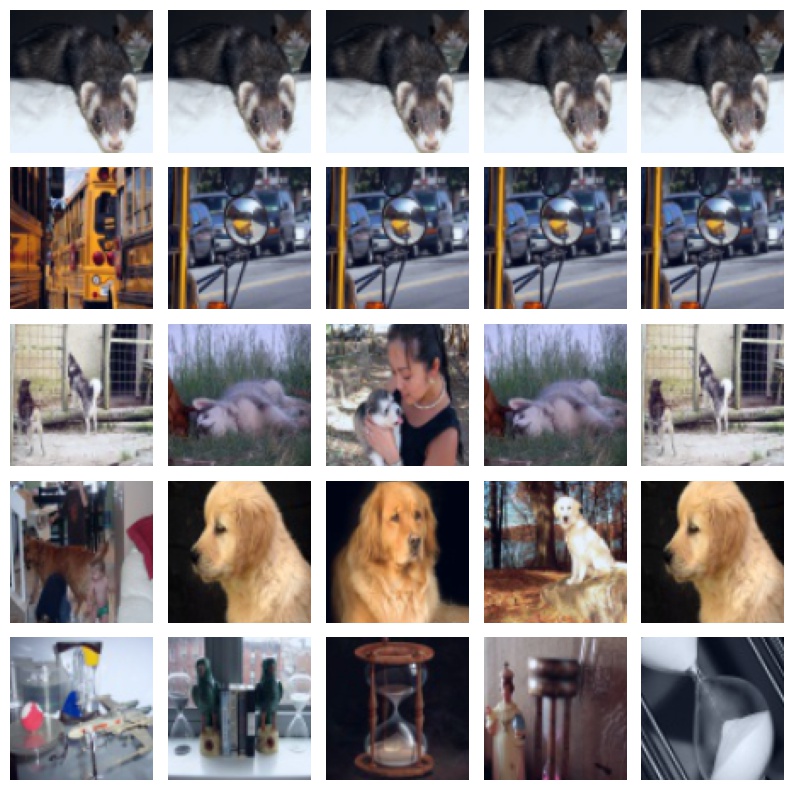}}
    \hspace{0.25cm}
    \fbox{\includegraphics[height=4.25cm]{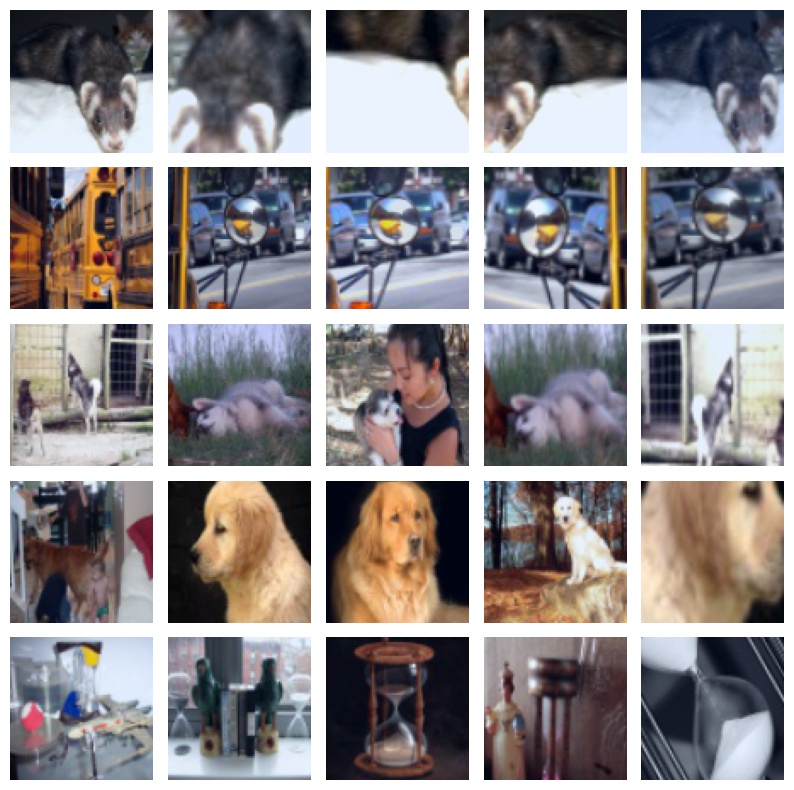}}
    \caption{Visualisation of linear 1-5shot support sets and Random Over-Sampling with (ROS+) or without (ROS) augmentations. \textbf{Left:} unbalanced case. \textbf{Middle:} rebalancing with ROS. \textbf{Right:} rebalancing with ROS+.}
    \label{fig ros_support}
\end{figure}

\subsection{Class Imbalance Techniques and Strategies}
We pair FSL methods with popular data-level class-imbalance strategies:
\begin{enumerate}
    \item \textbf{Random Over-Sampling (ROS)} \cite{Japkowicz2002imbalance} without and with data augmentation (\textbf{ROS} and \textbf{ROS+}). For the augmentations, we used: random-sized cropping between 0.15 and 1.1 scale of the image, random horizontal flip, and random color/brightness/contrast jitter. A visualization of ROS and ROS+ is presented in Figure~\ref{fig ros_support}. During meta-training, ROS+ augmentations were applied twice: once when sampling from the meta-training dataset and the second time during the support set resampling.
    \item \textbf{Random-Shot Meta-Training} \cite{Triantafillou2019meta, Lee2020baysiantaml, Chen2020mamlstop} was applied as specified in the main body of the paper (Section~\ref{appendix loss}).
    \item \textbf{Focal Loss} \cite{Lin2017focal}. Focal Loss has been found to be very effective in combating the class-imbalance problem on the one-stage object detectors. We exchanged the inner-loop cross-entropy loss of optimization-based algorithms and fine-tune baselines with the Focal Loss with $\gamma=2$ and $\alpha=1$. Results are presented in Figure~\ref{fig cost_func_mean} in the paper and in Appendix~\ref{appendix loss}.
    \item \textbf{Weighted Loss}. Weighted Loss is also commonly used to rebalance the effects of class-imbalance \cite{Buda2018imbalance,Leevy2018bigdata}. We weight the inner-loop cross-entropy loss of optimization-based algorithms and fine-tune baselines by inverse class frequency of support set samples. Results are presented in Figure~\ref{fig cost_func_mean} in the paper and in Appendix~\ref{appendix loss}.
    \item \textbf{CB Loss}. Class Balancing Loss (CB Loss) is a rebalancing loss multiplier that has been used effectively in the long-tail literature \cite{Cui2019cbloss}. The loss of each class is re-weighted based on the effective number of samples per class. Like other losses, we weight the inner-loop cross-entropy loss of optimization-based algorithms and the fine-tune baselines by a class multiplier. The class multiplier is calculated as $(1-\beta^n)/(1-\beta)$ where $n$ is the number of samples per class and $\beta=0.8$ based on $N-1/N$ formula, where $N$ is the number of classes. We also tried different values for $\beta \in \{0.1,0.2,0.3,0.4,0.5,0.6,0.7,0.8,0.9,0.99\}$ and we found that with $\beta~\to~1$ the performance gets closer to the weighted loss. In the main paper we report results with the standard value $\beta=N-1/N$ as suggested by the authors.
\end{enumerate}


\section{Verification of the Implementation}~\label{appendix verification}
\begin{figure}[htb]
    \centering
    \vspace{0.1cm}
    \includegraphics[width=0.32\linewidth]{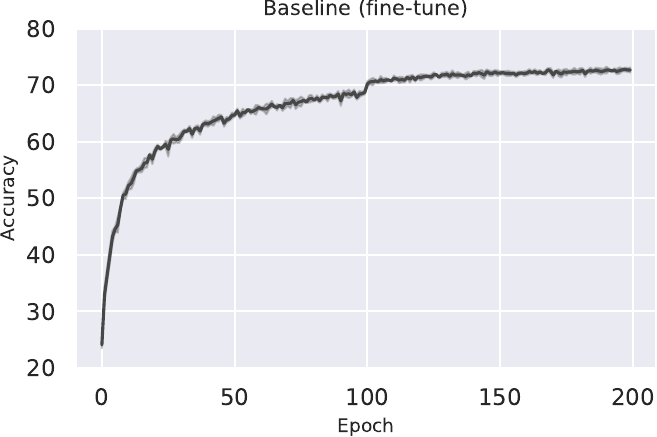}
    \vspace{0.1cm}
    \includegraphics[width=0.32\linewidth]{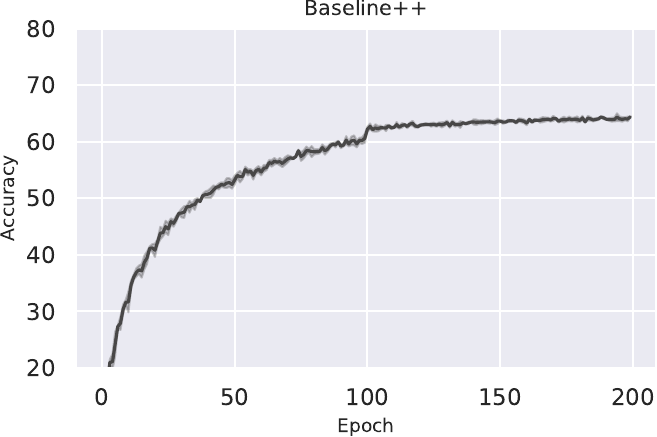}
    \vspace{0.1cm}
    \includegraphics[width=0.32\linewidth]{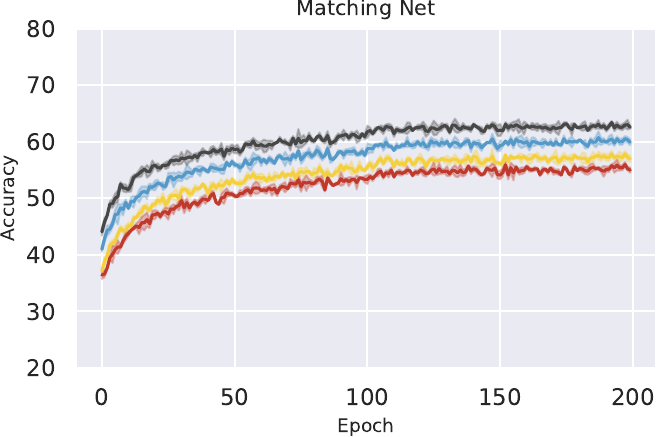}
    \vspace{0.1cm}
    \includegraphics[width=0.32\linewidth]{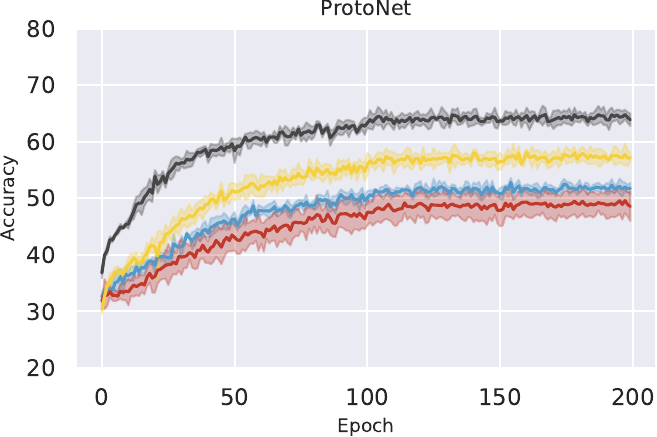}
    \vspace{0.1cm}
    \includegraphics[width=0.32\linewidth]{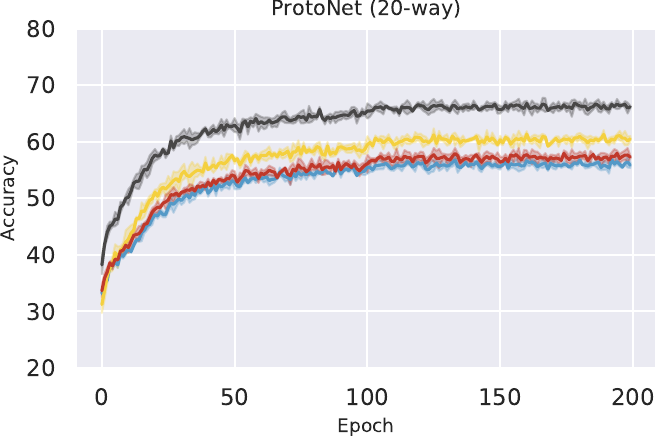}
    \vspace{0.1cm}
    \includegraphics[width=0.32\linewidth]{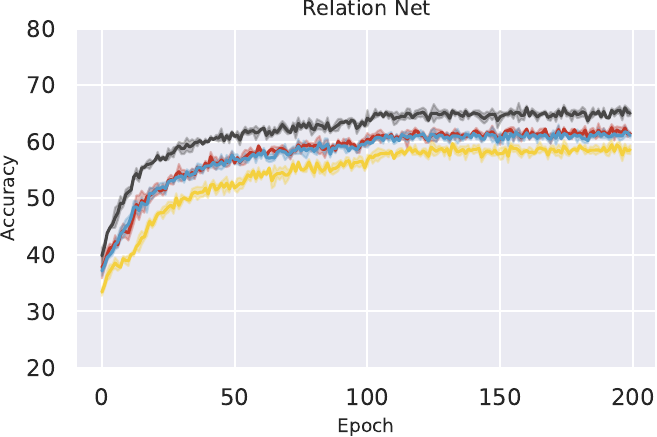}
    \vspace{0.1cm}
    \includegraphics[width=0.32\linewidth]{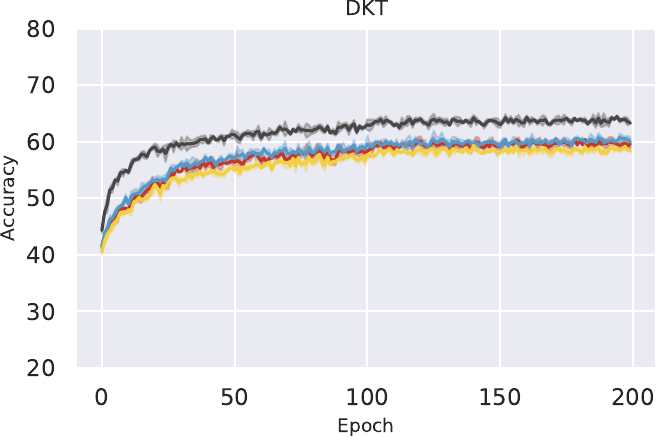}
    \vspace{0.1cm}
    \includegraphics[width=0.32\linewidth]{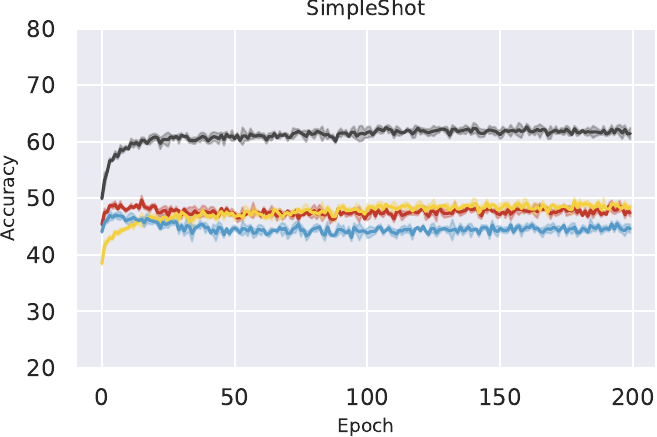}
    \vspace{0.1cm}
    \includegraphics[width=0.32\linewidth]{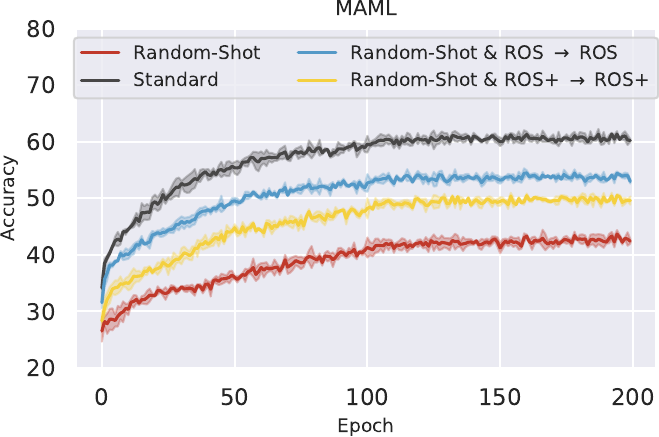}
    \vspace{0.1cm}
    \includegraphics[width=0.32\linewidth]{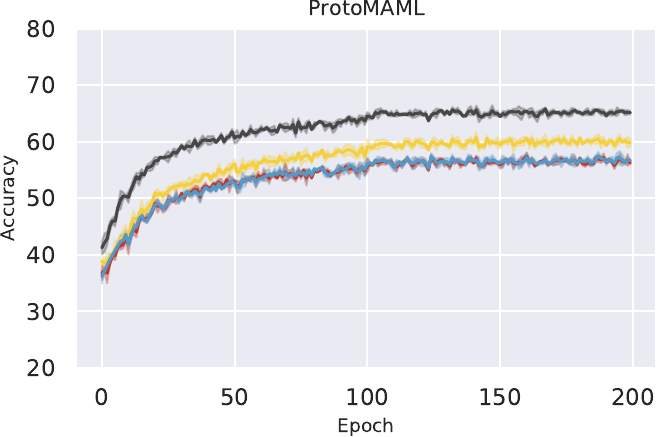}
    \vspace{0.1cm}
    \includegraphics[width=0.32\linewidth]{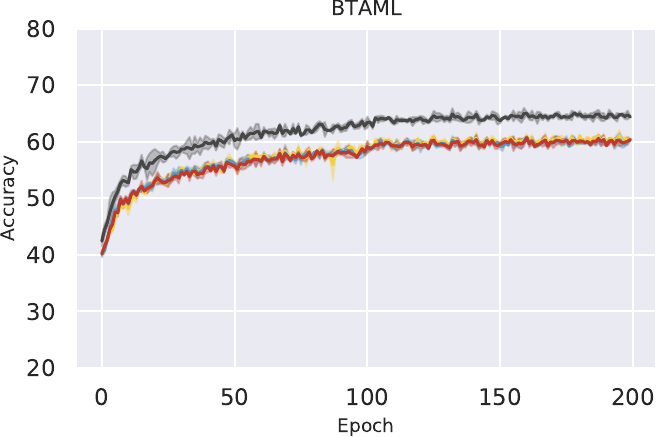}
    \vspace{0.1cm}
    \includegraphics[width=0.32\linewidth]{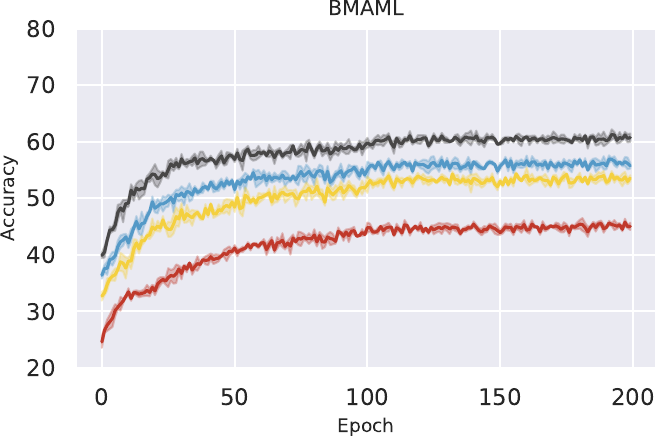}
    \caption{Validation performance through epochs of various meta-training strategies: Standard (black), Random-Shot (red), Random-Shot \& ROS (blue), and Random-Shot \& ROS+ (yellow). The shaded areas show $\pm 1$ standard deviation over three runs on different seeds.}
    \label{fig thr_epochs}
\end{figure}

We implement the FSL methods in PyTorch, adapting the implementation of \cite{Chen2019closer} but also borrowing from other implementations online (see individual method files in the source code for individual attribution). However, we have heavily modified these implementations to fit our imbalanced FSL framework, which also offers standard and continual FSL compatibility \cite{Antoniou2020cfsl}. We provide our implementations for ProtoMAML and BTAML, for which no open-source implementation in PyTorch existed at the time of writing. To verify our implementations, we compare methods on the standard balanced 5-shot 5-way task with reported accuracy. Results are presented in Table~\ref{tbl 5shot_5way}. All the algorithms achieve a score that is very close to the one reported in the original papers, with no less than 3\% accuracy points. Some of the discrepancies can be accounted to smaller training batches (e.g., SimpleShot), different augmentation strategies, and natural variance stemming from random initialization. The validation performances over epochs for each method in Figure~\ref{fig thr_epochs} suggest that models have sufficiently converged, and there would be little benefits from training on more tasks. Results in Appendix~\ref{appendix longer} show that training for five times as many episodes results in only $+1\%$ to $+3\%$ accuracy improvement. 

\begin{table}[htb]
    \caption[5-Shot 5-Way Classification]{Results of standard 5-shot 5-way experiments on Mini-ImageNet as achieved with our implementation compared to the original (reported) accuracy and other work. \emph{Other Sources} taken from: \ssymbol{1}~\cite{Chen2019closer}, \ssymbol{2}~\cite{Snell2020polya}, \ssymbol{3}~\cite{Vogelbaum2020protocontext}}\label{tbl 5shot_5way}
    \centering
    \scalebox{1}{\begin{tabular}{l|ccc}
\toprule
{} &                                       Acc (95\%CI) &                \specialcell{Original\\Acc(95\%CI)} &                      \specialcell{Other Sources\\Acc(95\%CI)} \\
\midrule
Baseline (fine-tune) \cite{Chen2019closer} &           62.67\textcolor{gray}{\tiny{$\pm$0.70}} &           62.53\textcolor{gray}{\tiny{$\pm$0.69}} &                                                             - \\
Baseline++      \cite{Chen2019closer}     &  \textbf{66.43\textcolor{gray}{\tiny{$\pm$0.66}}} &           66.43\textcolor{gray}{\tiny{$\pm$0.63}} &                                                             - \\
Matching Net    \cite{Vinyals2017matching}     &           62.27\textcolor{gray}{\tiny{$\pm$0.69}} &           55.31\textcolor{gray}{\tiny{$\pm$0.73}} &           63.48\textcolor{gray}{\tiny{$\pm$0.66}} \ssymbol{1} \\
ProtoNet     \cite{Snell2017proto}        &           64.37\textcolor{gray}{\tiny{$\pm$0.71}} &           65.77\textcolor{gray}{\tiny{$\pm$0.70}} &           64.24\textcolor{gray}{\tiny{$\pm$0.72}} \ssymbol{1} \\
ProtoNet (20-way) \cite{Snell2017proto}   &           65.76\textcolor{gray}{\tiny{$\pm$0.70}} &  \textbf{68.20\textcolor{gray}{\tiny{$\pm$0.66}}} &  \textbf{66.68\textcolor{gray}{\tiny{$\pm$0.68}} \ssymbol{1}} \\
Relation Net (CE) \cite{Sung2017relationnet}   &           64.76\textcolor{gray}{\tiny{$\pm$0.68}} &           65.32\textcolor{gray}{\tiny{$\pm$0.70}} &           66.60\textcolor{gray}{\tiny{$\pm$0.69}} \ssymbol{1} \\
DKT        \cite{Patacchiola2019gpshot}          &           62.92\textcolor{gray}{\tiny{$\pm$0.67}} &           64.00\textcolor{gray}{\tiny{$\pm$0.09}} &           62.88\textcolor{gray}{\tiny{$\pm$0.46}} \ssymbol{2} \\
SimpleShot      \cite{Wang2019simpleshot}     &           63.74\textcolor{gray}{\tiny{$\pm$0.69}} &           66.92\textcolor{gray}{\tiny{$\pm$0.17}} &                                                             - \\
MAML     \cite{Finn2017maml}            &           61.83\textcolor{gray}{\tiny{$\pm$0.71}} &           63.15\textcolor{gray}{\tiny{$\pm$0.91}} &           62.71\textcolor{gray}{\tiny{$\pm$0.71}} \ssymbol{1} \\
ProtoMAML     \cite{Triantafillou2019meta}       &           65.87\textcolor{gray}{\tiny{$\pm$0.71}} &                                                 - &           60.70\textcolor{gray}{\tiny{$\pm$0.99}} \ssymbol{3} \\
BMAML    \cite{Yoon2018bmaml}            &           59.89\textcolor{gray}{\tiny{$\pm$0.68}} &                                                 - &           59.23\textcolor{gray}{\tiny{$\pm$0.34}} \ssymbol{2} \\
BTAML   \cite{Lee2020baysiantaml}        &           64.96\textcolor{gray}{\tiny{$\pm$0.70}} &                                                 - &           - \\
\bottomrule
\end{tabular}

}
\end{table}


\clearpage

\section{More Details and Ablation Studies}\label{appendix ablation}
In this section, we present additional details and ablation studies. 

\subsection{Meta-Training with More Episodes}\label{appendix longer}
In this subsection, we meta-train MAML and ProtoNet five times longer than in the main experiments (500k meta-training tasks instead of 100k). The training procedure is similar to the one described in Section~\ref{appendix implementation}, with Conv4 backbone, and learning rate ($lr$) $1 \times 10^3$ decreased to $1 \times 10^{-4}$ halfway through training (i.e., after 250k tasks). Both $lr$ were selected based on best performance from the nine $lr$ combinations: ($1 \times 10^{-3}, 1 \times 10^{-4}$,$1 \times 10^{-4}, 1 \times 10^{-4}$,$1 \times 10^{-3}, 1 \times 10^{-3}$) where the second number indicates the learning rate after 125k, 200k, or 375k tasks. Table~\ref{tbl more_episodes} shows that the Standard meta-training continues to outperform Random-Shot mostly and does not improve much as the models are trained on more tasks. Only RelationNet and ProtoMAML see an advantage on the more imbalanced tasks, which is consistent with experiments trained on 100k episodes in the main paper. 

\begin{table}[htb]
    \caption{Accuracy of MAML and ProtoNet under different meta-training strategies after training on 500k tasks (instead of 100k as in the main experiments). The 95\% confidence interval is indicated in brackets. The results show that Standard episodic meta-training \cite{Vinyals2017matching} offers stronger performance than Random-Shot \cite{Triantafillou2019meta} on the imbalanced tasks. However, ProtoMAML and RelationNet see a slight advantage from Random-Shot on the 1-9shot random evaluation tasks.}\label{tbl more_episodes}
    \centering
    \scalebox{1}{
    \begin{tabular}{c|c|cccc}
\toprule
                  &          &                               \specialcell{5shot} &                     \specialcell{4-6shot\\linear} &                \specialcell{1-9shot\\step 1minor} &                     \specialcell{1-9shot\\random} \\
\midrule
\multirow{2}{*}{DKT} & Random-Shot &           63.24\textcolor{gray}{\tiny{$\pm$0.70}} &           63.14\textcolor{gray}{\tiny{$\pm$0.69}} &           61.31\textcolor{gray}{\tiny{$\pm$0.61}} &           59.19\textcolor{gray}{\tiny{$\pm$0.74}} \\
                  & Standard &  \textbf{64.08\textcolor{gray}{\tiny{$\pm$0.69}}} &  \textbf{63.60\textcolor{gray}{\tiny{$\pm$0.66}}} &  \textbf{62.02\textcolor{gray}{\tiny{$\pm$0.62}}} &  \textbf{60.04\textcolor{gray}{\tiny{$\pm$0.71}}} \\
\midrule
\multirow{2}{*}{MAML} & Random-Shot &           55.25\textcolor{gray}{\tiny{$\pm$0.75}} &           54.22\textcolor{gray}{\tiny{$\pm$0.74}} &           52.30\textcolor{gray}{\tiny{$\pm$0.65}} &           48.05\textcolor{gray}{\tiny{$\pm$0.78}} \\
                  & Standard &  \textbf{63.03\textcolor{gray}{\tiny{$\pm$0.74}}} &  \textbf{60.84\textcolor{gray}{\tiny{$\pm$0.67}}} &  \textbf{57.99\textcolor{gray}{\tiny{$\pm$0.59}}} &  \textbf{50.02\textcolor{gray}{\tiny{$\pm$0.88}}} \\
\midrule
\multirow{2}{*}{ProtoMAML} & Random-Shot &           62.84\textcolor{gray}{\tiny{$\pm$0.72}} &           62.13\textcolor{gray}{\tiny{$\pm$0.71}} &           59.97\textcolor{gray}{\tiny{$\pm$0.62}} &  \textbf{57.77\textcolor{gray}{\tiny{$\pm$0.75}}} \\
                  & Standard &  \textbf{67.09\textcolor{gray}{\tiny{$\pm$0.71}}} &  \textbf{65.12\textcolor{gray}{\tiny{$\pm$0.71}}} &  \textbf{60.80\textcolor{gray}{\tiny{$\pm$0.58}}} &           55.70\textcolor{gray}{\tiny{$\pm$0.83}} \\
\midrule
\multirow{2}{*}{ProtoNet} & Random-Shot &           58.01\textcolor{gray}{\tiny{$\pm$0.72}} &           57.80\textcolor{gray}{\tiny{$\pm$0.72}} &           55.96\textcolor{gray}{\tiny{$\pm$0.66}} &           54.81\textcolor{gray}{\tiny{$\pm$0.75}} \\
                  & Standard &  \textbf{67.53\textcolor{gray}{\tiny{$\pm$0.71}}} &  \textbf{66.14\textcolor{gray}{\tiny{$\pm$0.70}}} &  \textbf{61.45\textcolor{gray}{\tiny{$\pm$0.54}}} &  \textbf{56.54\textcolor{gray}{\tiny{$\pm$0.85}}} \\
\midrule
\multirow{2}{*}{Relation Net (CE)} & Random-Shot &           66.35\textcolor{gray}{\tiny{$\pm$0.71}} &           66.15\textcolor{gray}{\tiny{$\pm$0.71}} &  \textbf{65.90\textcolor{gray}{\tiny{$\pm$0.70}}} &  \textbf{63.96\textcolor{gray}{\tiny{$\pm$0.75}}} \\
                  & Standard &  \textbf{67.57\textcolor{gray}{\tiny{$\pm$0.70}}} &  \textbf{66.54\textcolor{gray}{\tiny{$\pm$0.70}}} &           62.28\textcolor{gray}{\tiny{$\pm$0.58}} &           60.34\textcolor{gray}{\tiny{$\pm$0.79}} \\
\bottomrule
\end{tabular}
    }
\end{table}

\subsection{Rebalancing Cost Functions}\label{appendix loss}
\begin{figure}[thb]
    \centering
    \includegraphics[width=0.99\linewidth]{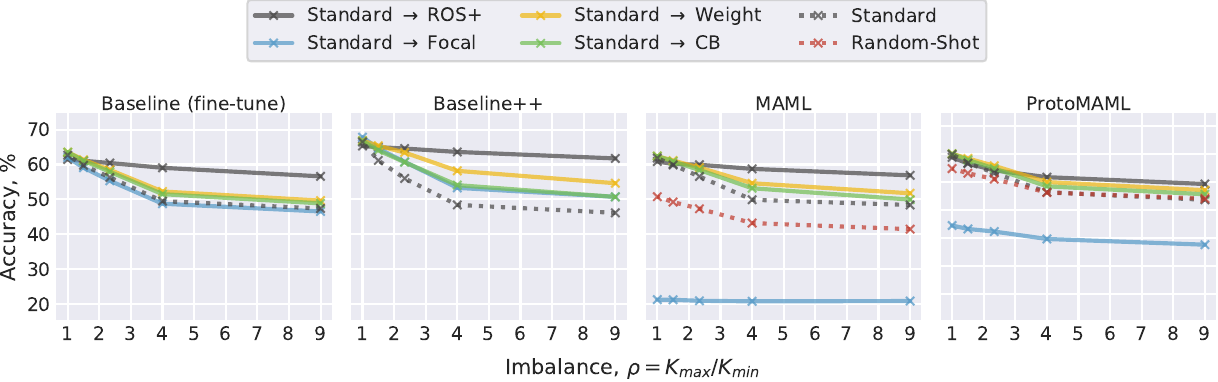}
    \caption{%
    Standard episodic meta-training \cite{Vinyals2017matching} and Random-Shot Meta-Training \cite{Triantafillou2019meta} with Weighted Loss \cite{Buda2018imbalance} and Focal Loss \cite{Lin2017focal}, applied to the inner-loop of optimization-based methods and fine-tune baselines.}
    \label{fig focal_loss}
\end{figure}

In Figure~\ref{fig focal_loss}, we compare Standard meta-training and Random-Shot Meta-Training with Focal Loss and Weighted Loss. Evaluation is performed on linear tasks constrained to 25 samples with increasing levels of imbalance. These results correspond to the per-model breakdown of Figure~\ref{fig cost_func_mean} in the paper. We observe no significant advantage of using Focal Loss Random-Shot Meta-Training over the Standard setting. Weighted Loss performs very similarly to ROS+.

\subsection{Linear Imbalance Experiments}\label{appendix linear}
\begin{figure}[tbh]
    \centering
    \includegraphics[width=0.99\linewidth]{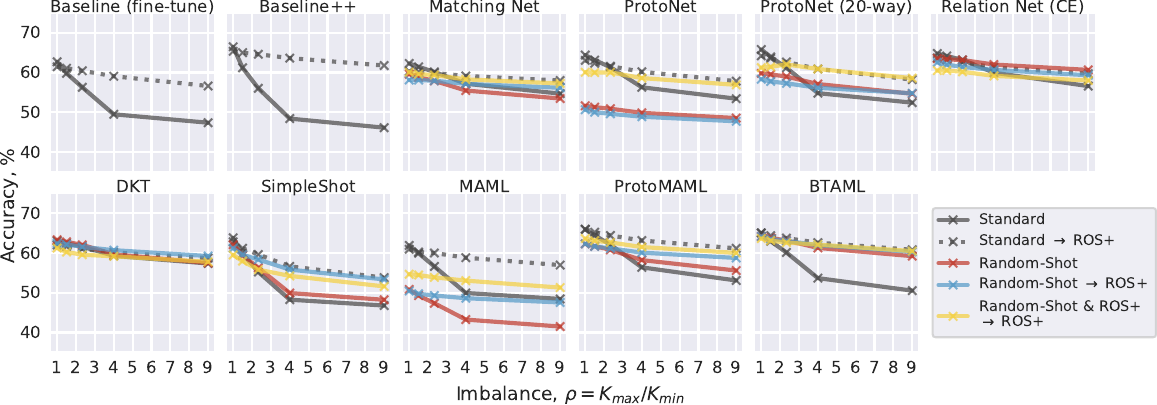}
    \caption{Standard episodic training \cite{Vinyals2017matching} vs. Random-Shot Meta-Training \cite{Triantafillou2019meta}, meta-tested on various linear imbalance tasks with a total support set size of 25 sampels. We used Random Over-Sampling with (ROS+) or without (ROS) augmentations.}
    \label{fig linear_imbalance_all}
\end{figure}

Figure~\ref{fig linear_imbalance_all} shows the performance of various methods over different levels of linear imbalance. When controlling for imbalance, the performances of many metric-based methods drop slightly as the imbalance increases (between $-3\%$ and $-8\%$). For optimization-based methods, the drop is more evident ($-15\%$ on average). The use of ROS/ROS+ significantly improves the performance. Note that Random-Shot Meta-Training on its own often has little or no advantage over Standard meta-training.

\subsection{Details into Backbone Experiments}\label{appendix backbone}
Tables~\ref{tab backbones_fsl_ret_acc} show the per-model break down of Figure~\ref{fig fsl_backbone}. The tabular results show random 1-9 shot 5-way performance on Conv6, ResNet10, and ResNet34. We can observe that ROS brings some benefits to most models (up to $+3\%$ gain on average). Note that we used the same hyperparameters of the four-layered CNN; additional gains could be obtained with an ad-hoc hyperparameter search for the larger backbones.

\begin{table}[ht]
    \begin{center}
        \caption{Accuracy on random 1-9shot evaluation after Standard and Random Shot meta-training (top), and Random Shot with ROS/ROS+ applied at both the meta-training and meta-testing time (bottom).}\label{tab backbones_fsl_ret_acc}
        \begin{tabular}{l|ccc|ccc}
\toprule
{} & \multicolumn{3}{c}{Standard} & \multicolumn{3}{c}{Random-Shot} \\
{} &                                             Conv6 &                                          ResNet10 &                                          ResNet34 &                                             Conv6 &                                          ResNet10 &                                          ResNet34 \\
\midrule
Baseline (1-NN)      &           51.21\textcolor{gray}{\tiny{$\pm$0.74}} &           53.18\textcolor{gray}{\tiny{$\pm$0.78}} &           53.25\textcolor{gray}{\tiny{$\pm$0.75}} &           50.01\textcolor{gray}{\tiny{$\pm$0.70}} &           52.24\textcolor{gray}{\tiny{$\pm$0.74}} &           52.49\textcolor{gray}{\tiny{$\pm$0.77}} \\
Baseline (fine-tune) &           54.12\textcolor{gray}{\tiny{$\pm$0.79}} &           58.55\textcolor{gray}{\tiny{$\pm$0.87}} &           59.02\textcolor{gray}{\tiny{$\pm$0.86}} &           54.62\textcolor{gray}{\tiny{$\pm$0.80}} &           58.36\textcolor{gray}{\tiny{$\pm$0.86}} &           60.54\textcolor{gray}{\tiny{$\pm$0.85}} \\
Baseline++           &           51.95\textcolor{gray}{\tiny{$\pm$0.83}} &           50.94\textcolor{gray}{\tiny{$\pm$0.84}} &           54.18\textcolor{gray}{\tiny{$\pm$0.86}} &           52.23\textcolor{gray}{\tiny{$\pm$0.83}} &           50.99\textcolor{gray}{\tiny{$\pm$0.83}} &           52.91\textcolor{gray}{\tiny{$\pm$0.80}} \\
Matching Net         &           54.99\textcolor{gray}{\tiny{$\pm$0.70}} &           58.60\textcolor{gray}{\tiny{$\pm$0.72}} &           60.00\textcolor{gray}{\tiny{$\pm$0.74}} &           53.70\textcolor{gray}{\tiny{$\pm$0.73}} &           56.54\textcolor{gray}{\tiny{$\pm$0.77}} &           56.34\textcolor{gray}{\tiny{$\pm$0.73}} \\
ProtoNet             &  \textbf{57.92\textcolor{gray}{\tiny{$\pm$0.78}}} &  \textbf{62.06\textcolor{gray}{\tiny{$\pm$0.74}}} &  \textbf{64.35\textcolor{gray}{\tiny{$\pm$0.74}}} &           56.31\textcolor{gray}{\tiny{$\pm$0.72}} &           60.00\textcolor{gray}{\tiny{$\pm$0.73}} &           59.81\textcolor{gray}{\tiny{$\pm$0.73}} \\
ProtoNet (20-way)    &           57.66\textcolor{gray}{\tiny{$\pm$0.76}} &           60.61\textcolor{gray}{\tiny{$\pm$0.80}} &           61.66\textcolor{gray}{\tiny{$\pm$0.80}} & \textbf{59.19\textcolor{gray}{\tiny{$\pm$0.74}}} &  \textbf{62.17\textcolor{gray}{\tiny{$\pm$0.79}}} & \textbf{62.46\textcolor{gray}{\tiny{$\pm$0.77}}} \\
DKT               &           55.60\textcolor{gray}{\tiny{$\pm$0.70}} &           57.98\textcolor{gray}{\tiny{$\pm$0.73}} &           57.07\textcolor{gray}{\tiny{$\pm$0.75}} &             57.44\textcolor{gray}{\tiny{$\pm$0.72}} &              58.77\textcolor{gray}{\tiny{$\pm$0.70}} &           57.83\textcolor{gray}{\tiny{$\pm$0.70}} \\
SimpleShot           &           54.36\textcolor{gray}{\tiny{$\pm$0.88}} &           60.95\textcolor{gray}{\tiny{$\pm$0.81}} &           60.48\textcolor{gray}{\tiny{$\pm$0.88}} &           54.90\textcolor{gray}{\tiny{$\pm$0.81}} &           61.33\textcolor{gray}{\tiny{$\pm$0.80}} &           55.25\textcolor{gray}{\tiny{$\pm$0.74}} \\
MAML                 &           52.53\textcolor{gray}{\tiny{$\pm$0.77}} &           57.69\textcolor{gray}{\tiny{$\pm$0.79}} &           52.52\textcolor{gray}{\tiny{$\pm$0.74}} &           46.91\textcolor{gray}{\tiny{$\pm$0.72}} &           52.61\textcolor{gray}{\tiny{$\pm$0.77}} &           48.83\textcolor{gray}{\tiny{$\pm$0.73}} \\
\bottomrule
\end{tabular}

        \begin{tabular}{l|ccc|ccc}
\toprule
{} & \multicolumn{3}{c}{Random-Shot \& ROS $\rightarrow$ ROS} & \multicolumn{3}{c}{Random-Shot \& ROS+ $\rightarrow$ ROS+} \\
{} &                                             Conv6 &                                          ResNet10 &                                          ResNet34 &                                             Conv6 &                                          ResNet10 &                                          ResNet34 \\
\midrule
Baseline (1-NN)      &           51.05\textcolor{gray}{\tiny{$\pm$0.75}} &           52.65\textcolor{gray}{\tiny{$\pm$0.75}} &           53.83\textcolor{gray}{\tiny{$\pm$0.74}} &           46.75\textcolor{gray}{\tiny{$\pm$0.79}} &           50.69\textcolor{gray}{\tiny{$\pm$0.76}} &           52.06\textcolor{gray}{\tiny{$\pm$0.76}}	 \\
Baseline (fine-tune) &           55.32\textcolor{gray}{\tiny{$\pm$0.80}} &           59.77\textcolor{gray}{\tiny{$\pm$0.88}} &           61.51\textcolor{gray}{\tiny{$\pm$0.83}} &           54.48\textcolor{gray}{\tiny{$\pm$0.77}} &           59.02\textcolor{gray}{\tiny{$\pm$0.80}} &           60.48\textcolor{gray}{\tiny{$\pm$0.80}} \\
Baseline++           &  \textbf{61.08\textcolor{gray}{\tiny{$\pm$0.75}}} &           57.41\textcolor{gray}{\tiny{$\pm$0.74}} &           58.07\textcolor{gray}{\tiny{$\pm$0.75}} &           56.95\textcolor{gray}{\tiny{$\pm$0.78}} &           54.81\textcolor{gray}{\tiny{$\pm$0.77}} &           55.49\textcolor{gray}{\tiny{$\pm$0.81}} \\
Matching Net         &           58.32\textcolor{gray}{\tiny{$\pm$0.72}} &  \textbf{60.48\textcolor{gray}{\tiny{$\pm$0.72}}} &           61.22\textcolor{gray}{\tiny{$\pm$0.74}} &           56.24\textcolor{gray}{\tiny{$\pm$0.74}} &           59.89\textcolor{gray}{\tiny{$\pm$0.70}} &           55.19\textcolor{gray}{\tiny{$\pm$0.71}} \\
ProtoNet             &           56.99\textcolor{gray}{\tiny{$\pm$0.79}} &           60.47\textcolor{gray}{\tiny{$\pm$0.73}} &           60.70\textcolor{gray}{\tiny{$\pm$0.76}} &           58.60\textcolor{gray}{\tiny{$\pm$0.72}} &  \textbf{62.05\textcolor{gray}{\tiny{$\pm$0.76}}} &           \textbf{63.29\textcolor{gray}{\tiny{$\pm$0.75}}}\\
ProtoNet (20-way)    &           57.66\textcolor{gray}{\tiny{$\pm$0.77}} &           59.85\textcolor{gray}{\tiny{$\pm$0.73}} &  \textbf{61.60\textcolor{gray}{\tiny{$\pm$0.73}}} &  \textbf{58.82\textcolor{gray}{\tiny{$\pm$0.72}}} &           61.15\textcolor{gray}{\tiny{$\pm$0.76}} &           61.78\textcolor{gray}{\tiny{$\pm$0.78}} \\
DKT               &           57.01\textcolor{gray}{\tiny{$\pm$0.72}} &           59.12\textcolor{gray}{\tiny{$\pm$0.71}} &           39.76\textcolor{gray}{\tiny{$\pm$0.62}}	 &           56.36\textcolor{gray}{\tiny{$\pm$0.73}} &           58.09\textcolor{gray}{\tiny{$\pm$0.73}} &           58.09\textcolor{gray}{\tiny{$\pm$0.72}} \\
SimpleShot           &           52.99\textcolor{gray}{\tiny{$\pm$0.82}} &           60.46\textcolor{gray}{\tiny{$\pm$0.80}} &           59.99\textcolor{gray}{\tiny{$\pm$0.81}} &           51.44\textcolor{gray}{\tiny{$\pm$0.79}} &           54.20\textcolor{gray}{\tiny{$\pm$0.83}} &           37.14\textcolor{gray}{\tiny{$\pm$0.69}} \\
MAML                 &           56.25\textcolor{gray}{\tiny{$\pm$0.74}} &           59.72\textcolor{gray}{\tiny{$\pm$0.76}} &           58.30\textcolor{gray}{\tiny{$\pm$0.76}} &           50.27\textcolor{gray}{\tiny{$\pm$0.78}} &           46.76\textcolor{gray}{\tiny{$\pm$0.74}} &           42.21\textcolor{gray}{\tiny{$\pm$0.74}}  \\
\bottomrule
\end{tabular}

    \end{center}
\end{table}

\clearpage

\subsection{Precision and Recall Tables}\label{sec pr_tables}
Tables~\ref{tbl pr_tables} show the average precision and recall on 1-9shot linear tasks. From these results, we can gain some additional insights. For instance, DKT~\cite{Patacchiola2019gpshot} offers strong performance in classes with a small number of shots and well-balanced performances for higher shots. This may be due to the partitioned Bayesian one-vs-rest scheme used for classification by DKT, with a separate Gaussian Process for each class; this could be more robust to imbalance. BMAML, on the other hand, fails to correctly classify samples with $K=1$ and $K=3$ samples, showing that the method has a strong bias towards majority classes.

\begin{table}[h!]
    \caption{Precision and recall for linear 1-9shot 5-way tasks after (from top to bottom): (i) \textbf{Standard} meta-training, (ii) \textbf{Standard} meta-training with \textbf{ROS+} applied at meta-test time, (iii) \textbf{Random-Shot} meta-training, (iv) \textbf{Random-Shot} meta-training with \textbf{ROS+} applied at meta-test time, and (v) \textbf{Random-Shot} meta-training with \textbf{ROS+} applied at both meta-train and meta-test time.}\label{tbl pr_tables}
    \centering
    \scalebox{0.9}{
    \vspace{0.5mm}
    \begin{tabular}{cl|ccccc|ccccc}
\toprule
 & {} & \multicolumn{5}{c}{Precision (95\%CI)} & \multicolumn{5}{c}{Recall (95\%CI)} \\
 & {} &                                          $K_0=1$ &                                          $K_1=3$ &                                          $K_2=5$ &                                          $K_3=7$ &                                          $K_4=9$ &                                          $K_0=1$ &                                          $K_1=3$ &                                          $K_2=5$ &                                          $K_3=7$ &                                          $K_4=9$ \\
\midrule
 \parbox[t]{2mm}{\multirow{11}{*}{\rotatebox[origin=c]{90}{\specialcell{Standard}}}} & Baseline (fine-tune) &           0.05\textcolor{gray}{\tiny{$\pm$0.02}} &           0.63\textcolor{gray}{\tiny{$\pm$0.03}} &           0.63\textcolor{gray}{\tiny{$\pm$0.02}} &           0.50\textcolor{gray}{\tiny{$\pm$0.01}} &           0.41\textcolor{gray}{\tiny{$\pm$0.01}} &           0.00\textcolor{gray}{\tiny{$\pm$0.00}} &           0.19\textcolor{gray}{\tiny{$\pm$0.02}} &           0.53\textcolor{gray}{\tiny{$\pm$0.02}} &           0.75\textcolor{gray}{\tiny{$\pm$0.01}} &           0.90\textcolor{gray}{\tiny{$\pm$0.01}} \\
 & Baseline++           &           0.02\textcolor{gray}{\tiny{$\pm$0.01}} &           0.45\textcolor{gray}{\tiny{$\pm$0.04}} &  \textbf{0.66\textcolor{gray}{\tiny{$\pm$0.02}}} &           0.52\textcolor{gray}{\tiny{$\pm$0.01}} &           0.39\textcolor{gray}{\tiny{$\pm$0.01}} &           0.00\textcolor{gray}{\tiny{$\pm$0.00}} &           0.13\textcolor{gray}{\tiny{$\pm$0.02}} &           0.48\textcolor{gray}{\tiny{$\pm$0.02}} &           0.76\textcolor{gray}{\tiny{$\pm$0.02}} &  \textbf{0.94\textcolor{gray}{\tiny{$\pm$0.01}}} \\
 & Matching Net         &           0.54\textcolor{gray}{\tiny{$\pm$0.04}} &           0.67\textcolor{gray}{\tiny{$\pm$0.02}} &           0.62\textcolor{gray}{\tiny{$\pm$0.01}} &           0.55\textcolor{gray}{\tiny{$\pm$0.01}} &           0.48\textcolor{gray}{\tiny{$\pm$0.01}} &           0.15\textcolor{gray}{\tiny{$\pm$0.02}} &           0.41\textcolor{gray}{\tiny{$\pm$0.02}} &           0.61\textcolor{gray}{\tiny{$\pm$0.02}} &           0.73\textcolor{gray}{\tiny{$\pm$0.02}} &           0.84\textcolor{gray}{\tiny{$\pm$0.01}} \\
 & ProtoNet             &           0.18\textcolor{gray}{\tiny{$\pm$0.03}} &           0.70\textcolor{gray}{\tiny{$\pm$0.02}} &           0.62\textcolor{gray}{\tiny{$\pm$0.02}} &           0.54\textcolor{gray}{\tiny{$\pm$0.01}} &           0.51\textcolor{gray}{\tiny{$\pm$0.01}} &           0.03\textcolor{gray}{\tiny{$\pm$0.01}} &           0.37\textcolor{gray}{\tiny{$\pm$0.02}} &           0.65\textcolor{gray}{\tiny{$\pm$0.02}} &           0.78\textcolor{gray}{\tiny{$\pm$0.01}} &           0.84\textcolor{gray}{\tiny{$\pm$0.01}} \\
 & ProtoNet (20-way)    &           0.10\textcolor{gray}{\tiny{$\pm$0.02}} &  \textbf{0.71\textcolor{gray}{\tiny{$\pm$0.03}}} &           0.64\textcolor{gray}{\tiny{$\pm$0.02}} &           0.55\textcolor{gray}{\tiny{$\pm$0.01}} &           0.49\textcolor{gray}{\tiny{$\pm$0.01}} &           0.01\textcolor{gray}{\tiny{$\pm$0.00}} &           0.31\textcolor{gray}{\tiny{$\pm$0.02}} &           0.63\textcolor{gray}{\tiny{$\pm$0.02}} &  \textbf{0.79\textcolor{gray}{\tiny{$\pm$0.01}}} &           0.88\textcolor{gray}{\tiny{$\pm$0.01}} \\
 & Relation Net (CE)    &           0.42\textcolor{gray}{\tiny{$\pm$0.04}} &           0.67\textcolor{gray}{\tiny{$\pm$0.02}} &           0.62\textcolor{gray}{\tiny{$\pm$0.01}} &           0.56\textcolor{gray}{\tiny{$\pm$0.01}} &           0.54\textcolor{gray}{\tiny{$\pm$0.01}} &           0.09\textcolor{gray}{\tiny{$\pm$0.01}} &  \textbf{0.48\textcolor{gray}{\tiny{$\pm$0.02}}} &  \textbf{0.68\textcolor{gray}{\tiny{$\pm$0.02}}} &           0.76\textcolor{gray}{\tiny{$\pm$0.01}} &           0.81\textcolor{gray}{\tiny{$\pm$0.01}} \\
 & DKT                  &  \textbf{0.57\textcolor{gray}{\tiny{$\pm$0.03}}} &           0.63\textcolor{gray}{\tiny{$\pm$0.02}} &           0.61\textcolor{gray}{\tiny{$\pm$0.01}} &  \textbf{0.57\textcolor{gray}{\tiny{$\pm$0.01}}} &  \textbf{0.55\textcolor{gray}{\tiny{$\pm$0.01}}} &  \textbf{0.22\textcolor{gray}{\tiny{$\pm$0.02}}} &           0.48\textcolor{gray}{\tiny{$\pm$0.02}} &           0.65\textcolor{gray}{\tiny{$\pm$0.02}} &           0.73\textcolor{gray}{\tiny{$\pm$0.02}} &           0.79\textcolor{gray}{\tiny{$\pm$0.01}} \\
 & SimpleShot           &           0.01\textcolor{gray}{\tiny{$\pm$0.01}} &           0.59\textcolor{gray}{\tiny{$\pm$0.04}} &           0.66\textcolor{gray}{\tiny{$\pm$0.02}} &           0.51\textcolor{gray}{\tiny{$\pm$0.01}} &           0.41\textcolor{gray}{\tiny{$\pm$0.01}} &           0.00\textcolor{gray}{\tiny{$\pm$0.00}} &           0.15\textcolor{gray}{\tiny{$\pm$0.02}} &           0.52\textcolor{gray}{\tiny{$\pm$0.02}} &           0.77\textcolor{gray}{\tiny{$\pm$0.01}} &           0.90\textcolor{gray}{\tiny{$\pm$0.01}} \\
 & MAML                 &           0.00\textcolor{gray}{\tiny{$\pm$0.00}} &           0.65\textcolor{gray}{\tiny{$\pm$0.02}} &           0.58\textcolor{gray}{\tiny{$\pm$0.02}} &           0.48\textcolor{gray}{\tiny{$\pm$0.01}} &           0.42\textcolor{gray}{\tiny{$\pm$0.01}} &           0.00\textcolor{gray}{\tiny{$\pm$0.00}} &           0.26\textcolor{gray}{\tiny{$\pm$0.02}} &           0.55\textcolor{gray}{\tiny{$\pm$0.02}} &           0.75\textcolor{gray}{\tiny{$\pm$0.01}} &           0.85\textcolor{gray}{\tiny{$\pm$0.01}} \\
 & ProtoMAML            &           0.28\textcolor{gray}{\tiny{$\pm$0.04}} &           0.71\textcolor{gray}{\tiny{$\pm$0.02}} &           0.63\textcolor{gray}{\tiny{$\pm$0.01}} &           0.53\textcolor{gray}{\tiny{$\pm$0.01}} &           0.45\textcolor{gray}{\tiny{$\pm$0.01}} &           0.04\textcolor{gray}{\tiny{$\pm$0.01}} &           0.34\textcolor{gray}{\tiny{$\pm$0.02}} &           0.62\textcolor{gray}{\tiny{$\pm$0.02}} &           0.78\textcolor{gray}{\tiny{$\pm$0.01}} &           0.87\textcolor{gray}{\tiny{$\pm$0.01}} \\
 & BTAML                &           0.11\textcolor{gray}{\tiny{$\pm$0.03}} &           0.69\textcolor{gray}{\tiny{$\pm$0.03}} &           0.61\textcolor{gray}{\tiny{$\pm$0.02}} &           0.51\textcolor{gray}{\tiny{$\pm$0.01}} &           0.43\textcolor{gray}{\tiny{$\pm$0.01}} &           0.01\textcolor{gray}{\tiny{$\pm$0.00}} &           0.29\textcolor{gray}{\tiny{$\pm$0.02}} &           0.58\textcolor{gray}{\tiny{$\pm$0.02}} &           0.77\textcolor{gray}{\tiny{$\pm$0.01}} &           0.88\textcolor{gray}{\tiny{$\pm$0.01}} \\
\bottomrule
\end{tabular}
    }
    \scalebox{0.9}{
    \vspace{0.5mm}
    \begin{tabular}{cl|ccccc|ccccc}
\toprule
 & {} & \multicolumn{5}{c}{Precision (95\%CI)} & \multicolumn{5}{c}{Recall (95\%CI)} \\
 & {} &                                          $K_0=1$ &                                          $K_1=3$ &                                          $K_2=5$ &                                          $K_3=7$ &                                          $K_4=9$ &                                          $K_0=1$ &                                          $K_1=3$ &                                          $K_2=5$ &                                          $K_3=7$ &                                          $K_4=9$ \\
\midrule
 \parbox[t]{2mm}{\multirow{11}{*}{\rotatebox[origin=c]{90}{\specialcell{Standard $\rightarrow$ ROS+}}}} & Baseline (fine-tune) &           0.61\textcolor{gray}{\tiny{$\pm$0.03}} &           0.62\textcolor{gray}{\tiny{$\pm$0.02}} &           0.60\textcolor{gray}{\tiny{$\pm$0.01}} &           0.57\textcolor{gray}{\tiny{$\pm$0.01}} &           0.55\textcolor{gray}{\tiny{$\pm$0.01}} &           0.22\textcolor{gray}{\tiny{$\pm$0.02}} &           0.50\textcolor{gray}{\tiny{$\pm$0.02}} &           0.64\textcolor{gray}{\tiny{$\pm$0.02}} &           0.71\textcolor{gray}{\tiny{$\pm$0.01}} &           0.75\textcolor{gray}{\tiny{$\pm$0.01}} \\
 & Baseline++           &  \textbf{0.65\textcolor{gray}{\tiny{$\pm$0.02}}} &  \textbf{0.66\textcolor{gray}{\tiny{$\pm$0.01}}} &  \textbf{0.65\textcolor{gray}{\tiny{$\pm$0.01}}} &  \textbf{0.62\textcolor{gray}{\tiny{$\pm$0.01}}} &  \textbf{0.60\textcolor{gray}{\tiny{$\pm$0.01}}} &           0.35\textcolor{gray}{\tiny{$\pm$0.02}} &  \textbf{0.58\textcolor{gray}{\tiny{$\pm$0.02}}} &  \textbf{0.67\textcolor{gray}{\tiny{$\pm$0.02}}} &           0.73\textcolor{gray}{\tiny{$\pm$0.02}} &           0.76\textcolor{gray}{\tiny{$\pm$0.01}} \\
 & Matching Net         &           0.58\textcolor{gray}{\tiny{$\pm$0.02}} &           0.61\textcolor{gray}{\tiny{$\pm$0.02}} &           0.61\textcolor{gray}{\tiny{$\pm$0.01}} &           0.58\textcolor{gray}{\tiny{$\pm$0.01}} &           0.57\textcolor{gray}{\tiny{$\pm$0.01}} &  \textbf{0.40\textcolor{gray}{\tiny{$\pm$0.02}}} &           0.53\textcolor{gray}{\tiny{$\pm$0.02}} &           0.61\textcolor{gray}{\tiny{$\pm$0.02}} &           0.67\textcolor{gray}{\tiny{$\pm$0.02}} &           0.68\textcolor{gray}{\tiny{$\pm$0.02}} \\
 & ProtoNet             &           0.63\textcolor{gray}{\tiny{$\pm$0.03}} &           0.64\textcolor{gray}{\tiny{$\pm$0.02}} &           0.61\textcolor{gray}{\tiny{$\pm$0.01}} &           0.58\textcolor{gray}{\tiny{$\pm$0.01}} &           0.57\textcolor{gray}{\tiny{$\pm$0.01}} &           0.23\textcolor{gray}{\tiny{$\pm$0.02}} &           0.53\textcolor{gray}{\tiny{$\pm$0.02}} &           0.66\textcolor{gray}{\tiny{$\pm$0.02}} &           0.72\textcolor{gray}{\tiny{$\pm$0.01}} &           0.75\textcolor{gray}{\tiny{$\pm$0.01}} \\
 & ProtoNet (20-way)    &           0.62\textcolor{gray}{\tiny{$\pm$0.03}} &           0.66\textcolor{gray}{\tiny{$\pm$0.02}} &           0.62\textcolor{gray}{\tiny{$\pm$0.01}} &           0.59\textcolor{gray}{\tiny{$\pm$0.01}} &           0.57\textcolor{gray}{\tiny{$\pm$0.01}} &           0.21\textcolor{gray}{\tiny{$\pm$0.02}} &           0.52\textcolor{gray}{\tiny{$\pm$0.02}} &           0.66\textcolor{gray}{\tiny{$\pm$0.02}} &  \textbf{0.74\textcolor{gray}{\tiny{$\pm$0.01}}} &           0.78\textcolor{gray}{\tiny{$\pm$0.01}} \\
 & Relation Net (CE)    &           0.62\textcolor{gray}{\tiny{$\pm$0.02}} &           0.63\textcolor{gray}{\tiny{$\pm$0.02}} &           0.63\textcolor{gray}{\tiny{$\pm$0.01}} &           0.60\textcolor{gray}{\tiny{$\pm$0.01}} &           0.59\textcolor{gray}{\tiny{$\pm$0.01}} &           0.35\textcolor{gray}{\tiny{$\pm$0.02}} &           0.56\textcolor{gray}{\tiny{$\pm$0.02}} &           0.66\textcolor{gray}{\tiny{$\pm$0.02}} &           0.69\textcolor{gray}{\tiny{$\pm$0.02}} &           0.73\textcolor{gray}{\tiny{$\pm$0.01}} \\
 & DKT                  &           0.58\textcolor{gray}{\tiny{$\pm$0.02}} &           0.61\textcolor{gray}{\tiny{$\pm$0.02}} &           0.62\textcolor{gray}{\tiny{$\pm$0.01}} &           0.59\textcolor{gray}{\tiny{$\pm$0.01}} &           0.59\textcolor{gray}{\tiny{$\pm$0.01}} &           0.37\textcolor{gray}{\tiny{$\pm$0.02}} &           0.55\textcolor{gray}{\tiny{$\pm$0.02}} &           0.63\textcolor{gray}{\tiny{$\pm$0.02}} &           0.68\textcolor{gray}{\tiny{$\pm$0.02}} &           0.70\textcolor{gray}{\tiny{$\pm$0.02}} \\
 & SimpleShot           &           0.52\textcolor{gray}{\tiny{$\pm$0.04}} &           0.66\textcolor{gray}{\tiny{$\pm$0.02}} &           0.61\textcolor{gray}{\tiny{$\pm$0.02}} &           0.55\textcolor{gray}{\tiny{$\pm$0.01}} &           0.51\textcolor{gray}{\tiny{$\pm$0.01}} &           0.11\textcolor{gray}{\tiny{$\pm$0.01}} &           0.41\textcolor{gray}{\tiny{$\pm$0.02}} &           0.62\textcolor{gray}{\tiny{$\pm$0.02}} &           0.74\textcolor{gray}{\tiny{$\pm$0.01}} &  \textbf{0.81\textcolor{gray}{\tiny{$\pm$0.01}}} \\
 & MAML                 &           0.58\textcolor{gray}{\tiny{$\pm$0.02}} &           0.57\textcolor{gray}{\tiny{$\pm$0.01}} &           0.58\textcolor{gray}{\tiny{$\pm$0.01}} &           0.58\textcolor{gray}{\tiny{$\pm$0.01}} &           0.57\textcolor{gray}{\tiny{$\pm$0.01}} &           0.34\textcolor{gray}{\tiny{$\pm$0.02}} &           0.55\textcolor{gray}{\tiny{$\pm$0.02}} &           0.62\textcolor{gray}{\tiny{$\pm$0.02}} &           0.65\textcolor{gray}{\tiny{$\pm$0.01}} &           0.67\textcolor{gray}{\tiny{$\pm$0.01}} \\
 & ProtoMAML            &           0.65\textcolor{gray}{\tiny{$\pm$0.02}} &           0.64\textcolor{gray}{\tiny{$\pm$0.02}} &           0.63\textcolor{gray}{\tiny{$\pm$0.01}} &           0.61\textcolor{gray}{\tiny{$\pm$0.01}} &           0.60\textcolor{gray}{\tiny{$\pm$0.01}} &           0.34\textcolor{gray}{\tiny{$\pm$0.02}} &           0.57\textcolor{gray}{\tiny{$\pm$0.02}} &           0.67\textcolor{gray}{\tiny{$\pm$0.02}} &           0.72\textcolor{gray}{\tiny{$\pm$0.01}} &           0.75\textcolor{gray}{\tiny{$\pm$0.01}} \\
 & BTAML                &           0.62\textcolor{gray}{\tiny{$\pm$0.02}} &           0.62\textcolor{gray}{\tiny{$\pm$0.01}} &           0.62\textcolor{gray}{\tiny{$\pm$0.01}} &           0.61\textcolor{gray}{\tiny{$\pm$0.01}} &           0.60\textcolor{gray}{\tiny{$\pm$0.01}} &           0.38\textcolor{gray}{\tiny{$\pm$0.02}} &           0.58\textcolor{gray}{\tiny{$\pm$0.02}} &           0.67\textcolor{gray}{\tiny{$\pm$0.02}} &           0.70\textcolor{gray}{\tiny{$\pm$0.01}} &           0.72\textcolor{gray}{\tiny{$\pm$0.01}} \\
\bottomrule
\end{tabular}
    }
    \scalebox{0.9}{
    \vspace{0.5mm}
    \begin{tabular}{cl|ccccc|ccccc}
\toprule
 & {} & \multicolumn{5}{c}{Precision (95\%CI)} & \multicolumn{5}{c}{Recall (95\%CI)} \\
 & {} &                                          $K_0=1$ &                                          $K_1=3$ &                                          $K_2=5$ &                                          $K_3=7$ &                                          $K_4=9$ &                                          $K_0=1$ &                                          $K_1=3$ &                                          $K_2=5$ &                                          $K_3=7$ &                                          $K_4=9$ \\
\midrule
 \parbox[t]{2mm}{\multirow{9}{*}{\rotatebox[origin=c]{90}{\specialcell{Random-Shot}}}} & Matching Net      &           0.58\textcolor{gray}{\tiny{$\pm$0.03}} &  \textbf{0.65\textcolor{gray}{\tiny{$\pm$0.02}}} &           0.62\textcolor{gray}{\tiny{$\pm$0.02}} &           0.55\textcolor{gray}{\tiny{$\pm$0.01}} &           0.45\textcolor{gray}{\tiny{$\pm$0.01}} &           0.24\textcolor{gray}{\tiny{$\pm$0.02}} &           0.36\textcolor{gray}{\tiny{$\pm$0.02}} &           0.55\textcolor{gray}{\tiny{$\pm$0.02}} &           0.68\textcolor{gray}{\tiny{$\pm$0.02}} &           0.84\textcolor{gray}{\tiny{$\pm$0.01}} \\
 & ProtoNet          &           0.47\textcolor{gray}{\tiny{$\pm$0.03}} &           0.51\textcolor{gray}{\tiny{$\pm$0.02}} &           0.51\textcolor{gray}{\tiny{$\pm$0.01}} &           0.49\textcolor{gray}{\tiny{$\pm$0.01}} &           0.48\textcolor{gray}{\tiny{$\pm$0.01}} &           0.17\textcolor{gray}{\tiny{$\pm$0.02}} &           0.44\textcolor{gray}{\tiny{$\pm$0.02}} &           0.57\textcolor{gray}{\tiny{$\pm$0.02}} &           0.61\textcolor{gray}{\tiny{$\pm$0.02}} &           0.63\textcolor{gray}{\tiny{$\pm$0.02}} \\
 & ProtoNet (20-way) &           0.52\textcolor{gray}{\tiny{$\pm$0.03}} &           0.60\textcolor{gray}{\tiny{$\pm$0.02}} &           0.58\textcolor{gray}{\tiny{$\pm$0.01}} &           0.55\textcolor{gray}{\tiny{$\pm$0.01}} &           0.54\textcolor{gray}{\tiny{$\pm$0.01}} &           0.16\textcolor{gray}{\tiny{$\pm$0.02}} &           0.50\textcolor{gray}{\tiny{$\pm$0.02}} &  \textbf{0.65\textcolor{gray}{\tiny{$\pm$0.02}}} &           0.70\textcolor{gray}{\tiny{$\pm$0.01}} &           0.73\textcolor{gray}{\tiny{$\pm$0.01}} \\
 & Relation Net (CE) &           0.55\textcolor{gray}{\tiny{$\pm$0.02}} &           0.62\textcolor{gray}{\tiny{$\pm$0.01}} &  \textbf{0.64\textcolor{gray}{\tiny{$\pm$0.01}}} &  \textbf{0.63\textcolor{gray}{\tiny{$\pm$0.01}}} &  \textbf{0.64\textcolor{gray}{\tiny{$\pm$0.01}}} &  \textbf{0.48\textcolor{gray}{\tiny{$\pm$0.02}}} &  \textbf{0.58\textcolor{gray}{\tiny{$\pm$0.02}}} &           0.64\textcolor{gray}{\tiny{$\pm$0.02}} &           0.66\textcolor{gray}{\tiny{$\pm$0.02}} &           0.67\textcolor{gray}{\tiny{$\pm$0.02}} \\
 & DKT               &           0.58\textcolor{gray}{\tiny{$\pm$0.03}} &           0.63\textcolor{gray}{\tiny{$\pm$0.02}} &           0.61\textcolor{gray}{\tiny{$\pm$0.01}} &           0.57\textcolor{gray}{\tiny{$\pm$0.01}} &           0.55\textcolor{gray}{\tiny{$\pm$0.01}} &           0.21\textcolor{gray}{\tiny{$\pm$0.02}} &           0.48\textcolor{gray}{\tiny{$\pm$0.02}} &           0.65\textcolor{gray}{\tiny{$\pm$0.02}} &           0.74\textcolor{gray}{\tiny{$\pm$0.02}} &           0.79\textcolor{gray}{\tiny{$\pm$0.01}} \\
 & SimpleShot        &           0.06\textcolor{gray}{\tiny{$\pm$0.02}} &           0.64\textcolor{gray}{\tiny{$\pm$0.03}} &           0.62\textcolor{gray}{\tiny{$\pm$0.02}} &           0.50\textcolor{gray}{\tiny{$\pm$0.01}} &           0.44\textcolor{gray}{\tiny{$\pm$0.01}} &           0.01\textcolor{gray}{\tiny{$\pm$0.00}} &           0.22\textcolor{gray}{\tiny{$\pm$0.02}} &           0.56\textcolor{gray}{\tiny{$\pm$0.02}} &  \textbf{0.76\textcolor{gray}{\tiny{$\pm$0.01}}} &  \textbf{0.87\textcolor{gray}{\tiny{$\pm$0.01}}} \\
 & MAML              &           0.07\textcolor{gray}{\tiny{$\pm$0.02}} &           0.45\textcolor{gray}{\tiny{$\pm$0.03}} &           0.50\textcolor{gray}{\tiny{$\pm$0.02}} &           0.42\textcolor{gray}{\tiny{$\pm$0.01}} &           0.36\textcolor{gray}{\tiny{$\pm$0.01}} &           0.01\textcolor{gray}{\tiny{$\pm$0.00}} &           0.16\textcolor{gray}{\tiny{$\pm$0.02}} &           0.44\textcolor{gray}{\tiny{$\pm$0.02}} &           0.64\textcolor{gray}{\tiny{$\pm$0.02}} &           0.81\textcolor{gray}{\tiny{$\pm$0.01}} \\
 & ProtoMAML         &           0.56\textcolor{gray}{\tiny{$\pm$0.03}} &           0.64\textcolor{gray}{\tiny{$\pm$0.02}} &           0.60\textcolor{gray}{\tiny{$\pm$0.01}} &           0.54\textcolor{gray}{\tiny{$\pm$0.01}} &           0.49\textcolor{gray}{\tiny{$\pm$0.01}} &           0.18\textcolor{gray}{\tiny{$\pm$0.02}} &           0.45\textcolor{gray}{\tiny{$\pm$0.02}} &           0.63\textcolor{gray}{\tiny{$\pm$0.02}} &           0.72\textcolor{gray}{\tiny{$\pm$0.01}} &           0.79\textcolor{gray}{\tiny{$\pm$0.01}} \\
 & BTAML             &  \textbf{0.59\textcolor{gray}{\tiny{$\pm$0.02}}} &           0.64\textcolor{gray}{\tiny{$\pm$0.02}} &           0.62\textcolor{gray}{\tiny{$\pm$0.01}} &           0.59\textcolor{gray}{\tiny{$\pm$0.01}} &           0.56\textcolor{gray}{\tiny{$\pm$0.01}} &           0.32\textcolor{gray}{\tiny{$\pm$0.02}} &           0.50\textcolor{gray}{\tiny{$\pm$0.02}} &           0.65\textcolor{gray}{\tiny{$\pm$0.02}} &           0.72\textcolor{gray}{\tiny{$\pm$0.01}} &           0.76\textcolor{gray}{\tiny{$\pm$0.01}} \\
\bottomrule
\end{tabular}
    }
    \scalebox{0.9}{
    \vspace{0.5mm}
    \begin{tabular}{cl|ccccc|ccccc}
\toprule
 & {} & \multicolumn{5}{c}{Precision (95\%CI)} & \multicolumn{5}{c}{Recall (95\%CI)} \\
 & {} &                                          $K_0=1$ &                                          $K_1=3$ &                                          $K_2=5$ &                                          $K_3=7$ &                                          $K_4=9$ &                                          $K_0=1$ &                                          $K_1=3$ &                                          $K_2=5$ &                                          $K_3=7$ &                                          $K_4=9$ \\
\midrule
 \parbox[t]{3mm}{\multirow{9}{*}{\rotatebox[origin=c]{90}{\specialcell{Random-Shot\\ $\rightarrow$ ROS+}}}} & Matching Net      &           0.53\textcolor{gray}{\tiny{$\pm$0.02}} &           0.60\textcolor{gray}{\tiny{$\pm$0.02}} &           0.61\textcolor{gray}{\tiny{$\pm$0.01}} &           0.56\textcolor{gray}{\tiny{$\pm$0.01}} &           0.58\textcolor{gray}{\tiny{$\pm$0.01}} &           0.44\textcolor{gray}{\tiny{$\pm$0.02}} &           0.52\textcolor{gray}{\tiny{$\pm$0.02}} &           0.57\textcolor{gray}{\tiny{$\pm$0.02}} &           0.65\textcolor{gray}{\tiny{$\pm$0.02}} &           0.62\textcolor{gray}{\tiny{$\pm$0.02}} \\
 & ProtoNet          &           0.45\textcolor{gray}{\tiny{$\pm$0.02}} &           0.49\textcolor{gray}{\tiny{$\pm$0.02}} &           0.49\textcolor{gray}{\tiny{$\pm$0.01}} &           0.49\textcolor{gray}{\tiny{$\pm$0.01}} &           0.48\textcolor{gray}{\tiny{$\pm$0.01}} &           0.28\textcolor{gray}{\tiny{$\pm$0.02}} &           0.45\textcolor{gray}{\tiny{$\pm$0.02}} &           0.52\textcolor{gray}{\tiny{$\pm$0.02}} &           0.56\textcolor{gray}{\tiny{$\pm$0.02}} &           0.57\textcolor{gray}{\tiny{$\pm$0.02}} \\
 & ProtoNet (20-way) &           0.55\textcolor{gray}{\tiny{$\pm$0.02}} &           0.58\textcolor{gray}{\tiny{$\pm$0.02}} &           0.57\textcolor{gray}{\tiny{$\pm$0.01}} &           0.56\textcolor{gray}{\tiny{$\pm$0.01}} &           0.55\textcolor{gray}{\tiny{$\pm$0.01}} &           0.31\textcolor{gray}{\tiny{$\pm$0.02}} &           0.52\textcolor{gray}{\tiny{$\pm$0.02}} &           0.60\textcolor{gray}{\tiny{$\pm$0.02}} &           0.64\textcolor{gray}{\tiny{$\pm$0.02}} &           0.66\textcolor{gray}{\tiny{$\pm$0.02}} \\
 & Relation Net (CE) &           0.56\textcolor{gray}{\tiny{$\pm$0.02}} &           0.60\textcolor{gray}{\tiny{$\pm$0.02}} &  \textbf{0.62\textcolor{gray}{\tiny{$\pm$0.01}}} &  \textbf{0.61\textcolor{gray}{\tiny{$\pm$0.01}}} &  \textbf{0.61\textcolor{gray}{\tiny{$\pm$0.01}}} &  \textbf{0.45\textcolor{gray}{\tiny{$\pm$0.02}}} &  \textbf{0.57\textcolor{gray}{\tiny{$\pm$0.02}}} &           0.63\textcolor{gray}{\tiny{$\pm$0.02}} &           0.65\textcolor{gray}{\tiny{$\pm$0.02}} &           0.67\textcolor{gray}{\tiny{$\pm$0.02}} \\
 & DKT               &  \textbf{0.59\textcolor{gray}{\tiny{$\pm$0.02}}} &           0.61\textcolor{gray}{\tiny{$\pm$0.01}} &           0.62\textcolor{gray}{\tiny{$\pm$0.01}} &           0.60\textcolor{gray}{\tiny{$\pm$0.01}} &           0.59\textcolor{gray}{\tiny{$\pm$0.01}} &           0.37\textcolor{gray}{\tiny{$\pm$0.02}} &           0.55\textcolor{gray}{\tiny{$\pm$0.02}} &           0.64\textcolor{gray}{\tiny{$\pm$0.02}} &           0.68\textcolor{gray}{\tiny{$\pm$0.02}} &           0.71\textcolor{gray}{\tiny{$\pm$0.02}} \\
 & SimpleShot        &           0.52\textcolor{gray}{\tiny{$\pm$0.03}} &  \textbf{0.63\textcolor{gray}{\tiny{$\pm$0.02}}} &           0.59\textcolor{gray}{\tiny{$\pm$0.01}} &           0.54\textcolor{gray}{\tiny{$\pm$0.01}} &           0.51\textcolor{gray}{\tiny{$\pm$0.01}} &           0.13\textcolor{gray}{\tiny{$\pm$0.01}} &           0.43\textcolor{gray}{\tiny{$\pm$0.02}} &           0.61\textcolor{gray}{\tiny{$\pm$0.02}} &  \textbf{0.72\textcolor{gray}{\tiny{$\pm$0.01}}} &  \textbf{0.77\textcolor{gray}{\tiny{$\pm$0.01}}} \\
 & MAML              &           0.42\textcolor{gray}{\tiny{$\pm$0.02}} &           0.46\textcolor{gray}{\tiny{$\pm$0.01}} &           0.48\textcolor{gray}{\tiny{$\pm$0.01}} &           0.49\textcolor{gray}{\tiny{$\pm$0.01}} &           0.49\textcolor{gray}{\tiny{$\pm$0.01}} &           0.41\textcolor{gray}{\tiny{$\pm$0.02}} &           0.47\textcolor{gray}{\tiny{$\pm$0.02}} &           0.51\textcolor{gray}{\tiny{$\pm$0.02}} &           0.49\textcolor{gray}{\tiny{$\pm$0.02}} &           0.50\textcolor{gray}{\tiny{$\pm$0.02}} \\
 & ProtoMAML         &           0.57\textcolor{gray}{\tiny{$\pm$0.02}} &           0.59\textcolor{gray}{\tiny{$\pm$0.01}} &           0.60\textcolor{gray}{\tiny{$\pm$0.01}} &           0.59\textcolor{gray}{\tiny{$\pm$0.01}} &           0.58\textcolor{gray}{\tiny{$\pm$0.01}} &           0.40\textcolor{gray}{\tiny{$\pm$0.02}} &           0.56\textcolor{gray}{\tiny{$\pm$0.02}} &           0.63\textcolor{gray}{\tiny{$\pm$0.02}} &           0.66\textcolor{gray}{\tiny{$\pm$0.02}} &           0.68\textcolor{gray}{\tiny{$\pm$0.02}} \\
 & BTAML             &           0.58\textcolor{gray}{\tiny{$\pm$0.02}} &           0.61\textcolor{gray}{\tiny{$\pm$0.01}} &           0.61\textcolor{gray}{\tiny{$\pm$0.01}} &           0.61\textcolor{gray}{\tiny{$\pm$0.01}} &           0.61\textcolor{gray}{\tiny{$\pm$0.01}} &           0.42\textcolor{gray}{\tiny{$\pm$0.02}} &           0.57\textcolor{gray}{\tiny{$\pm$0.02}} &  \textbf{0.65\textcolor{gray}{\tiny{$\pm$0.02}}} &           0.68\textcolor{gray}{\tiny{$\pm$0.02}} &           0.69\textcolor{gray}{\tiny{$\pm$0.02}} \\
\bottomrule
\end{tabular}
    }
    \scalebox{0.9}{
    \vspace{0.5mm}
    \begin{tabular}{cl|ccccc|ccccc}
\toprule
 & {} & \multicolumn{5}{c}{Precision (95\%CI)} & \multicolumn{5}{c}{Recall (95\%CI)} \\
 & {} &                                          $K_0=1$ &                                          $K_1=3$ &                                          $K_2=5$ &                                          $K_3=7$ &                                          $K_4=9$ &                                          $K_0=1$ &                                          $K_1=3$ &                                          $K_2=5$ &                                          $K_3=7$ &                                          $K_4=9$ \\
\midrule
 \parbox[t]{3mm}{\multirow{9}{*}{\rotatebox[origin=c]{90}{\specialcell{Random-Shot \\\& ROS+ $\rightarrow$ ROS+}}}} & Matching Net      &           0.55\textcolor{gray}{\tiny{$\pm$0.02}} &           0.60\textcolor{gray}{\tiny{$\pm$0.02}} &  \textbf{0.62\textcolor{gray}{\tiny{$\pm$0.01}}} &           0.59\textcolor{gray}{\tiny{$\pm$0.01}} &           0.56\textcolor{gray}{\tiny{$\pm$0.01}} &           0.43\textcolor{gray}{\tiny{$\pm$0.02}} &           0.47\textcolor{gray}{\tiny{$\pm$0.02}} &           0.58\textcolor{gray}{\tiny{$\pm$0.02}} &           0.66\textcolor{gray}{\tiny{$\pm$0.02}} &           0.72\textcolor{gray}{\tiny{$\pm$0.02}} \\
 & ProtoNet          &           0.59\textcolor{gray}{\tiny{$\pm$0.03}} &           0.55\textcolor{gray}{\tiny{$\pm$0.01}} &           0.56\textcolor{gray}{\tiny{$\pm$0.01}} &           0.59\textcolor{gray}{\tiny{$\pm$0.01}} &           0.63\textcolor{gray}{\tiny{$\pm$0.01}} &           0.28\textcolor{gray}{\tiny{$\pm$0.02}} &           0.58\textcolor{gray}{\tiny{$\pm$0.02}} &           0.67\textcolor{gray}{\tiny{$\pm$0.02}} &           0.68\textcolor{gray}{\tiny{$\pm$0.02}} &           0.64\textcolor{gray}{\tiny{$\pm$0.02}} \\
 & ProtoNet (20-way) &  \textbf{0.62\textcolor{gray}{\tiny{$\pm$0.03}}} &           0.56\textcolor{gray}{\tiny{$\pm$0.01}} &           0.57\textcolor{gray}{\tiny{$\pm$0.01}} &  \textbf{0.62\textcolor{gray}{\tiny{$\pm$0.01}}} &  \textbf{0.68\textcolor{gray}{\tiny{$\pm$0.01}}} &           0.25\textcolor{gray}{\tiny{$\pm$0.02}} &  \textbf{0.63\textcolor{gray}{\tiny{$\pm$0.02}}} &  \textbf{0.72\textcolor{gray}{\tiny{$\pm$0.02}}} &           0.69\textcolor{gray}{\tiny{$\pm$0.02}} &           0.63\textcolor{gray}{\tiny{$\pm$0.02}} \\
 & Relation Net (CE) &           0.55\textcolor{gray}{\tiny{$\pm$0.02}} &           0.57\textcolor{gray}{\tiny{$\pm$0.02}} &           0.60\textcolor{gray}{\tiny{$\pm$0.01}} &           0.60\textcolor{gray}{\tiny{$\pm$0.01}} &           0.62\textcolor{gray}{\tiny{$\pm$0.01}} &           0.41\textcolor{gray}{\tiny{$\pm$0.02}} &           0.54\textcolor{gray}{\tiny{$\pm$0.02}} &           0.63\textcolor{gray}{\tiny{$\pm$0.02}} &           0.66\textcolor{gray}{\tiny{$\pm$0.02}} &           0.66\textcolor{gray}{\tiny{$\pm$0.02}} \\
 & DKT               &           0.56\textcolor{gray}{\tiny{$\pm$0.02}} &  \textbf{0.61\textcolor{gray}{\tiny{$\pm$0.02}}} &           0.62\textcolor{gray}{\tiny{$\pm$0.01}} &           0.58\textcolor{gray}{\tiny{$\pm$0.01}} &           0.57\textcolor{gray}{\tiny{$\pm$0.01}} &           0.37\textcolor{gray}{\tiny{$\pm$0.02}} &           0.45\textcolor{gray}{\tiny{$\pm$0.02}} &           0.61\textcolor{gray}{\tiny{$\pm$0.02}} &           0.70\textcolor{gray}{\tiny{$\pm$0.02}} &           0.76\textcolor{gray}{\tiny{$\pm$0.01}} \\
 & SimpleShot        &           0.50\textcolor{gray}{\tiny{$\pm$0.04}} &           0.56\textcolor{gray}{\tiny{$\pm$0.02}} &           0.58\textcolor{gray}{\tiny{$\pm$0.02}} &           0.52\textcolor{gray}{\tiny{$\pm$0.01}} &           0.52\textcolor{gray}{\tiny{$\pm$0.01}} &           0.11\textcolor{gray}{\tiny{$\pm$0.01}} &           0.41\textcolor{gray}{\tiny{$\pm$0.02}} &           0.51\textcolor{gray}{\tiny{$\pm$0.02}} &  \textbf{0.76\textcolor{gray}{\tiny{$\pm$0.01}}} &  \textbf{0.79\textcolor{gray}{\tiny{$\pm$0.01}}} \\
 & MAML              &           0.48\textcolor{gray}{\tiny{$\pm$0.02}} &           0.47\textcolor{gray}{\tiny{$\pm$0.01}} &           0.52\textcolor{gray}{\tiny{$\pm$0.01}} &           0.53\textcolor{gray}{\tiny{$\pm$0.01}} &           0.56\textcolor{gray}{\tiny{$\pm$0.01}} &           0.38\textcolor{gray}{\tiny{$\pm$0.02}} &           0.47\textcolor{gray}{\tiny{$\pm$0.02}} &           0.54\textcolor{gray}{\tiny{$\pm$0.02}} &           0.59\textcolor{gray}{\tiny{$\pm$0.02}} &           0.60\textcolor{gray}{\tiny{$\pm$0.02}} \\
 & ProtoMAML         &           0.60\textcolor{gray}{\tiny{$\pm$0.02}} &           0.61\textcolor{gray}{\tiny{$\pm$0.01}} &           0.61\textcolor{gray}{\tiny{$\pm$0.01}} &           0.60\textcolor{gray}{\tiny{$\pm$0.01}} &           0.59\textcolor{gray}{\tiny{$\pm$0.01}} &           0.38\textcolor{gray}{\tiny{$\pm$0.02}} &           0.57\textcolor{gray}{\tiny{$\pm$0.02}} &           0.65\textcolor{gray}{\tiny{$\pm$0.02}} &           0.69\textcolor{gray}{\tiny{$\pm$0.01}} &           0.70\textcolor{gray}{\tiny{$\pm$0.01}} \\
 & BTAML             &           0.59\textcolor{gray}{\tiny{$\pm$0.02}} &           0.60\textcolor{gray}{\tiny{$\pm$0.01}} &           0.61\textcolor{gray}{\tiny{$\pm$0.01}} &           0.62\textcolor{gray}{\tiny{$\pm$0.01}} &           0.62\textcolor{gray}{\tiny{$\pm$0.01}} &  \textbf{0.44\textcolor{gray}{\tiny{$\pm$0.02}}} &           0.59\textcolor{gray}{\tiny{$\pm$0.02}} &           0.65\textcolor{gray}{\tiny{$\pm$0.02}} &           0.67\textcolor{gray}{\tiny{$\pm$0.02}} &           0.67\textcolor{gray}{\tiny{$\pm$0.02}} \\
\bottomrule
\end{tabular}
    }
\end{table}

\clearpage

\subsection{Imbalanced Query Set vs Imbalanced Support Set}\label{appendix imbalanced query}
In the main paper, we examined the problem of imbalanced support sets. In this section, we investigate imbalanced query sets in the meta-training task to see how they affect generalization to imbalanced support sets at test time. MAML-like algorithms often involve second-order derivative approximations, so they are not directly influenced by imbalance at the support set level. Metric-learning algorithms such as ProtoNet also optimize the query set performance rather than the support set loss. Thus, imbalance in the query set has a direct effect on the optimization of meta-learners. This section investigates four meta-training settings, along two axes:  1) balanced vs imbalanced, and 2) support vs query set. Here, we define imbalanced tasks as having 1-9 shot/queries linear imbalance in the support/query set and 5shot/query in the balanced equivalent. At meta-test time, all methods are evaluated on tasks with imbalanced support sets and balanced query sets. We want to investigate whether an imbalanced query set during meta-training affects the imbalanced support set task during evaluation. Table~\ref{tbl imbalanced query} shows the performance of algorithms with balanced and imbalanced query sets. The results show that having the query set balanced during meta-learning is generally preferred by metric-based methods; however, the trend is not so clear for optimization-based algorithms. Future work could investigate the effect of query set imbalance on transductive inference methods \cite{Antoniou2019sca,liu2019learning,Dhillon2020baseline,boudiaf2020transductive}. 

\begin{table}[htb]
    \caption{} 
    \label{tbl imbalanced query}
    \centering
    \scalebox{0.99}{
    \begin{tabular}{c|cc|c}
\toprule
Model               &   Support Imbalance   &   Query Imbalance &   Accuracy    \\
\midrule
MAML                &   balanced            &   balanced        &   47.50         \\
MAML                &   linear              &   balanced        &   44.98        \\
MAML                &   balanced            &   linear          &   37.34       \\
MAML                &   linear              &   linear          &   \textbf{47.87}        \\
\midrule
ProtoMAML                &   balanced        &   balanced        &   \textbf{46.60}        \\
ProtoMAML                &   linear          &   balanced        &   44.86        \\
ProtoMAML                &   balanced        &   linear          &   46.16        \\
ProtoMAML                &   linear          &   linear          &   45.99        \\
\midrule
ProtoNet                &   balanced        &   balanced        &   \textbf{46.60}        \\
ProtoNet                &   linear          &   balanced        &   44.86        \\
ProtoNet                &   balanced        &   linear          &   46.16        \\
ProtoNet                &   linear          &   linear          &   45.99        \\
\midrule
Relation Net (CE)                &   balanced        &   balanced       &   56.16       \\
Relation Net (CE)                &   linear          &   balanced       &   \textbf{58.58}        \\
Relation Net (CE)                &   balanced        &   linear         &   55.86        \\
Relation Net (CE)                &   linear          &   linear         &   54.82       \\
\midrule
DKT                &   balanced         &   balanced        &   \textbf{56.41}        \\
DKT                &   linear           &   balanced        &   55.73        \\
DKT                &   balanced         &   linear          &   55.86        \\
DKT                &   linear           &   linear          &   54.34        \\
\bottomrule
\end{tabular}

    }
\end{table}

\subsection{Small, Imbalanced Meta-Training Dataset}\label{appendix reduced}
In this section, we examine whether a smaller dataset size could influence the effect of imbalance. Specifically, we construct a new set of datasets denoted by $\mathcal{D}''_{train}$ containing 1/8\textsuperscript{th} of samples in $\mathcal{D}_{train}$ of Mini-ImageNet, and 32 classes selected uniformly at random, $|\mathcal{D}''_{train}|=4800$. Table~\ref{tbl redu_mini_table_combined} are a break-down of Figure~\ref{fig reduced_meta}. Overall, the performance drops as the number of minority classes increases, but the effect remains quite low ($<3\%$ absolute difference).

\begin{table}[htb]
    \caption{Evaluation accuracy after meta-training on $\mathcal{D}''_{train}$ derived from Mini-ImageNet with various imbalance distributions. Small differences in accuracy between $balanced$ and other distributions, suggest a small effect of imbalance at dataset level. \emph{Left:} Evaluation on the meta-testing dataset of Mini-ImageNet. \emph{Right:} Evaluation on the meta-testing dataset of CUB.
    } \label{tbl redu_mini_table_combined}
    \centering
    \scalebox{0.9}{ \begin{tabular}{r|cccc|cccc}
\toprule
$\mathcal{D}''_{train} \rightarrow \mathcal{D}_{test}$ & \multicolumn{4}{c|}{Mini-ImageNet $\rightarrow$ Mini-ImageNet} & \multicolumn{4}{c}{Mini-ImageNet $\rightarrow$ CUB} \\
Imbalance in $\mathcal{D}''_{train}$  &  \specialcell{ $balanced$ \\ {\scriptsize(150, 150, 32, -) }} & \specialcell{  $step$-8 \\ {\scriptsize(30, 190, 32, 8)} } & \specialcell{  $step$-16 \\ {\scriptsize(30, 270, 32, 16)} } & \specialcell{  $step$-24 \\ {\scriptsize(30, 510, 32, 24)} }  &  \specialcell{ $balanced$ \\ {\scriptsize(150, 150, 32, -) }} & \specialcell{  $step$-8 \\ {\scriptsize(30, 190, 32, 8)} } & \specialcell{  $step$-16 \\ {\scriptsize(30, 270, 32, 16)} } & \specialcell{  $step$-24 \\ {\scriptsize(30, 510, 32, 24)} } \\
\midrule
Baseline (fine-tune) &           56.07\textcolor{gray}{\tiny{$\pm$0.71}} &           56.19\textcolor{gray}{\tiny{$\pm$0.72}} &           56.17\textcolor{gray}{\tiny{$\pm$0.71}} &           55.81\textcolor{gray}{\tiny{$\pm$0.72}} &           55.80\textcolor{gray}{\tiny{$\pm$0.68}} &           55.19\textcolor{gray}{\tiny{$\pm$0.70}} &           55.14\textcolor{gray}{\tiny{$\pm$0.70}} &           53.38\textcolor{gray}{\tiny{$\pm$0.69}} \\
Baseline++           &           54.35\textcolor{gray}{\tiny{$\pm$0.67}} &           54.24\textcolor{gray}{\tiny{$\pm$0.67}} &           53.90\textcolor{gray}{\tiny{$\pm$0.67}} &           53.25\textcolor{gray}{\tiny{$\pm$0.68}} &           50.72\textcolor{gray}{\tiny{$\pm$0.69}} &           50.64\textcolor{gray}{\tiny{$\pm$0.70}} &           50.44\textcolor{gray}{\tiny{$\pm$0.69}} &           49.50\textcolor{gray}{\tiny{$\pm$0.67}} \\
Matching Net         &           55.79\textcolor{gray}{\tiny{$\pm$0.68}} &           54.76\textcolor{gray}{\tiny{$\pm$0.69}} &           53.44\textcolor{gray}{\tiny{$\pm$0.68}} &           51.54\textcolor{gray}{\tiny{$\pm$0.67}} &           52.27\textcolor{gray}{\tiny{$\pm$0.71}} &           50.44\textcolor{gray}{\tiny{$\pm$0.72}} &           49.61\textcolor{gray}{\tiny{$\pm$0.72}} &           49.60\textcolor{gray}{\tiny{$\pm$0.71}} \\
ProtoNet             &           55.90\textcolor{gray}{\tiny{$\pm$0.71}} &           54.71\textcolor{gray}{\tiny{$\pm$0.70}} &           54.75\textcolor{gray}{\tiny{$\pm$0.71}} &           53.75\textcolor{gray}{\tiny{$\pm$0.71}} &           51.24\textcolor{gray}{\tiny{$\pm$0.73}} &           50.93\textcolor{gray}{\tiny{$\pm$0.73}} &           50.96\textcolor{gray}{\tiny{$\pm$0.72}} &           49.47\textcolor{gray}{\tiny{$\pm$0.73}} \\
Relation Net (CE)    &           55.41\textcolor{gray}{\tiny{$\pm$0.70}} &           53.80\textcolor{gray}{\tiny{$\pm$0.69}} &           51.15\textcolor{gray}{\tiny{$\pm$0.70}} &           50.24\textcolor{gray}{\tiny{$\pm$0.68}} &           51.42\textcolor{gray}{\tiny{$\pm$0.70}} &           50.06\textcolor{gray}{\tiny{$\pm$0.72}} &           49.12\textcolor{gray}{\tiny{$\pm$0.71}} &           47.12\textcolor{gray}{\tiny{$\pm$0.70}} \\
DKT                  &           57.21\textcolor{gray}{\tiny{$\pm$0.67}} &           56.23\textcolor{gray}{\tiny{$\pm$0.68}} &           54.65\textcolor{gray}{\tiny{$\pm$0.67}} &           53.65\textcolor{gray}{\tiny{$\pm$0.69}} &  \textbf{56.36\textcolor{gray}{\tiny{$\pm$0.69}}} &           54.76\textcolor{gray}{\tiny{$\pm$0.68}} &           54.69\textcolor{gray}{\tiny{$\pm$0.69}} &  \textbf{54.98\textcolor{gray}{\tiny{$\pm$0.70}}} \\
SimpleShot           &  \textbf{58.80\textcolor{gray}{\tiny{$\pm$0.74}}} &  \textbf{58.61\textcolor{gray}{\tiny{$\pm$0.74}}} &  \textbf{58.46\textcolor{gray}{\tiny{$\pm$0.75}}} &  \textbf{58.05\textcolor{gray}{\tiny{$\pm$0.75}}} &           56.04\textcolor{gray}{\tiny{$\pm$0.71}} &  \textbf{55.73\textcolor{gray}{\tiny{$\pm$0.72}}} &  \textbf{56.01\textcolor{gray}{\tiny{$\pm$0.71}}} &           53.55\textcolor{gray}{\tiny{$\pm$0.72}} \\
MAML                 &           53.17\textcolor{gray}{\tiny{$\pm$0.75}} &           53.16\textcolor{gray}{\tiny{$\pm$0.74}} &           52.64\textcolor{gray}{\tiny{$\pm$0.75}} &           51.79\textcolor{gray}{\tiny{$\pm$0.73}} &           53.56\textcolor{gray}{\tiny{$\pm$0.73}} &           52.90\textcolor{gray}{\tiny{$\pm$0.76}} &           52.50\textcolor{gray}{\tiny{$\pm$0.72}} &           53.13\textcolor{gray}{\tiny{$\pm$0.73}} \\
ProtoMAML            &           57.75\textcolor{gray}{\tiny{$\pm$0.71}} &           56.74\textcolor{gray}{\tiny{$\pm$0.71}} &           56.05\textcolor{gray}{\tiny{$\pm$0.71}} &           54.34\textcolor{gray}{\tiny{$\pm$0.70}} &           54.85\textcolor{gray}{\tiny{$\pm$0.72}} &           54.02\textcolor{gray}{\tiny{$\pm$0.75}} &           53.16\textcolor{gray}{\tiny{$\pm$0.71}} &           53.20\textcolor{gray}{\tiny{$\pm$0.71}} \\
\midrule
\specialcell{Avr. Diff. to $balanced$} &         - & 0.7 &	-1.5 &	-2.4	& - &	-0.8	& -1.2	& -2.0 \\
\bottomrule
\end{tabular}

} 
\end{table}

\end{document}